\documentclass{article}

\usepackage{amsmath,amsfonts,bm}

\def\figref#1{Figure~\ref{#1}}

\def\tabref#1{Table~\ref{#1}}

\def\secref#1{Section~\ref{#1}}

\def\eqref#1{Eq.~(\ref{#1})}

\def\algref#1{Algorithm~\ref{#1}}

\def\apxref#1{Appendix~\ref{#1}}

\def\1{\bm{1}}

\DeclareMathAlphabet{\mathsfit}{\encodingdefault}{\sfdefault}{m}{sl}
\SetMathAlphabet{\mathsfit}{bold}{\encodingdefault}{\sfdefault}{bx}{n}

\def\gV{{\mathcal{V}}}

\usepackage[numbers, compress]{natbib}
\usepackage[eandd, final]{neurips_2026}

\usepackage[utf8]{inputenc} 
\usepackage[T1]{fontenc}    
\usepackage{hyperref}       
\usepackage{url}            
\usepackage{booktabs}       
\usepackage{amsfonts}       
\usepackage{nicefrac}       
\usepackage{microtype}      
\usepackage{xcolor}         
\usepackage{subcaption}
\usepackage{graphicx}
\usepackage{relsize}
\usepackage{amsmath}
\usepackage{amssymb}
\usepackage{multirow}
\usepackage[most]{tcolorbox}
\usepackage{makecell}
\usepackage{fvextra}
\usepackage{longtable}
\usepackage{tabularx}
\usepackage{pifont}
\usepackage{wasysym}
\usepackage{enumitem}
\usepackage{wrapfig}
\usepackage{siunitx}
\usepackage[table]{xcolor}
\usepackage{times}

\usepackage[ruled,vlined,linesnumbered]{algorithm2e}
\SetKwInOut{Input}{Input}
\SetKwInOut{Output}{Output}
\SetKw{Return}{return}

\definecolor{citecolor}{rgb}{0.21,0.49,0.74} 
\definecolor{linkcolor}{RGB}{255, 0, 0} 
\definecolor{urlcolor}{RGB}{255, 105, 180} 
\definecolor{TV}{HTML}{A65310}
\definecolor{Human}{HTML}{205F95}

\hypersetup{
  colorlinks=true, 
  urlcolor=urlcolor, 
  linkcolor=linkcolor, 
  citecolor=citecolor 
}

\newtcolorbox{promptbox}[1]{
  colback=gray!5,
  colframe=gray!60,
  title=#1,
  fonttitle=\small\bfseries,
  breakable,
  boxrule=0.5pt,
  arc=2pt,
  left=6pt,
  right=6pt,
  top=6pt,
  bottom=6pt
}

\newcommand{\paratitle}[1]{\noindent\textbf{#1}\enspace}
\newcommand{\cmark}{\ding{51}} 
\newcommand{\xmark}{\ding{55}} 

\newcommand{\casepanel}[6]{%
  \begin{minipage}[t]{0.237\linewidth}
    \vspace{0pt}\raggedright\fontsize{8}{10}\selectfont
    \parbox[t][12pt][t]{\linewidth}{%
      \fontsize{8}{10}\selectfont\bfseries #1%
    }\par
    \includegraphics[width=\linewidth]{#2}\par\vspace{-2pt}
    {\fontsize{7}{9}\selectfont
    \textcolor{darkgray}{\tiny{#5}}\par\vspace{3pt}
    \begin{tabular*}{\linewidth}{@{}l@{\extracolsep{\fill}}l@{}}
      \textcolor{TV}{\textbf{TerraVis}} & \textcolor{Human}{\textbf{Human}}\\
      \textcolor{TV}{$#3$} & \textcolor{Human}{$#4$}
    \end{tabular*}\par}\vspace{4pt}
    #6\par
  \end{minipage}}

\newcommand{\casegroup}[2]{%
  \par\noindent
  {\fontsize{8}{10}\selectfont\bfseries #1}%
  \hfill
  {\fontsize{7}{9}\selectfont\color{darkgray}#2}\par
  \vspace{2pt}
  {\color{gray!40}\hrule height 0.4pt}
  \vspace{4pt}\noindent
}

\newcommand{\casepanelwide}[5]{%
  \begin{minipage}[t]{0.485\linewidth}
    \vspace{0pt}
    \begin{minipage}[t]{0.48\linewidth}
      \vspace{0pt}\raggedright
      \includegraphics[width=\linewidth]{#1}\par\vspace{-2pt}
      {\fontsize{7}{9}\selectfont
        \textcolor{darkgray}{\tiny{#4}}\par\vspace{3pt}
        \noindent\makebox[\linewidth][s]{
          \textcolor{TV}{\textbf{T}~$#2$}
          \hfill
          \textcolor{Human}{\textbf{H}~$#3$}
      }\par
      }
    \end{minipage}\hfill
    \begin{minipage}[t]{0.50\linewidth}
      \vspace{0pt}\raggedright
      \fontsize{8}{10}\selectfont
      #5\par
    \end{minipage}
  \end{minipage}
}

\newcommand{\baselinecase}[7]{%
  \begin{minipage}[t]{0.485\linewidth}
    \vspace{0pt}\raggedright
    \begin{minipage}[t]{0.48\linewidth}
      \vspace{0pt}\raggedright
      \includegraphics[width=\linewidth]{#1}\par\vspace{1pt}
      {\tiny\color{darkgray}\mbox{#2}\par}%
    \end{minipage}\hfill
    \begin{minipage}[t]{0.50\linewidth}
      \vspace{0pt}\raggedright\fontsize{8}{10}\selectfont
      {\fontsize{7}{9}\selectfont
        #3\par\vspace{2pt}
        \textcolor{TV}{\textbf{TerraVis}~#4}\hfill
        \textcolor{Human}{\textbf{Human}~#5}\par}%
      \vspace{3pt}
      \textit{Prompt:} #6\par\vspace{3pt}
      \textit{Discussion:} #7\par
    \end{minipage}%
  \end{minipage}%
}

\newcommand{\rahfscores}[4]{%
  \textbf{RAHF}~#1\par\vspace{1pt}
  \begin{tabular}{@{}l@{\hspace{0.6em}}r@{}}
    RAHF-Plausibility & #2\\
    RAHF-Alignment    & #3\\
    RAHF-Aesthetics   & #4
  \end{tabular}%
}

\title{TerraVis: Towards Evaluation of World-Grounded Visual Consistency in Text-to-Image Generation via MLLM Workflows}

\author{
Shuai Fu$^{1}$
\quad
Jing Gu$^{2}$
\quad
Jian Zhou$^{1}$
\quad
Zicheng Duan$^{1}$
\quad
Gengze Zhou$^{1}$
\quad
Qi Wu$^{1,3}$\thanks{Corresponding Author}
\vspace{3pt}
\\
$^{1}$AIML, Adelaide University
\quad
$^{2}$xAI 
\quad
$^{3}$Responsible AI Research Centre 
\\
\small{\texttt{\{shuai.fu,j.zhou,zicheng.duan,gengze.zhou,qi.wu01\}@adelaide.edu.au}} \\
\small{\texttt{jgu110@ucsc.edu}}
}

\begin{document}

\maketitle

\vspace{-3mm}
\begin{abstract}
Recent text-to-image models have made substantial progress in photorealism, aesthetics, and text-image alignment. 
Yet visually appealing images can still violate real-world plausibility, exhibiting malformed object structures, impossible anatomy, physically implausible interactions, or inconsistent spatial relationships. 
Such failures are not well captured by existing fidelity, aesthetics, preference, or alignment metrics. To address this gap, we introduce \textbf{TerraVis}, a framework for evaluating \textbf{world-grounded visual consistency} in generated images. 
TerraVis defines a structured taxonomy of world-consistency violations spanning object-, interaction-, and scene-level failures, and employs a multi-stage evaluation framework to identify and quantify them. 
Given an image, TerraVis first uses an MLLM to assess its eligibility for evaluation, then detects violations across 18 taxonomy-defined types and classifies them as minor or major to derive an overall world-consistency score. 
Across diverse open-source and proprietary text-to-image models on two widely used benchmarks, TerraVis achieves the strongest correlation with human judgments of world consistency among existing metrics. 
Our benchmark results further show that models that achieve strong performance on conventional metrics can still exhibit substantial world-consistency failures. 
These findings highlight world consistency as a complementary evaluation dimension and demonstrate that TerraVis enables systematic quantification, diagnosis, and comparison of such failures. 
Our code is publicly available at \href{https://github.com/ShyFoo/TerraVis}{https://github.com/ShyFoo/TerraVis}.
\end{abstract}

\section{Introduction}
Generative models, especially text-to-image generation models, have made remarkable progress in recent years.
From earlier generative adversarial network (GAN)-based models~\cite{gan, wgan, srgan, cyclegan, stylegan, fu2023cooperative, biggan, stackgan}, to diffusion-based models~\cite{ddpm, adm, ldm, edm, glide, dalle-2, imagen}, and more recently autoregressive-based models~\cite{dalle, parti, var, emu3, janus, infinity, bagel} and diffusion-autoregressive hybrid models~\cite{transfusion, hart, hunyuanimage}, their ability to synthesize increasingly realistic imagery has advanced substantially.
This progress has been accompanied by the development of a rich ecosystem of evaluation metrics.
These span distributional similarity metrics such as Fréchet Inception Distance (FID)~\cite{fid}, text-image alignment metrics such as CLIPScore~\cite{clipscore}, human preference scores such as ImageReward~\cite{imagereward}, as well as technical quality metrics~\cite{niqe, brisque, musiq} and aesthetic quality metrics~\cite{ava, nima, aesthetic_v25}. 
By providing quantitative signals along these dimensions, existing metrics have served as important optimization objectives and evaluation references, encouraging progress in photorealism, aesthetics, and alignment with user prompts.

However, existing evaluation metrics are not designed to directly assess whether a generated image is visually consistent with the real-world referent it attempts to depict.
For example, a generated image may achieve a high level of photorealism, strong text-image alignment, or favorable human preference scores, while still containing implausible object structures, physically implausible interactions, impossible spatial relationships, or other world-level visual errors.
Such failures can undermine the reliability of generated images in applications where visual content is expected to represent a physically plausible and trustworthy depiction of the world.
We argue that a key reason this gap persists is the absence of an evaluation metric that explicitly targets this dimension.

\begin{figure}[t]
    \centering
    \includegraphics[width=\textwidth]{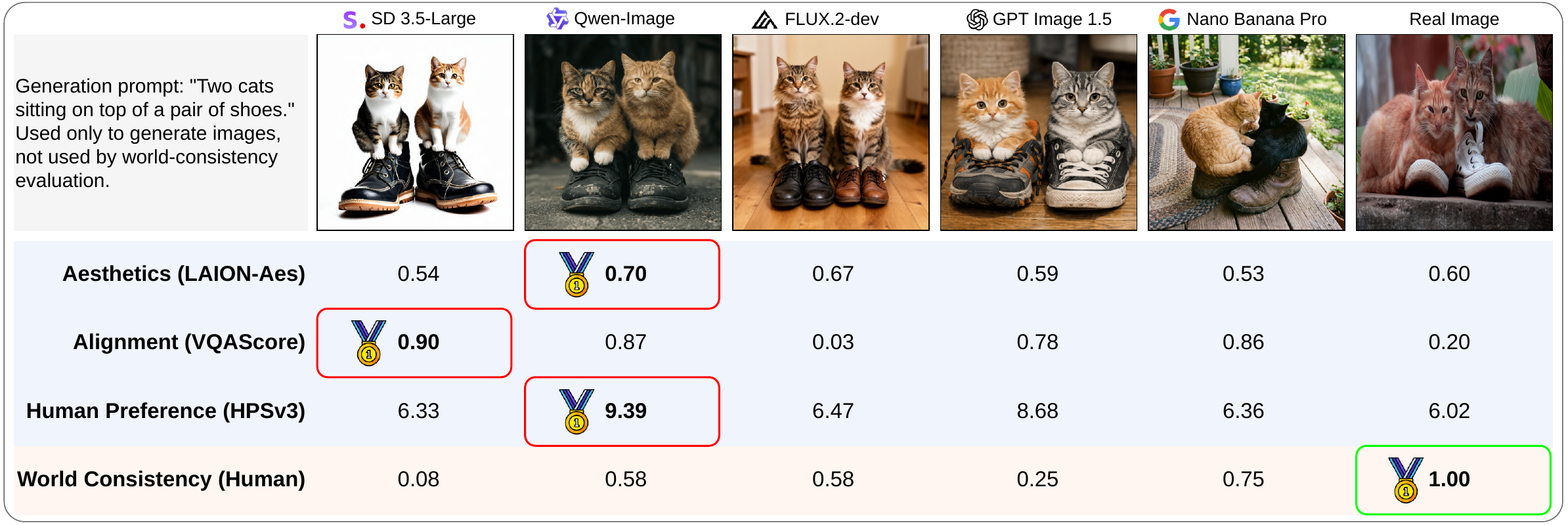}
    \caption{
    \textbf{World consistency is not reliably captured by existing metrics.}
    Given the same prompt, existing metrics may assign high aesthetics, alignment, or preference scores to generated images that still contain real-world implausibilities, such as cats appearing unnaturally embedded in, or physically supported by shoes.
    In contrast, \textbf{world consistency} evaluates whether the depicted entities, structures, interactions, and spatial relationships are visually plausible with respect to the real world.
    The real image is taken from MSCOCO~\cite{mscoco} and paired with the same caption used for generation.
    }
    \label{fig.1}
    \vspace{-4mm}
\end{figure}

To bridge this gap, we introduce the concept of \textbf{World-Grounded Visual Consistency}, referred to as \textbf{World Consistency}, which evaluates whether the depicted entities, structures, interactions, and phenomena are visually plausible with respect to the real world.
As illustrated in \figref{fig.1}, images that receive high aesthetic, alignment, or preference scores can still exhibit low world consistency.
This suggests that world consistency captures a complementary evaluation dimension for text-to-image generation. 
A structured comparison between world consistency and common text-to-image evaluation dimensions is provided in \tabref{tab.1}.
More discussions can be found in \apxref{apx: discussion_wc}.

Building on this concept, we propose \textbf{TerraVis}, a systematic evaluation framework for detecting and scoring world-grounded visual inconsistencies in generated images.
TerraVis first defines a fine-grained and structured taxonomy of world-consistency violations, covering common object-, interaction-, and scene-level failures.
It then leverages multimodal large language models (MLLMs) in a sequential workflow that checks evaluation eligibility, detects taxonomy-guided violations, classifies their severity, and aggregates the resulting major and minor violations into an overall score using a negative exponential function.
Experimental results show that TerraVis achieves stronger correlation with human judgments than \textbf{TerraVis-H}, a holistic MLLM baseline that is prompted with the same taxonomy descriptions and evaluation instructions provided to human annotators.
This demonstrates the advantage of decomposing world-consistency evaluation through taxonomy guidance and a multi-stage MLLM workflow.
Ablation results further show that our severity-aware exponential aggregation outperforms uniform checklist-style pass-rate aggregation, which treats all items equally and is commonly adopted in prior QA-based evaluation methods~\cite{tifa,dsg}.

This work makes the following contributions:
\begin{itemize}
    \item We introduce \textbf{world-grounded visual consistency} as a complementary evaluation dimension for text-to-image generation, focusing on perceptible violations of real-world structural, spatial, biological, and physical constraints.
    
    \item We propose \textbf{TerraVis}, a taxonomy-guided evaluation framework that decomposes world-consistency assessment into object-level, interaction-level, and scene-level violations through a multi-stage MLLM-based workflow.
    
    \item We benchmark TerraVis against existing quality, alignment, and preference metrics across multiple text-to-image models and datasets, showing substantially stronger correlation with human judgments of world consistency.
\end{itemize}

\begin{table}[t]
    \centering
    \setlength\tabcolsep{3pt}
    \caption{
    Comparison between world consistency and common evaluation dimensions in text-to-image generation. 
    Here, $\LEFTcircle$ indicates that the dimension might partially capture world plausibility, but does not explicitly target perceptible world-consistency violations.
    }
    \label{tab.1}
    \resizebox{\linewidth}{!}{%
    \begin{tabular}{l l c c l}
    \toprule
    \textbf{Evaluation Dimension} &
    \textbf{Representative Methods} &
    \textbf{Evaluation Unit} &
    \textbf{World Plausibility} &
    \textbf{Detection Target} \\
    \midrule
    Technical quality       & NIQE~\cite{niqe}, BRISQUE~\cite{brisque}, MUSIQ~\cite{musiq}                     & Single image            & \xmark  & Low-level image quality and artifacts \\
    Aesthetics              & AVA~\cite{ava}, NIMA~\cite{nima}, LAION-Aes~\cite{aesthetic_v25} & Single image            & \xmark  & Visual appeal and artistic quality \\
    Photorealism            & Human rating~\cite{stackgan, srgan, glide, dalle-2, imagen}                      & Single image            & $\LEFTcircle$ & Photo-like visual appearance \\
    Text-image alignment    & CLIPScore~\cite{clipscore}, TIFA~\cite{tifa}, VQAScore~\cite{vqascore}           & Text-image pair       & $\LEFTcircle$ & Text-image semantic consistency \\
    Human preference        & ImageReward~\cite{imagereward}, PickScore~\cite{pickscore}, HPS~\cite{hps}       & Prompt-cond. image(s)       & $\LEFTcircle$ & Overall subjective human preference \\
    Distributional similarity & FID~\cite{fid}, KID~\cite{kid}, Precision \& Recall~\cite{precision_recall}    & Ref./gen.\ image sets   & $\LEFTcircle$ & Distribution-level similarity to training data \\
    \midrule
    \textbf{World consistency} & \textbf{TerraVis (ours)} & Single image  & \cmark  & Perceptible violations of real-world constraints \\
    \bottomrule
    \end{tabular}}
    \vspace{-4mm}
\end{table}

\section{Related Work}
\label{sec2: related_work}

\subsection{Prompt-Conditioned Metrics for Generated Images}
\label{sec2.1: related_work--prompt_cond}

\textbf{Image-text alignment} metrics have been widely used to evaluate text-to-image models~\cite{clipscore, tifa, vpeval, llmscore, dsg, viescore, vqascore}.
For example, CLIPScore~\cite{clipscore} is originally proposed to measure the semantic similarity between text prompts and generated images.
TIFA~\cite{tifa} decomposes a text prompt into multiple question-answer pairs using large language models (LLMs), enabling a more fine-grained evaluation of text-to-image alignment. 
VQAScore adopts a simpler approach by deriving an alignment score from the probability of the “Yes" token under a fixed text-image alignment template.

Recently, a growing number of \textbf{human preference} scores have been proposed to better align the outputs of text-to-image models with human preferences~\cite{imagereward, pickscore, hps, hpsv2, mps, hpsv3}.  
Unlike image-text alignment metrics, these methods aim to capture how well a generated image matches human judgments of overall quality and preference.  
Representative examples include ImageReward~\cite{imagereward}, PickScore~\cite{pickscore}, and the Human Preference Score (HPS) series~\cite{hps, hpsv2, hpsv3}, which are typically trained on large-scale human preference data.  
As a result, they provide a complementary perspective that goes beyond traditional text-image semantic alignment metrics.

\subsection{Prompt-Independent Metrics for Generated Images}
\label{sec2.2: related_work--prompt_ind}

\textbf{Technical quality} metrics, such as NIQE~\cite{niqe}, BRISQUE~\cite{brisque}, and MUSIQ~\cite{musiq}, have also been used to evaluate generated images. 
These metrics focus on the intrinsic technical quality of an image, measuring factors such as blur, compression artifacts, and distortion. 
However, their importance has diminished as modern text-to-image models~\cite{sd_v3, qwen-image, flux2} generally no longer exhibit the obvious low-quality artifacts common in earlier generations, particularly GAN-based models~\cite{gan, wgan, cyclegan, stylegan, biggan}.

In contrast, \textbf{aesthetic quality} metrics~\cite{ava, nima, aesthetic_v25} assess image quality in terms of visual appeal, including composition, pleasing lighting, harmonious color combinations, clear subject emphasis, and overall stylistic attractiveness. 
These metrics are therefore particularly useful in applications such as photography, poster design, advertising, and other scenarios where generated images need to be ranked by aesthetic preference rather than factual correctness.

More recently, \citet{counthallu} found that better FID or MUSIQ scores do not necessarily correspond to fewer counting hallucinations, in which images depict incorrect numbers of object parts or instances.
While such errors can be viewed as a specific form of real-world consistency violation, we focus on \textbf{world-grounded visual consistency}, a prompt-independent evaluation target assessing whether the generated image itself conforms to real-world constraints.
This target encompasses a broader range of visual failures than counting errors alone and complements existing prompt-independent metrics.

\section{Evaluation of World-Grounded Visual Consistency Using TerraVis}
\label{sec3: terravis}

This section presents our definition of world-grounded visual consistency in generated images and describes how TerraVis evaluates it through a carefully designed protocol.

\subsection{World-Grounded Visual Consistency}
\label{sec3.1: wc}

\paratitle{General definition.}
World-grounded visual consistency (hereafter, \textit{world consistency}) measures the extent to which a representational depiction of a macroscopic object or scene visually conforms to real-world constraints.
These constraints include, for example, plausible object structure and anatomy, valid text or symbolic content when present, physically plausible interactions and environmental effects, spatially plausible scale, depth ordering, and continuity, as well as commonsense compatibility between objects, actions, events, and scene context.

\paratitle{Evaluation scope.}
World consistency is not tied to photorealism, so it can also be assessed in stylized images such as cartoons and sketches, provided they depict identifiable real-world objects or scenes with sufficient visual structure to support plausibility judgments.
However, the following content types are excluded: purely abstract content (e.g., color blocks and geometric patterns); symbolic content (e.g., logos, icons, emblems, flags, and emojis); informational content (e.g., charts, diagrams, maps, and tables); interface content (e.g., UI mockups and app/website screenshots); layout-based compositions (e.g., posters, advertisements, book covers, collages, multi-panel layouts); and microscopic content (e.g., cell structures, fabric fibers, and material surfaces).
These categories are excluded because they do not depict content depicting macroscopic objects or scenes to which world-consistency constraints can be meaningfully applied.

\paratitle{Style-tolerant but world-constrained.}
Stylistic rendering choices that intentionally reduce photorealism, such as simplified texture, flattened color, schematic lighting, or attenuated high-frequency detail, do not in themselves constitute world-consistency violations. 
A violation arises only when stylized content contradicts fundamental real-world constraints, rather than merely departing from photorealistic appearance.
For instance, if a person’s hand interpenetrates a cup they are shown holding, this constitutes a world-consistency violation regardless of image style, because it violates the physical integrity of the depicted objects.

\begin{figure}[t]
    \centering
    \includegraphics[width=\textwidth]{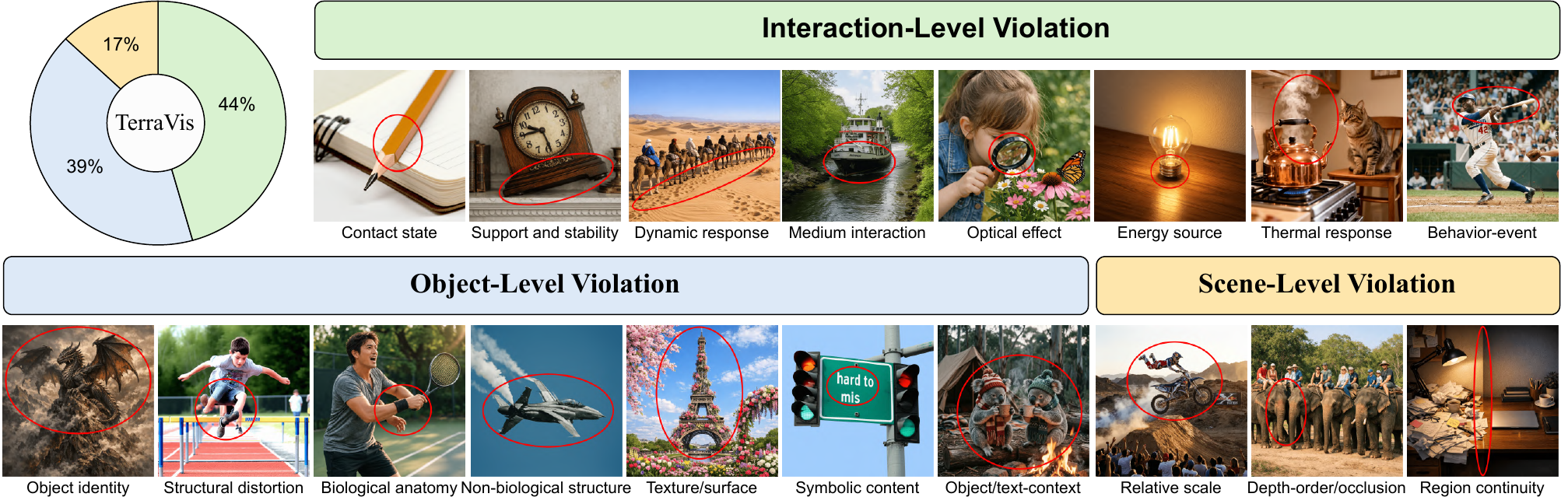}
    \caption{
    Overview of the TerraVis world-consistency violation taxonomy. 
    TerraVis organizes world-consistency violations into three levels: object-level, interaction-level, and scene-level violations. 
    The example images illustrate representative violation types within each level. 
    Red ellipses indicate the regions where the corresponding violations are visually observed.
    }
    \label{fig.2}
    \vspace{-4mm}
\end{figure}

\subsection{TerraVis Violation Taxonomy}
\label{sec3.2: taxonomy_terravis}

To systematically evaluate world consistency, TerraVis organizes violations into three hierarchical levels: object, interaction, and scene, covering seven, eight, and three specific violation types, respectively.
\figref{fig.2} illustrates representative examples of each type.
Detailed definitions and criteria for each violation type are provided in \apxref{apx: violation_criteria}.

\subsection{TerraVis Scoring Framework}
\label{sec3.3: framework_terravis}

Existing text-to-image evaluation dimensions, such as aesthetics~\cite{rahf}, fidelity~\cite{fid, otani2023toward}, alignment~\cite{imagereward, otani2023toward, tifa, vqascore}, and human preference~\cite{hps, hpsv2, hpsv3}, are commonly formulated as holistic or aggregate image-level assessments.
However, it remains unclear whether evaluating world consistency benefits from a more structured evaluation process.
Unlike text-image alignment, which primarily assesses whether the generated content matches the textual prompt, world consistency concerns the intrinsic plausibility of the depicted world, including object structures, spatial layouts, physical interactions, and whether the scene conforms to real-world constraints.
To investigate this, we conducted a pilot annotation study with human volunteers under our task definition of world consistency.
The study suggests that human judgments naturally follow four recurring stages: eligibility checking, violation detection, severity assessment, and score assignment.

\begin{algorithm}[t]
\caption{TerraVis Scoring Procedure}
\label{alg: terravis}
\KwIn{Input image $I$; hyperparameters $\lambda > 0$, $\alpha \in [0, 1]$}
\KwOut{$\mathrm{N/A}$ or TerraVis score $S$}

\BlankLine
\textbf{Eligibility checking:} determine whether $I$ depicts a macroscopic object or scene\;

\If{$I$ is \textbf{not} eligible}{
    \Return $\mathrm{N/A}$\;
}

Initialize $N_{\mathrm{maj}} \leftarrow 0$ and $N_{\mathrm{min}} \leftarrow 0$\;

\textbf{Violation detection:} evaluate $I$ using the TerraVis violation taxonomy, which consists of 18 violation types: 7 object-level, 8 interaction-level, and 3 scene-level violations\;

Let $\gV_{\mathrm{det}}$ denote the set of detected violations\;

\For{each violation $v \in \gV_{\mathrm{det}}$}{
    \textbf{Severity classification}: determine whether $v$ affects a semantically central element in $I$\;
    
    \eIf{$v$ affects a semantically central element}{
        $N_{\mathrm{maj}} \leftarrow N_{\mathrm{maj}} + 1$\;
    }{
        $N_{\mathrm{min}} \leftarrow N_{\mathrm{min}} + 1$\;
    }
}

\textbf{Calculate the TerraVis score}:
\begin{equation*} 
    S = \exp\!\left[-\lambda \left(N_{\mathrm{maj}} + \alpha N_{\mathrm{min}}\right)\right] \nonumber
\end{equation*}

\Return $S \in (0,1]$\;
\end{algorithm}

Hence, TerraVis formalizes this process into a four-step scoring framework, as shown in \algref{alg: terravis}.
Specifically, TerraVis is model-agnostic by design: eligibility checking, violation detection, and severity classification can be carried out by human annotators or instantiated with a multimodal large language model (MLLM), while the final score is computed deterministically from the detected violations and their severities.
The following paragraphs describe each step in detail.

\paratitle{Step 1: Eligibility checking.}
Given an input image $I$, TerraVis first determines whether the image falls within the evaluation scope.
Specifically, the image is considered eligible only if it depicts a macroscopic object or scene with an intended real-world referent.
Images that do not satisfy this condition, such as purely abstract, symbolic, informational, interface-based, layout-based, or microscopic content, are excluded from evaluation and assigned $\mathrm{N/A}$, following the scope definition in \secref{sec3.1: wc}.
This eligibility check ensures that TerraVis is applied only to images for which world consistency is meaningful.
Eligible images are then passed to the violation-detection stage.

\paratitle{Step 2: Violation detection.} 
For each eligible image $I$, TerraVis evaluates whether it exhibits world consistency violations according to the TerraVis violation taxonomy defined in \secref{sec3.2: taxonomy_terravis}.
The taxonomy comprises 18 violation types, including 7 object-level violations, 8 interaction-level violations, and 3 scene-level violations.
Each violation type is checked independently based on visible evidence in the image.
The output of this stage is the detected violation set $\gV_{\mathrm{det}}$, where each element $v \in \gV_{\mathrm{det}}$ corresponds to a violation type observed in $I$.
If no violation is detected, $\gV_{\mathrm{det}} = \varnothing$.

\paratitle{Step 3: Violation severity classification when applicable.}
If one or more violations are detected, i.e., $\gV_{\mathrm{det}} \neq \varnothing$, TerraVis classifies the severity of each violation $v \in \gV_{\mathrm{det}}$ based on the semantic centrality of the affected element.
A violation is categorized as major if it affects a primary subject, a central object, or an element essential to the image content.
In contrast, it is categorized as minor if it affects only a peripheral, background, or otherwise semantically non-central element.
Accordingly, TerraVis records the numbers of major and minor violations as $N_{\mathrm{maj}}$ and $N_{\mathrm{min}}$, respectively.
This distinction allows the scoring procedure to penalize severe errors more strongly while accounting for less salient inconsistencies.

\paratitle{Step 4: Score calculation.}
Finally, TerraVis computes a continuous score $S$ based on the numbers of detected major and minor violations:
\begin{equation} \label{eq.1}
    S = \exp[-\lambda (N_{\mathrm{maj}} + \alpha N_{\mathrm{min}})],
\end{equation}
where $\lambda > 0$ controls the overall penalty strength and $\alpha \in [0,1]$ controls the relative contribution of minor violations compared with major violations.
Thus, higher TerraVis scores indicate stronger world consistency, with $S=1$ reserved for eligible images in which no violation is detected.

This scoring design applies a strong penalty to the first few violations, with diminishing marginal decreases as additional violations accumulate. 
The rationale is that even a small but salient world inconsistency can substantially undermine an image’s usability and credibility, regardless of its aesthetic quality or prompt alignment. 
Moreover, the exponential form helps reduce score compression among high-performing images by clearly separating violation-free outputs from those with visible world-consistency errors, instead of assigning a uniform linear cost to each additional violation.

\subsection{MLLM-based Implementation and Annotation Design}
\label{sec3.4: implementation_terravis}

\paratitle{MLLM workflows.}
We leverage MLLMs to instantiate eligibility checking, violation detection, and severity classification as a sequence of targeted visual question-answering steps. 
Rather than prompting MLLMs to directly assign a holistic score of world-consistency, TerraVis decomposes the evaluation into targeted atomic questions at each stage, thereby constraining the output space and reducing open-ended model hallucination.
Prompting details for each judgment stage are provided in \apxref{apx: prompts}.
This structured execution provides interpretable evidence for each decision and improves alignment with holistic human scores, suggesting that the decomposition captures the visual cues humans rely on when judging world consistency. 

In this work, we instantiate violation detection with two MLLM-based variants: a holistic variant, \textbf{TerraVis-H}, and a workflow-based variant, \textbf{TerraVis}. 
TerraVis-H directly prompts the MLLM with the same taxonomy descriptions and evaluation instructions provided to human annotators, without decomposing the evaluation process. 
In contrast, TerraVis uses the MLLM to execute the sequential workflow defined in \algref{alg: terravis}, including eligibility checking, taxonomy-guided violation detection, severity classification, and severity-aware exponential score aggregation.

\paratitle{Human annotation design.}
To evaluate whether TerraVis aligns with human judgments, we collect human reference scores using Amazon Mechanical Turk (AMT) and measure their correlation with TerraVis scores.
Following prior studies~\cite{rahf, otani2023toward, tifa, vqascore}, annotators are asked to rate each image on a 5-point Likert scale, with an additional N/A option, based on the presence and severity of perceptible world-consistency violations.
Each image is annotated by three independent annotators.
The annotation details are provided in \apxref{apx: human_annotation}.

\section{Experiment Results}
\label{sec4: experiments}

\subsection{Experiment Setup}
\label{sec4.1: exp_settings}

\paratitle{Text-to-image benchmarks.}
Although image-level world-consistency evaluation is inherently prompt-independent, comprehensive benchmarking still requires diverse generated samples to ensure broad scenario coverage. 
We therefore use prompts from two widely used text-to-image benchmarks, COCO-T2I~\cite{coco-t2i} with 200 prompts and GenAI-Bench~\cite{genai-bench} with 1,600 prompts, to evaluate TerraVis alongside existing metrics on world consistency.

\paratitle{Text-to-image models.}
For each sampled text prompt, we generate an image using five popular text-to-image models, including SD 3.5-Large~\cite{sd_v3}, Qwen-Image~\cite{qwen-image}, FLUX.2-dev~\cite{flux2}, GPT Image 1.5~\cite{gptimage1.5}, and Nano Banana Pro~\cite{gemini}.
All images are generated at a resolution of 
$1024 \times 1024$ pixels and each model uses its 
default sampling configuration.

\paratitle{Baseline metrics.}
We compare TerraVis with eight representative text-to-image evaluation metrics:
MUSIQ, LAION-Aesthetic, CLIPScore, TIFA, VQAScore, ImageReward, RAHF, and HPSv3.
For reproducibility, 
MUSIQ is evaluated with the KonIQ checkpoint~\cite{musiq}; LAION-Aesthetic uses Aesthetic Predictor V2.5~\cite{aesthetic_v25}, which consists of a pretrained SigLIP image encoder~\cite{siglip} and the released MLP regression head; 
CLIPScore~\cite{clipscore} uses the SigLIP2 checkpoint~\cite{siglip2}; 
for TIFA~\cite{tifa}, we use Gemma 4 31B~\cite{gemma4} to generate QA pairs from prompts and perform visual question answering;
and VQAScore~\cite{vqascore} computes the ``yes''-token probability using Qwen-3.5~\cite{qwen3.5}.
ImageReward~\cite{imagereward}, RAHF~\cite{rahf} and HPSv3~\cite{hpsv3} are evaluated using their official released checkpoints.
We also include GPTScore as an MLLM-based baseline, where the MLLM is given only the generated image and asked: ``\texttt{Please rate the world consistency of this image using an integer score from 1 to 5, where 1 indicates severe world-consistency errors and 5 indicates no noticeable world-consistency errors.}'' 
These baselines span image quality, alignment, human preference, and holistic MLLM judgment, enabling us to test whether existing metrics can reliably capture world-consistency errors.

\begin{table}[t]
    \centering
    \caption{
    \textbf{Correlation with human judgments of world consistency across evaluation metrics.}
    We report Spearman's $\rho$ and Kendall's $\tau$ between automatic metrics and human ratings on two text-to-image benchmarks.
    Model-wise correlations are computed for each text-to-image model, while \textbf{pooled} correlations aggregate images from all evaluated models within each benchmark.
    Higher values indicate stronger agreement with human judgments.
    TerraVis-H is a holistic taxonomy-informed baseline that uses the full TerraVis taxonomy in a single prompt, while TerraVis decomposes the evaluation into violation identification and severity assessment.
    ``--'' indicates that the metric produced constant predictions, making rank correlations undefined.
    }
    \label{tab.2}
    \setlength{\tabcolsep}{1.6pt}
    \resizebox{\linewidth}{!}{%
    \begin{tabular}{lcccccccccccc}
    \toprule
    \multirow{3}{*}{Metric}
    & \multicolumn{6}{c}{Open-source models}
    & \multicolumn{4}{c}{Proprietary models}
    & \multicolumn{2}{c}{\cellcolor{blue!8}\textbf{Pooled}} \\
    \cmidrule(lr){2-7}
    \cmidrule(lr){8-11}
    \cmidrule(lr){12-13}
    & \multicolumn{2}{c}{SD 3.5-Large~\cite{sd_v3}}
    & \multicolumn{2}{c}{Qwen-Image~\cite{qwen-image}}
    & \multicolumn{2}{c}{FLUX.2-dev~\cite{flux2}}
    & \multicolumn{2}{c}{GPT Image 1.5~\cite{gptimage1.5}}
    & \multicolumn{2}{c}{Nano Banana Pro~\cite{gemini}}
    & \cellcolor{blue!8}$\rho \uparrow$ & \cellcolor{blue!8}$\tau \uparrow$ \\
    \cmidrule(lr){2-3}
    \cmidrule(lr){4-5}
    \cmidrule(lr){6-7}
    \cmidrule(lr){8-9}
    \cmidrule(lr){10-11}
    \cmidrule(lr){12-13}
    & $\rho \uparrow$ & $\tau \uparrow$
    & $\rho \uparrow$ & $\tau \uparrow$
    & $\rho \uparrow$ & $\tau \uparrow$
    & $\rho \uparrow$ & $\tau \uparrow$
    & $\rho \uparrow$ & $\tau \uparrow$
    & \cellcolor{blue!8}$\rho \uparrow$ & \cellcolor{blue!8}$\tau \uparrow$ \\

    \midrule
    \multicolumn{13}{l}{\textbf{Benchmark: COCO-T2I}~\cite{coco-t2i}} \\
    \midrule

    \multicolumn{13}{l}{\textit{Image quality metrics}} \\
    
    MUSIQ~\cite{musiq}
    & 0.07 & 0.05 & 0.24 & 0.17 & 0.01 & 0.01 & 0.06 & 0.04 & 0.06 & 0.05
    & \cellcolor{blue!5}0.12 & \cellcolor{blue!5}0.09 \\

    LAION-Aes~\cite{aesthetic_v25}
    & 0.18 & 0.13 & 0.08 & 0.06 & 0.09 & 0.06 & 0.30 & 0.23 & 0.14 & 0.12
    & \cellcolor{blue!5}0.21 & \cellcolor{blue!5}0.15 \\

    \midrule
    \multicolumn{13}{l}{\textit{Text-image alignment metrics}} \\
    
    CLIPScore~\cite{clipscore}
    & -0.09 & -0.06 & 0.16 & 0.11 & 0.07 & 0.05 & 0.08 & 0.06 & 0.21 & 0.15
    & \cellcolor{blue!5}0.06 & \cellcolor{blue!5}0.04 \\

    TIFA~\cite{tifa}
    & -0.01 & -0.01 & 0.17 & 0.14 & 0.06 & 0.05 & 0.17 & 0.14 & 0.11 & 0.09
    & \cellcolor{blue!5}0.16 & \cellcolor{blue!5}0.13 \\

    VQAScore~\cite{vqascore}
    & 0.12 & 0.09 & 0.33 & 0.25 & 0.27 & 0.19 & 0.25 & 0.18 & 0.05 & 0.03
    & \cellcolor{blue!5}0.26 & \cellcolor{blue!5}0.19 \\

    \midrule
    \multicolumn{13}{l}{\textit{Human preference metrics}} \\
    
    ImageReward~\cite{imagereward}
    & -0.04 & -0.03 & 0.22 & 0.15 & 0.07 & 0.05 & 0.04 & 0.02 & 0.14 & 0.09
    & \cellcolor{blue!5}0.18 & \cellcolor{blue!5}0.12 \\

    RAHF~\cite{rahf}
    & 0.54 & 0.39 & 0.54 & 0.40 & 0.29 & 0.21 & 0.27 & 0.19 & 0.33 & 0.24
    & \cellcolor{blue!5}0.29 & \cellcolor{blue!5}0.21 \\

    HPSv3~\cite{hpsv3}
    & 0.29 & 0.22 & 0.25 & 0.18 & 0.11 & 0.08 & 0.03 & 0.02 & -0.04 & -0.02
    & \cellcolor{blue!5}0.28 & \cellcolor{blue!5}0.20 \\

    \midrule
    \multicolumn{13}{l}{\textit{World-consistency metrics}} \\
    
    GPTScore
    & -- & -- & -- & -- & -- & -- & -- & -- & -- & --
    & -- & -- \\
    
    \rowcolor{green!10}
    TerraVis-H
    & 0.44 & 0.37 & 0.30 & 0.26 & 0.09 & 0.08 & 0.10 & 0.09 & 0.23 & 0.20
    & \cellcolor{blue!12}0.28 & \cellcolor{blue!12}0.24 \\

    \rowcolor{green!10}
    TerraVis
    & 0.43 & 0.34 & 0.55 & 0.46 & 0.22 & 0.19 & 0.33 & 0.29 & 0.36 & 0.31
    & \cellcolor{blue!12}\textbf{0.44} & \cellcolor{blue!12}\textbf{0.37} \\

    \midrule
    \multicolumn{13}{l}{\textbf{Benchmark: GenAI-Bench}~\cite{genai-bench}} \\
    \midrule

    \multicolumn{13}{l}{\textit{Image quality metrics}} \\
    
    MUSIQ~\cite{musiq}
    & -0.12 & -0.08 & 0.02 & 0.01 & -0.13 & -0.10 & 0.01 & 0.01 & 0.06 & 0.05
    & \cellcolor{blue!5}-0.02 & \cellcolor{blue!5}-0.02 \\

    LAION-Aes~\cite{aesthetic_v25}
    & -0.01 & -0.01 & -0.03 & -0.02 & -0.08 & -0.06 & -0.01 & -0.01 & 0.14 & 0.12
    & \cellcolor{blue!5}0.02 & \cellcolor{blue!5}0.01 \\

    \midrule
    \multicolumn{13}{l}{\textit{Text-image alignment metrics}} \\
    
    CLIPScore~\cite{clipscore}
    & -0.15 & -0.11 & -0.18 & -0.13 & -0.19 & -0.13 & -0.26 & -0.18 & 0.21 & 0.15
    & \cellcolor{blue!5}-0.16 & \cellcolor{blue!5}-0.11 \\

    TIFA~\cite{tifa}
    & -0.01 & -0.01 & -0.02 & -0.01 & 0.02 & 0.02 & -0.04 & -0.03 & -0.01 & -0.01
    & \cellcolor{blue!5}0.06 & \cellcolor{blue!5}0.05 \\
    
    VQAScore~\cite{vqascore}
    & 0.02 & 0.02 & 0.04 & 0.03 & 0.06 & 0.04 & -0.02 & -0.01 & 0.05 & 0.03
    & \cellcolor{blue!5}0.12 & \cellcolor{blue!5}0.08 \\

    \midrule
    \multicolumn{13}{l}{\textit{Human preference metrics}} \\
    
    ImageReward~\cite{imagereward}
    & -0.12 & -0.08 & -0.07 & -0.05 & -0.18 & -0.13 & -0.20 & -0.14 & 0.14 & 0.09
    & \cellcolor{blue!5}-0.08 & \cellcolor{blue!5}-0.06 \\

    RAHF~\cite{rahf}
    & 0.36 & 0.26 & 0.45 & 0.32 & 0.45 & 0.33 & 0.35 & 0.25 & 0.36 & 0.26
    & \cellcolor{blue!5}0.37 & \cellcolor{blue!5}0.27 \\

    HPSv3~\cite{hpsv3}
    & 0.10 & 0.07 & -0.03 & -0.02 & -0.05 & -0.03 & -0.03 & -0.02 & -0.04 & -0.02
    & \cellcolor{blue!5}0.10 & \cellcolor{blue!5}0.07 \\

    \midrule
    \multicolumn{13}{l}{\textit{World-consistency metrics}} \\
    GPTScore
    & 0.04 & 0.03 & 0.04 & 0.03 & -- & -- & -- & -- & -- & --
    & 0.03 & 0.03 \\
    \rowcolor{green!10}
    TerraVis-H
    & 0.13 & 0.11 & 0.23 & 0.19 & 0.35 & 0.30 & 0.37 & 0.31 & 0.23 & 0.20
    & \cellcolor{blue!12}0.28 & \cellcolor{blue!12}0.23 \\

    \rowcolor{green!10}
    TerraVis
    & 0.33 & 0.26 & 0.37 & 0.30 & 0.43 & 0.36 & 0.48 & 0.40 & 0.37 & 0.30
    & \cellcolor{blue!12}\textbf{0.43} & \cellcolor{blue!12}\textbf{0.35} \\

    \bottomrule
    \end{tabular}}
    \vspace{-4mm}
\end{table}

\paratitle{Implementation details.}
We exclude generated images that are not applicable (i.e., N/A images) to world-consistency evaluation, where applicability is determined from the generated image itself rather than inferred from the text-guided prompt. 
These cases are rare: on average across the evaluated text-to-image models, less than 1\% of GenAI-Bench samples are out of scope for world-consistency assessment. 
All experiments are conducted using the GPT-5.5 API~\cite{gpt5} with default generation parameters.
To improve efficiency, we use the Batch API to parallelize all stage-wise QA calls.
We preserve each text-to-image model's stylistic choices under the given prompt, so the evaluated images include both photorealistic outputs and a small number of stylized generations.
For human annotation, we annotate a randomly selected 50\% subset of the generated images and use this subset for correlation analysis between automatic metrics and human judgments.
We set $\lambda$ and $\alpha$ in \eqref{eq.1} as 1 and 0.5, respectively.

\begin{table}[t]
    \centering
    \caption{
        Correlations with human judgments using different MLLM evaluators.
        Each cell reports Spearman's $\rho$ / Kendall's $\tau$;
        best results for each benchmark and metric are bolded.
    }
    \label{tab.3}
    \small
    \setlength{\tabcolsep}{3pt}
    \renewcommand{\arraystretch}{0.95}
    \begin{tabular*}{\linewidth}{@{\extracolsep{\fill}}lccccc@{}}
        \toprule
        Benchmark
        & \makecell{Gemma 4\\31B}
        & \makecell{Qwen3-VL\\32B-Instruct}
        & \makecell{Gemma 4\\26B-A4B}
        & \makecell{Qwen3-VL\\30B-A3B-Instruct}
        & GPT-5.5 \\
        \midrule
        COCO-T2I
        & 0.42 / 0.32
        & 0.30 / 0.25
        & 0.37 / 0.29
        & 0.27 / 0.23
        & \textbf{0.44} / \textbf{0.37} \\
        GenAI-Bench
        & \textbf{0.51} / \textbf{0.38}
        & 0.36 / 0.29
        & 0.42 / 0.33
        & 0.35 / 0.28
        & 0.43 / 0.35 \\
        \bottomrule
    \end{tabular*}
    \vspace{-3mm}
\end{table}

\begin{figure}[t]
    \centering
    \begin{subfigure}{0.48\linewidth}
        \centering
        \includegraphics[width=\linewidth]{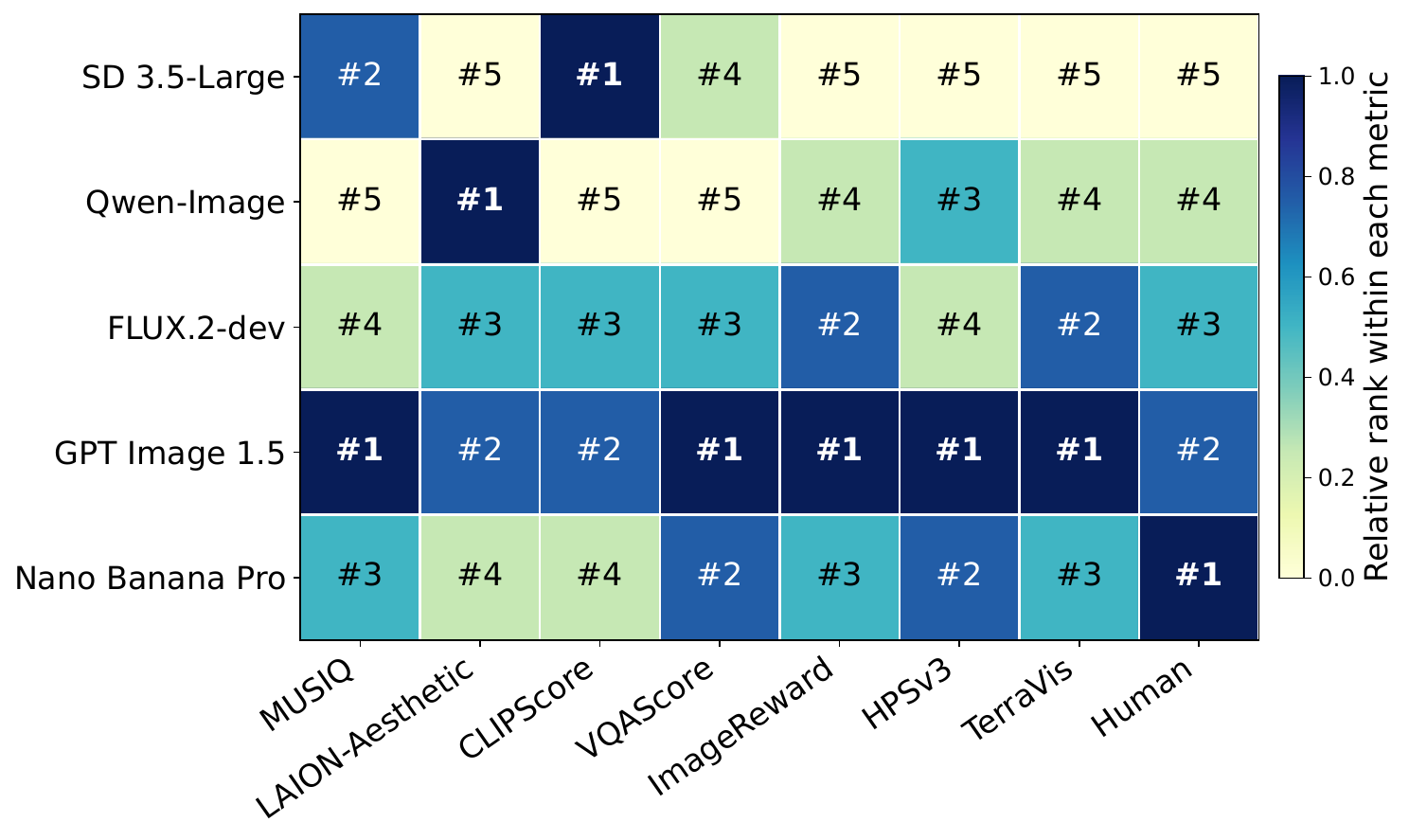}
        \caption{COCO-T2I}
    \end{subfigure}
    \hfill
    \begin{subfigure}{0.48\linewidth}
        \centering
        \includegraphics[width=\linewidth]{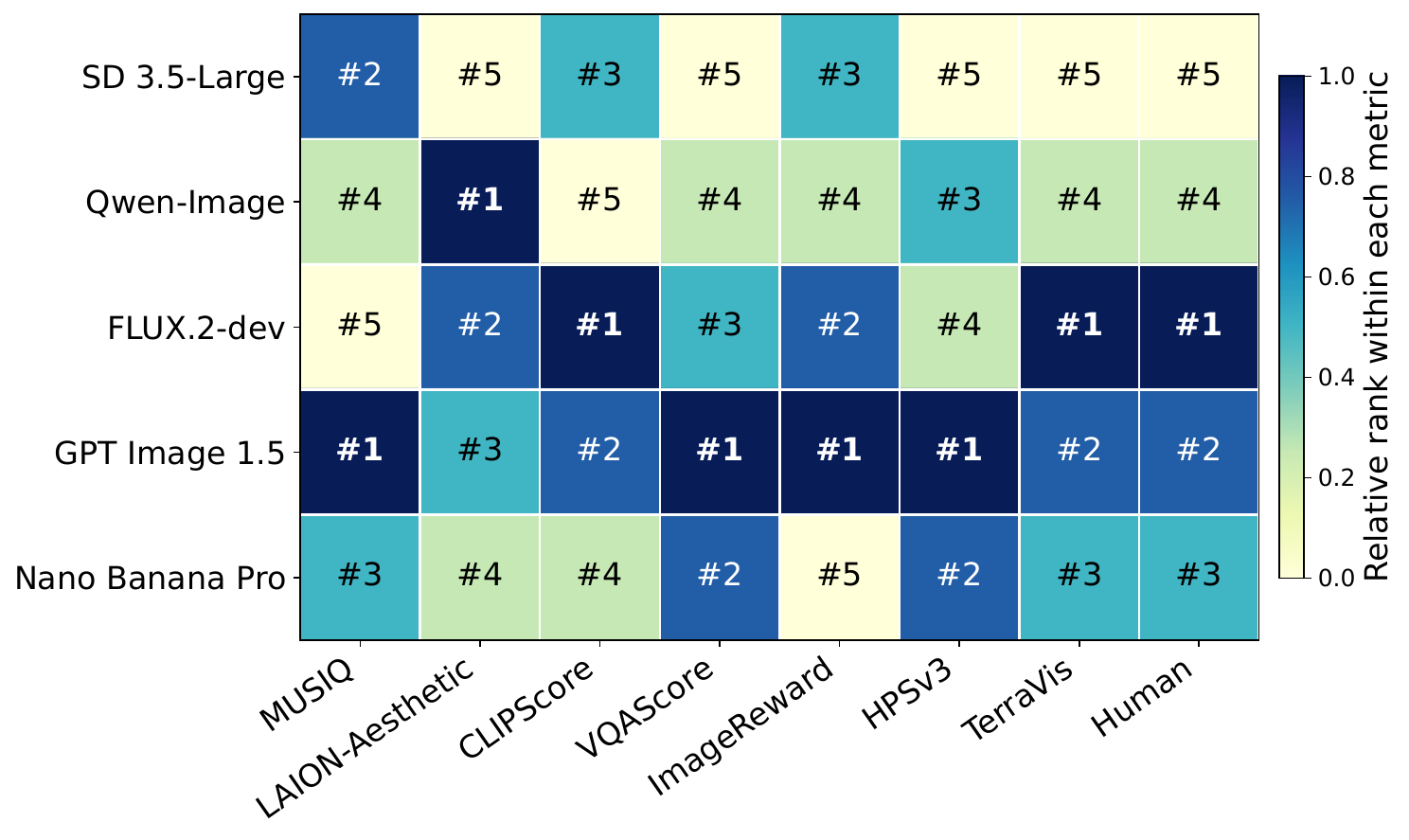}
        \caption{GenAI-Bench}
    \end{subfigure}
    \caption{
    Rank heatmaps comparing T2I models across multiple evaluation metrics on two benchmarks.
    Human denotes the ranking derived from human judgments of world consistency.
    Lower ranks (with darker shades) indicate better performance.
    }
    \label{fig.3}
    \vspace{-4mm}
\end{figure}

\begin{table}[t]
    \centering
    \caption{Ablation studies of TerraVis.
    (a) Per-level results are aggregated across both benchmarks.
    \textsc{Only} reports Spearman's $\rho$ with human ratings using that level alone;
    \textsc{LOO} means change in $\rho$ when that level is removed.
    (b) Aggregation rules are evaluated using Pearson's $r$,
    Spearman's $\rho$, and Kendall's $\tau$ against human judgments.
    The last three rows in (b) use $\lambda=1$ and $\alpha=0.5$.}
    \label{tab.4}
    \small
    \setlength{\tabcolsep}{5pt}
    \renewcommand{\arraystretch}{1.15}
    \begin{subtable}[t]{0.44\linewidth}
        \centering
        \caption{Evaluation levels.}
        \label{tab.4.a}
        \begin{tabularx}{\linewidth}{@{}Xccc@{}}
            \toprule
            Level & \#Types & \textsc{Only}: $\rho$ & \textsc{LOO}: $\Delta\rho$ \\
            \midrule
            Object      & 7  & 0.384 & \textminus 0.149 \\
            Interaction & 8  & 0.253 & \textminus 0.033 \\
            Scene       & 3  & 0.174 & \textminus 0.016 \\
            \midrule
            \textbf{Full} & 18 & \textbf{0.428} & -- \\
            \bottomrule
        \end{tabularx}
    \end{subtable}\hfill
    \begin{subtable}[t]{0.46\linewidth}
        \centering
        \caption{Aggregation rules.}
        \label{tab.4.b}
        \begin{tabularx}{\linewidth}{@{}Xccc@{}}
            \toprule
            Aggregation & $r$ & $\rho$ & $\tau$ \\
            \midrule
            Uniform pass-rate  & 0.3634 & 0.4264 & 0.3471 \\
            Uniform exp.       & 0.4212 & 0.4264 & 0.3471 \\
            Severity pass-rate & 0.3630 & \textbf{0.4281} & \textbf{0.3477} \\
            \midrule
            \textbf{TerraVis exp. (ours)} & \textbf{0.4229} & \textbf{0.4281} & \textbf{0.3477} \\
            \bottomrule
        \end{tabularx}
    \end{subtable}
    \vspace{-5mm}
\end{table}

\subsection{Correlation with Human Judgments}
\label{sec4.2: exp_correlation}

\paratitle{Existing metrics are not reliable indicators of world consistency.}
\tabref{tab.2} reports the correlations between automatic metrics and human judgments of world consistency.
Existing metrics show highly model-dependent agreement with human judgments, as illustrated by the near-zero or negative correlations for TIFA on SD 3.5-Large and HPSv3 on Nano Banana Pro across both benchmarks.
GPTScore also exhibits near-zero correlations, assigning almost uniformly high scores to generated images.
Among existing metrics, RAHF achieves the strongest correlations ($\rho=0.29$ on COCO-T2I and $0.37$ on GenAI-Bench).
This may reflect its plausibility dimension, which partially overlaps with TerraVis in detecting violations such as structural distortions.
However, TerraVis outperforms all existing metrics by large margins on both benchmarks.
These findings suggest that existing quality, alignment, and preference metrics provide only partial and inconsistent signals of world consistency.

\paratitle{Structured world-consistency evaluation improves agreement with humans.}
As shown in \tabref{tab.2}, TerraVis consistently achieves stronger correlations with human judgments than TerraVis-H.
For example, on COCO-T2I, the correlation improves from TerraVis-H ($\rho=0.28$, $\tau=0.24$) to TerraVis ($\rho=0.44$, $\tau=0.37$).
Violation detection analysis provides further insight into this improvement: TerraVis-H exhibits high precision but low recall, whereas TerraVis substantially improves recall while maintaining high precision (see \apxref{apx: exp_holistic_vs_structured} for detailed results).
Together, these findings suggest that decomposing evaluation into eligibility checking, taxonomy-guided violation detection, severity classification, and severity-aware aggregation provides a more reliable approximation of human judgments than a holistic MLLM assessment.

\paratitle{Can TerraVis evaluate stylized images?}
Yes.
To examine TerraVis's applicability to stylized imagery, we use SigLIP-so400m~\cite{siglip} to partition the evaluation images into photorealistic and stylized subsets.
We exclude images assigned N/A by either TerraVis or human annotators.
TerraVis achieves a higher Spearman correlation with human judgments on stylized images ($\rho=0.539$) than on photorealistic images ($\rho=0.413$).
These results suggest that TerraVis can not only assess world-consistency violations on photorealistic images but also stylized ones.

\paratitle{How robust is TerraVis to the choice of MLLM evaluators?}
TerraVis maintains comparable correlations with human judgments and largely consistent T2I model rankings when GPT-5.5 is replaced with Gemma 4 31B.
To assess TerraVis's robustness to evaluator choice, we rerun it with four additional open-source MLLMs spanning dense and MoE architectures, assessing both correlations with human judgments (\tabref{tab.3}) and agreement among the resulting T2I model rankings (\tabref{tab: evaluator_ranking_agreement}).
Gemma 4 31B achieves a Spearman correlation comparable to that of GPT-5.5 on COCO-T2I ($\rho=0.42$ vs. $0.44$) and a higher correlation on GenAI-Bench ($\rho=0.51$ vs. $0.43$).
For T2I model rankings, the mean pairwise Kendall's $\tau$ across all five evaluators is $0.64$ on COCO-T2I and $0.68$ on GenAI-Bench.
However, GPT-5.5 and Gemma 4 31B show strong ranking agreement ($\tau=0.80$ on both benchmarks), while the remaining three evaluators also agree closely with one another (mean pairwise $\tau=0.87$).
These results indicate stronger ranking agreement within the two evaluator groups than between them.
Together, the correlation and ranking results support \textbf{Gemma 4 31B as an effective open-source alternative} to GPT-5.5 for TerraVis evaluation.

\subsection{Multi-Dimensional Evaluation of T2I Models}
\label{sec4.3: exp_model_comparison}

We benchmark five text-to-image models (three open-source and two proprietary) across technical quality, aesthetics, text-image alignment, human preference, and world consistency.
As shown in \figref{fig.3}, GPT Image 1.5 ranks first in 9 of 16 comparisons across both benchmarks, including 8 of 12 on conventional metrics.
On COCO-T2I, GPT Image 1.5 achieves the best mean rank across all eight metrics (1.38), followed by Nano Banana Pro (2.75), FLUX.2-dev (3.00), Qwen-Image (3.88), and SD 3.5-Large (4.00).
Notably, FLUX.2-dev ranks first in world consistency on GenAI-Bench according to both TerraVis and human judgments, ahead of both proprietary models.
A simple qualitative inspection suggests that FLUX.2-dev often generates simpler scenes than the proprietary models, potentially reducing the risk of world-consistency violations.
These findings suggest that visual richness and strong conventional performance do not necessarily imply better world consistency, highlighting its value as a complementary evaluation dimension.

\subsection{Ablation Study}
\label{sec4.4: exp_ablation}

\begin{figure}[t]
    \centering
    \casepanel{(a) Human Oversight}
    {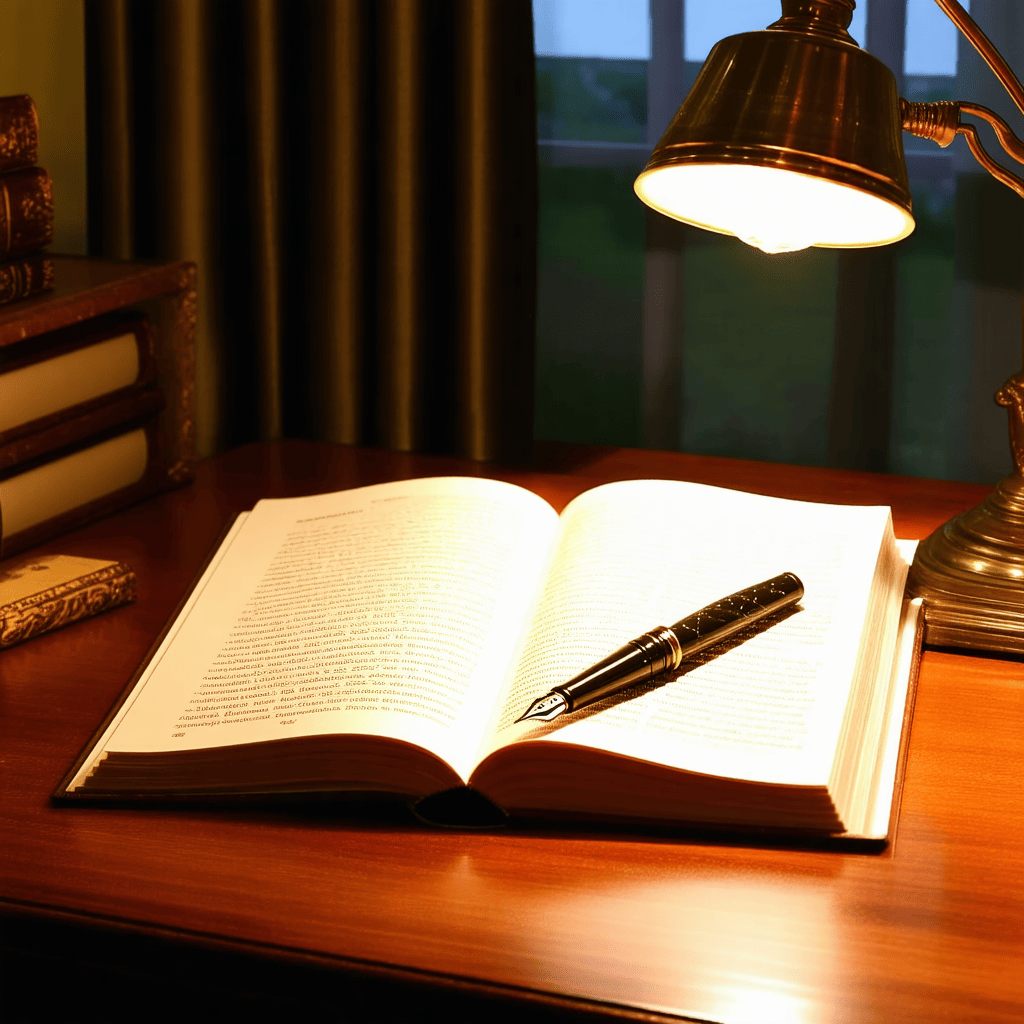}{0.135}{1.0\,(3,5,5)}{SD 3.5-Large / GenAI-Bench-0347}{
    \textbf{Symbolic content}\par
    {\footnotesize\color{darkgray}Major $\cdot$ object-level}\par\vspace{3pt}
    \textit{Affected aspect:} open book text. \\
    \textit{Evidence:} the printed text on the open pages appears as tiny, malformed, nonsensical glyph-like lines rather than plausible readable writing.\par\vspace{3pt}
    {\footnotesize\textbf{Other detected types:} optical effect.}
    }
    \hfill
    \casepanel{(b) False Positives}
    {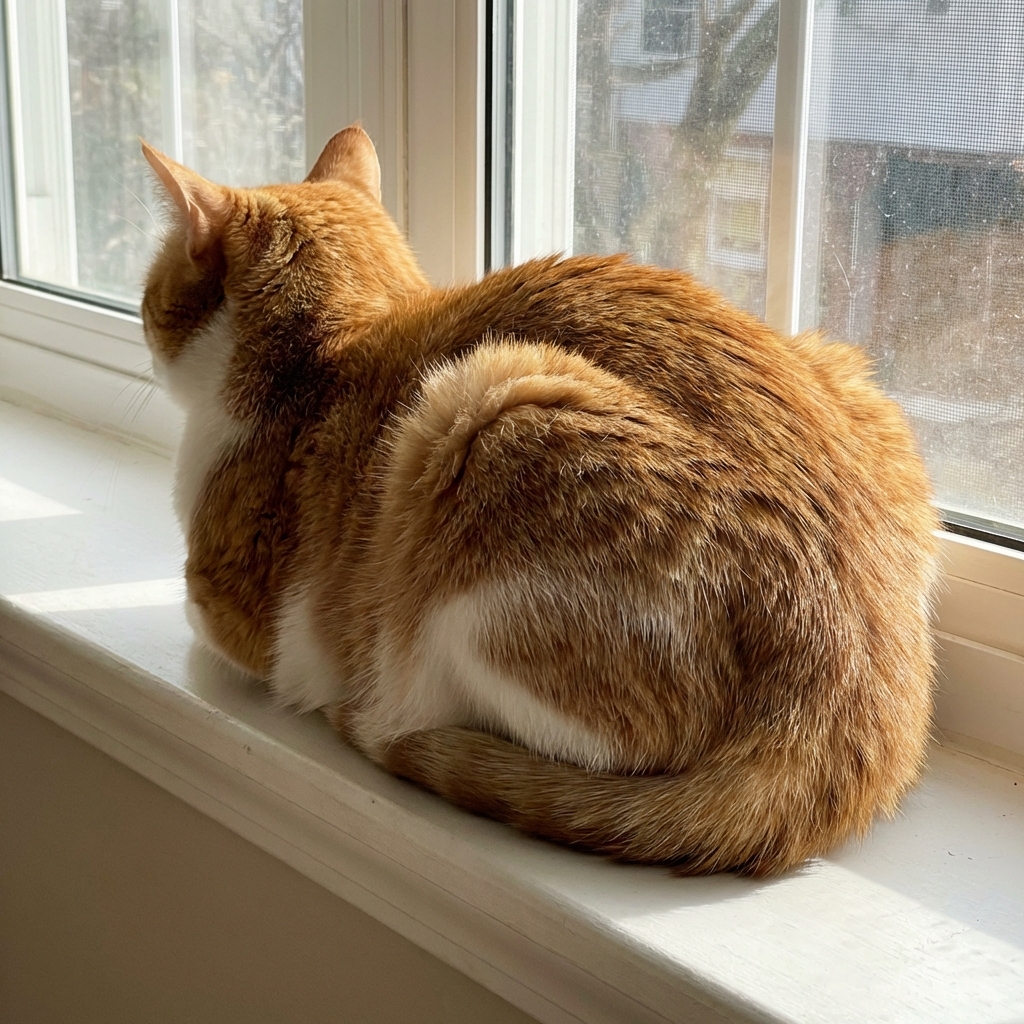}{0.050}{1.0\,(5,5,5)}{Nano Banana Pro / GenAI-Bench-0296}{
    \textbf{Object identity}\par
    {\footnotesize\color{darkgray}Major $\cdot$ object-level}\par\vspace{3pt}
    \textit{Affected aspect:} cat. \\
    \textit{Evidence:} the cat has a large circular coiled furry protrusion on its back/flank that resembles an extra curled tail or shell, which is not normal real cat anatomy.\par\vspace{3pt}
    {\footnotesize\textbf{Other detected types:} structural distortion, biological anatomy.}
    }
    \hfill
    \casepanel{(c) Missed Violations}
    {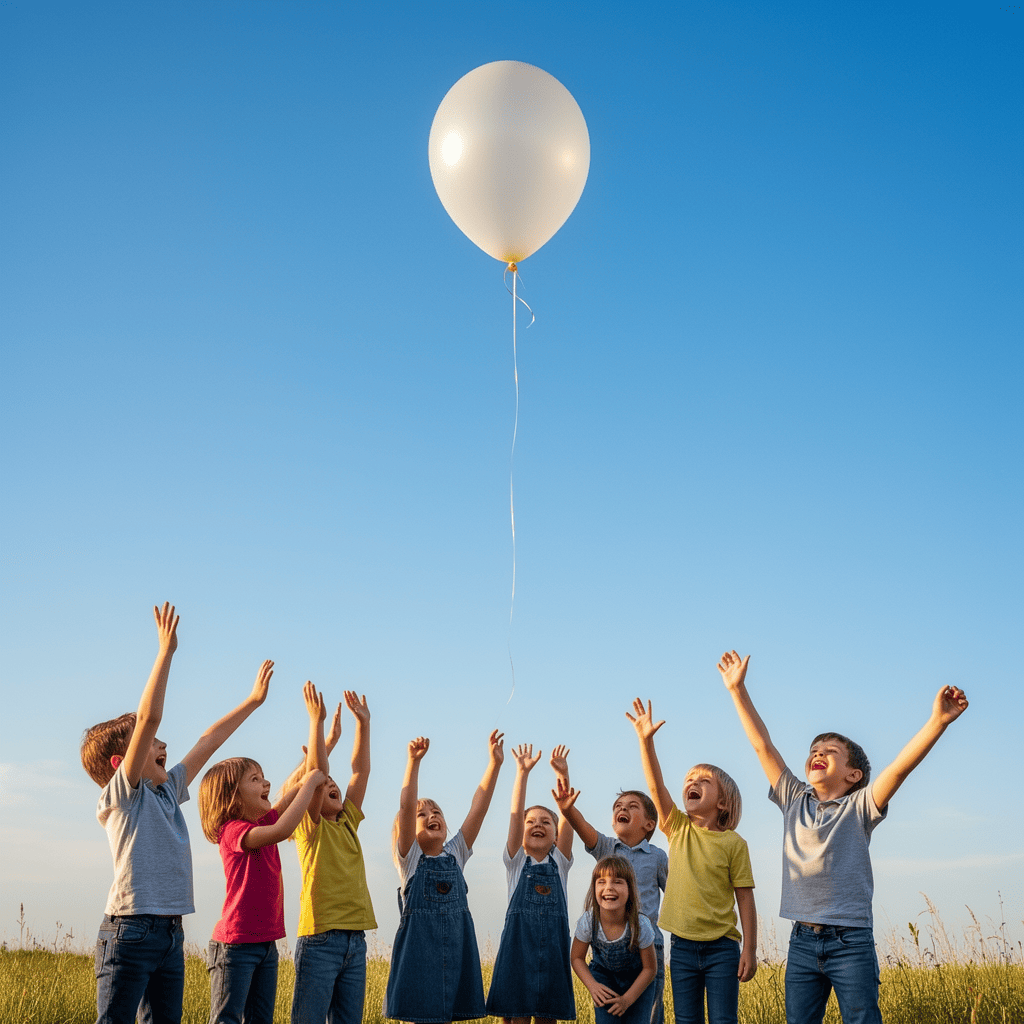}{1.000}{0.0\,(1,1,1)}{FLUX.2-dev / GenAI-Bench-0354}{
    \textcolor{TV}{\textbf{No violation detected}}\par\vspace{4pt}
    \textbf{Disagreement:} TerraVis detects no violations and assigns the maximum world-consistency score of 1.0, whereas all three human raters assign the lowest rating (1/5). Our manual review confirms anatomical distortions in the children's hands and arms that TerraVis fails to detect.
    }
    \hfill
    \casepanel{(d) Rater Disagreement}
    {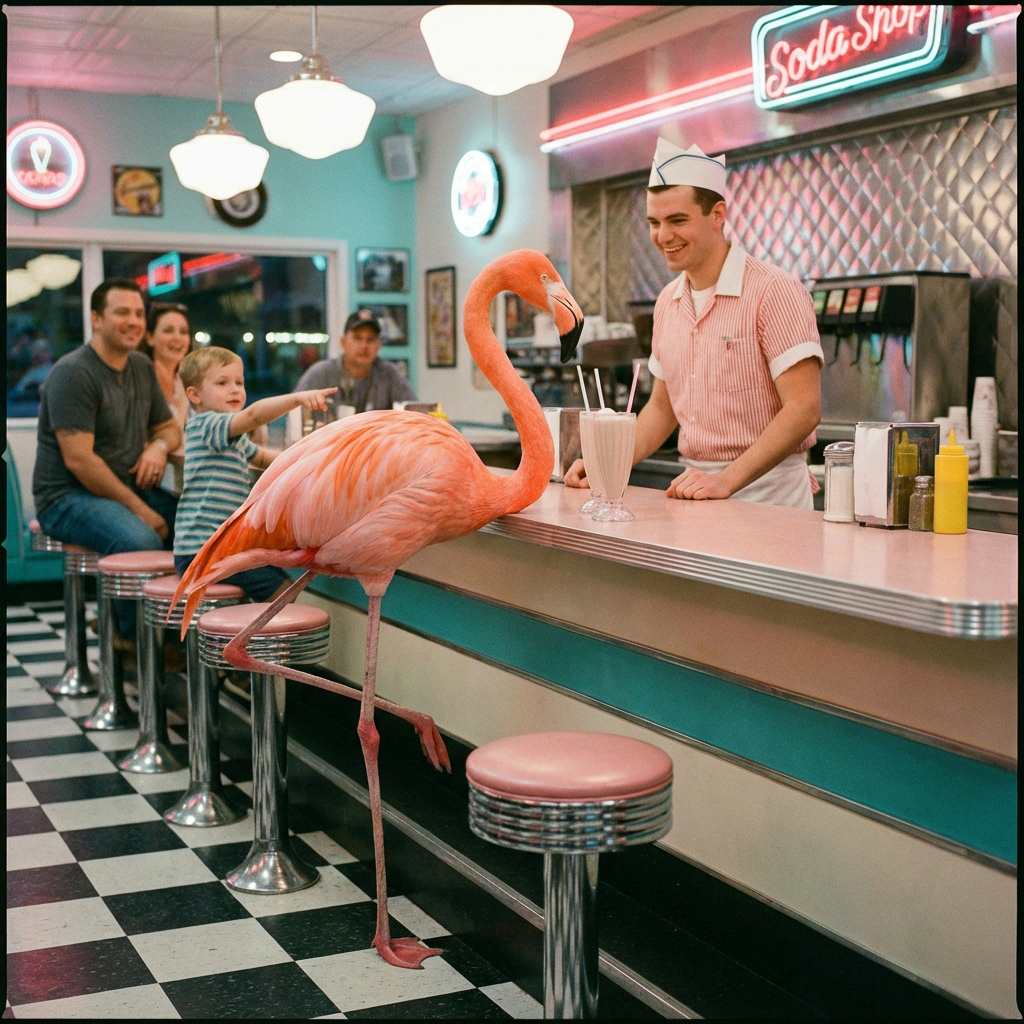}{0.011}{0.5\,(3,5,1)}{Nano Banana Pro / GenAI-Bench-0584}{
    \textcolor{Human}{\textbf{Human judgments}}\par\vspace{4pt}
    \textbf{Rater 1 (3/5):} Partially consistent.\par\vspace{3pt}
    \textbf{Rater 2 (5/5):} Fully consistent.\par\vspace{3pt}
    \textbf{Rater 3 (1/5):} Completely inconsistent.\par\vspace{4pt}
    \textbf{Disagreement:} Ratings span the full 1--5 scale for the same image, indicating substantial inter-rater disagreement.
    }
    \caption{
        \textbf{Four patterns of disagreement in world-consistency evaluation.}
        (a) TerraVis detects genuine violations overlooked by all three human raters. 
        (b) TerraVis produces likely false-positive detections in an image that all three human raters judge fully consistent. 
        (c) TerraVis fails to detect anatomical distortions in the children’s hands and arms that human raters identify and our manual review confirms. 
        (d) Human raters assign substantially different scores to the same image.
        The violation descriptions in (a) and (b) are reproduced verbatim from TerraVis outputs.
    }
    \label{fig.4}
    \vspace{-4mm}
\end{figure}

\paragraph{Ablation on evaluation levels.}
\tabref{tab.4.a} shows that removing any evaluation level reduces correlation on both benchmarks, with the full framework consistently performing best.
The object level contributes most, followed by the interaction level.
The scene level yields the smallest marginal gain, which may reflect its smaller set of violation types (3, compared with 7 for objects and 8 for interactions) and the tendency of scene-level violations, such as implausible relative scale, to co-occur with those at other levels.
Nevertheless, its standalone Spearman correlation ($\rho=0.174$, aggregated over both benchmarks) suggests that it captures genuine and useful signal.
An ablation study of individual violation types is presented in \apxref{apx: exp_ablation_types}.

\paratitle{Ablation of aggregation rules.}
\tabref{tab.4.b} compares four rules for aggregating the same TerraVis detections into evaluation scores.
The \emph{Uniform pass-rate} baseline treats each detected violation as a failed check regardless of severity, yielding $S=1-(N_{\mathrm{maj}}+N_{\mathrm{min}})/M$.
\emph{Uniform exp.} uses the same unweighted violation count but replaces the linear score with an exponential penalty, $S=\exp[-\lambda(N_{\mathrm{maj}}+N_{\mathrm{min}})]$.
\emph{Severity pass-rate} retains the linear formulation while down-weighting minor violations by a factor of $\alpha$, giving $S=1-(N_{\mathrm{maj}}+\alpha N_{\mathrm{min}})/M$.
Our default \emph{TerraVis exp.} combines severity weighting with exponential aggregation, $S=\exp[-\lambda(N_{\mathrm{maj}}+\alpha N_{\mathrm{min}})]$.
These variants allow us to assess the effects of severity weighting and exponential aggregation separately.
A sensitivity analysis of the aggregation hyperparameters is provided in \apxref{apx: sensitivity_lamda_alpha}.

\subsection{Case Studies}
\label{sec4.5: case_studies}

To complement the quantitative results, we present a qualitative analysis of representative cases in \figref{fig.4}, which illustrates four patterns of disagreement in world-consistency evaluation: human oversight, false positives, missed violations, and rater disagreement. 
TerraVis can effectively detect textual and symbolic violations overlooked by humans, but may also produce false positives or miss violations in dense scenes. 
Human raters may also disagree on the presence and severity of violations in unusual scenes. 
\apxref{apx: case_studies} provides additional examples involving stylized images and qualitative comparisons between human world-consistency ratings and scores from existing metrics, highlighting their limitations in capturing world-consistency violations.

\subsection{Discussion and Limitations}
\label{sec4.6: discussion}

Although TerraVis achieves stronger correlations with human judgments of world consistency than existing automatic metrics, several limitations remain.
\textbf{First}, its multi-stage workflow makes TerraVis less efficient than single-pass scoring metrics (see \apxref{apx: exp_efficiency}).
\textbf{Second}, violation detection performance remains a potential bottleneck.
As discussed in \secref{sec4.5: case_studies}, TerraVis may produce false positives or miss genuine violations.
However, our preliminary results suggest that in-context learning with representative violation examples can improve detection performance, supporting data-centric refinement as a promising direction for future work.
\textbf{Third}, although the taxonomy and prompting strategies were developed through iterative refinement combining principled design with data-driven validation, the taxonomy may not cover all forms of world-consistency violations in practice.
Continued refinement of the taxonomy and associated prompts may therefore be needed as T2I models evolve.
\textbf{Finally}, the benchmark relies on human annotations as its reference.
Judgments of ambiguous cases and violation severity may vary across annotators, introducing uncertainty into the measured agreement between automatic metrics and human judgments.

\section{Conclusion}
\label{sec5: conclusion}

In this work, we introduce world consistency as a new perspective for text-to-image evaluation.
We propose TerraVis, a taxonomy-guided MLLM workflow that performs eligibility checking, violation detection, severity classification, and severity-aware aggregation.
To validate TerraVis, we collect over 13K human annotations across two benchmarks and five text-to-image models.
Experiments on COCO-T2I and GenAI-Bench show that TerraVis achieves the strongest agreement with human judgments, outperforming existing quality, alignment, and preference metrics.
This work demonstrates the need to evaluate not only visual appeal or text-image alignment, but also whether generated images visually conform to real-world constraints.
We hope this work helps establish world consistency as a core evaluation dimension for future text-to-image models.



\bibliography{references}

@String(CVPR  = {CVPR})

@String(ICCV  = {ICCV})

@String(ECCV  = {ECCV})

@String(ICML  = {ICML})

@String(ICLR  = {ICLR})

@String(EMNLP= {EMNLP})

@String(ACL= {ACL})

@article{ddpm,
  title={Denoising diffusion probabilistic models},
  author={Ho, Jonathan and Jain, Ajay and Abbeel, Pieter},
  journal={Advances in Neural Information Processing Systems},
  volume={33},
  pages={6840--6851},
  year={2020}
}

@inproceedings{ldm,
  title={High-resolution image synthesis with latent diffusion models},
  author={Rombach, Robin and Blattmann, Andreas and Lorenz, Dominik and Esser, Patrick and Ommer, Bj{\"o}rn},
  booktitle=CVPR,
  pages={10684--10695},
  year={2022}
}

@article{adm,
  title={Diffusion Models Beat GANs on Image Synthesis},
  author={Dhariwal, Prafulla and Nichol, Alexander},
  journal={Advances in Neural Information Processing Systems},
  volume={34},
  pages={8780--8794},
  year={2021}
}

@article{edm,
  title={Elucidating the Design Space of Diffusion-based Generative Models},
  author={Karras, Tero and Aittala, Miika and Aila, Timo and Laine, Samuli},
  journal={Advances in Neural Information Processing Systems},
  volume={35},
  pages={26565--26577},
  year={2022}
}

@inproceedings{clipscore,
  title={CLIPScore: A Reference-free Evaluation Metric for Image Captioning},
  author={Hessel, Jack and Holtzman, Ari and Forbes, Maxwell and Le Bras, Ronan and Choi, Yejin},
  booktitle=EMNLP,
  pages={7514--7528},
  year={2021}
}

@article{fid,
  title={Gans trained by a two time-scale update rule converge to a local nash equilibrium},
  author={Heusel, Martin and Ramsauer, Hubert and Unterthiner, Thomas and Nessler, Bernhard and Hochreiter, Sepp},
  journal={Advances in Neural Information Processing Systems},
  volume={30},
  year={2017}
}

@article{precision_recall,
  title={Assessing generative models via precision and recall},
  author={Sajjadi, Mehdi SM and Bachem, Olivier and Lucic, Mario and Bousquet, Olivier and Gelly, Sylvain},
  journal={Advances in Neural Information Processing Systems},
  volume={31},
  year={2018}
}

@inproceedings{sd_v3,
  title={Scaling rectified flow transformers for high-resolution image synthesis},
  author={Esser, Patrick and Kulal, Sumith and Blattmann, Andreas and Entezari, Rahim and M{\"u}ller, Jonas and Saini, Harry and Levi, Yam and Lorenz, Dominik and Sauer, Axel and Boesel, Frederic and others},
  booktitle=ICML,
  pages={12606-12633},
  year={2024}
}

@inproceedings{vqascore,
  title={Evaluating text-to-visual generation with image-to-text generation},
  author={Lin, Zhiqiu and Pathak, Deepak and Li, Baiqi and Li, Jiayao and Xia, Xide and Neubig, Graham and Zhang, Pengchuan and Ramanan, Deva},
  booktitle=ECCV,
  pages={366--384},
  year={2024},
}

@article{llmscore,
  title={Llmscore: Unveiling the power of large language models in text-to-image synthesis evaluation},
  author={Lu, Yujie and Yang, Xianjun and Li, Xiujun and Wang, Xin Eric and Wang, William Yang},
  journal={Advances in Neural Information Processing Systems},
  volume={36},
  pages={23075--23093},
  year={2023}
}

@inproceedings{tifa,
  title={Tifa: Accurate and interpretable text-to-image faithfulness evaluation with question answering},
  author={Hu, Yushi and Liu, Benlin and Kasai, Jungo and Wang, Yizhong and Ostendorf, Mari and Krishna, Ranjay and Smith, Noah A},
  booktitle=ICCV,
  pages={20406--20417},
  year={2023}
}

@article{vpeval,
  title={Visual programming for step-by-step text-to-image generation and evaluation},
  author={Cho, Jaemin and Zala, Abhay and Bansal, Mohit},
  journal={Advances in Neural Information Processing Systems},
  volume={36},
  pages={6048--6069},
  year={2023}
}

@inproceedings{viescore,
  title={Viescore: Towards explainable metrics for conditional image synthesis evaluation},
  author={Ku, Max and Jiang, Dongfu and Wei, Cong and Yue, Xiang and Chen, Wenhu},
  booktitle=ACL,
  pages={12268--12290},
  year={2024}
}

@inproceedings{dsg,
  title={Davidsonian Scene Graph: Improving Reliability in Fine-grained Evaluation for Text-to-Image Generation},
  author={Cho, Jaemin and Hu, Yushi and Baldridge, Jason Michael and Garg, Roopal and Anderson, Peter and Krishna, Ranjay and Bansal, Mohit and Pont-Tuset, Jordi and Wang, Su},
  booktitle=ICLR,
  pages={1--21},
  year={2024}
}

@article{imagereward,
  title={Imagereward: Learning and evaluating human preferences for text-to-image generation},
  author={Xu, Jiazheng and Liu, Xiao and Wu, Yuchen and Tong, Yuxuan and Li, Qinkai and Ding, Ming and Tang, Jie and Dong, Yuxiao},
  journal={Advances in Neural Information Processing Systems},
  volume={36},
  pages={15903--15935},
  year={2023}
}

@article{pickscore,
  title={Pick-a-pic: An open dataset of user preferences for text-to-image generation},
  author={Kirstain, Yuval and Polyak, Adam and Singer, Uriel and Matiana, Shahbuland and Penna, Joe and Levy, Omer},
  journal={Advances in Neural Information Processing Systems},
  volume={36},
  pages={36652--36663},
  year={2023}
}

@inproceedings{hps,
  title={Human preference score: Better aligning text-to-image models with human preference},
  author={Wu, Xiaoshi and Sun, Keqiang and Zhu, Feng and Zhao, Rui and Li, Hongsheng},
  booktitle=ICCV,
  pages={2096--2105},
  year={2023}
}

@article{hpsv2,
  title={Human preference score v2: A solid benchmark for evaluating human preferences of text-to-image synthesis},
  author={Wu, Xiaoshi and Hao, Yiming and Sun, Keqiang and Chen, Yixiong and Zhu, Feng and Zhao, Rui and Li, Hongsheng},
  journal={arXiv preprint arXiv:2306.09341},
  year={2023}
}

@inproceedings{mps,
  title={Learning multi-dimensional human preference for text-to-image generation},
  author={Zhang, Sixian and Wang, Bohan and Wu, Junqiang and Li, Yan and Gao, Tingting and Zhang, Di and Wang, Zhongyuan},
  booktitle=CVPR,
  pages={8018--8027},
  year={2024}
}

@inproceedings{hpsv3,
  title={Hpsv3: Towards wide-spectrum human preference score},
  author={Ma, Yuhang and Wu, Xiaoshi and Sun, Keqiang and Li, Hongsheng},
  booktitle=ICCV,
  pages={15086--15095},
  year={2025}
}

@inproceedings{ava,
  title={AVA: A large-scale database for aesthetic visual analysis},
  author={Murray, Naila and Marchesotti, Luca and Perronnin, Florent},
  booktitle=CVPR,
  pages={2408--2415},
  year={2012}
}

@article{nima,
  title={NIMA: Neural image assessment},
  author={Talebi, Hossein and Milanfar, Peyman},
  journal={IEEE Transactions on Image Processing},
  volume={27},
  number={8},
  pages={3998--4011},
  year={2018}
}

@misc{aesthetic_v25,
  title        = {Aesthetic Predictor V2.5},
  author       = {Verb, discus0434@gmail.com},
  year         = {2024},
  howpublished = {\url{https://github.com/discus0434/aesthetic-predictor-v2-5}},
  note         = {Version 2024.12.18.1,  GitHub repository}
}

@article{niqe,
  title={Making a “completely blind” image quality analyzer},
  author={Mittal, Anish and Soundararajan, Rajiv and Bovik, Alan C},
  journal={IEEE Signal Processing Letters},
  volume={20},
  number={3},
  pages={209--212},
  year={2012}
}

@article{brisque,
  title={No-reference image quality assessment in the spatial domain},
  author={Mittal, Anish and Moorthy, Anush Krishna and Bovik, Alan Conrad},
  journal={IEEE Transactions on Image Processing},
  volume={21},
  number={12},
  pages={4695--4708},
  year={2012}
}

@inproceedings{musiq,
  title={Musiq: Multi-scale image quality transformer},
  author={Ke, Junjie and Wang, Qifei and Wang, Yilin and Milanfar, Peyman and Yang, Feng},
  booktitle=ICCV,
  pages={5148--5157},
  year={2021}
}

@misc{flux2,
  author={Black Forest Labs},
  title={{FLUX.2: Frontier Visual Intelligence}},
  year={2025},
  howpublished={\url{https://bfl.ai/blog/flux-2}},
}

@misc{gptimage1.5,
  author={OpenAI},
  title={{GPT Image 1.5}},
  year={2025},
  howpublished = {\url{https://developers.openai.com/api/docs/models/gpt-image-1.5}}
}

@article{gemini,
  title={Gemini: a family of highly capable multimodal models},
  author={Team, Gemini and Anil, Rohan and Borgeaud, Sebastian and Alayrac, Jean-Baptiste and Yu, Jiahui and Soricut, Radu and Schalkwyk, Johan and Dai, Andrew M and Hauth, Anja and Millican, Katie and others},
  journal={arXiv preprint arXiv:2312.11805},
  year={2023}
}

@article{qwen-image,
  title={Qwen-image technical report},
  author={Wu, Chenfei and Li, Jiahao and Zhou, Jingren and Lin, Junyang and Gao, Kaiyuan and Yan, Kun and Yin, Sheng-ming and Bai, Shuai and Xu, Xiao and Chen, Yilei and others},
  journal={arXiv preprint arXiv:2508.02324},
  year={2025}
}

@article{gan,
  title={Generative adversarial nets},
  author={Goodfellow, Ian J and Pouget-Abadie, Jean and Mirza, Mehdi and Xu, Bing and Warde-Farley, David and Ozair, Sherjil and Courville, Aaron and Bengio, Yoshua},
  journal={Advances in Neural Information Processing Systems},
  volume={27},
  year={2014}
}

@inproceedings{wgan,
  title={Wasserstein generative adversarial networks},
  author={Arjovsky, Martin and Chintala, Soumith and Bottou, L{\'e}on},
  booktitle=ICML,
  pages={214--223},
  year={2017}
}

@inproceedings{cyclegan,
  title={Unpaired image-to-image translation using cycle-consistent adversarial networks},
  author={Zhu, Jun-Yan and Park, Taesung and Isola, Phillip and Efros, Alexei A},
  booktitle=ICCV,
  pages={2223--2232},
  year={2017}
}

@inproceedings{stylegan,
  title={A style-based generator architecture for generative adversarial networks},
  author={Karras, Tero and Laine, Samuli and Aila, Timo},
  booktitle=CVPR,
  pages={4401--4410},
  year={2019}
}

@inproceedings{biggan,
  title={Large Scale GAN Training for High Fidelity Natural Image Synthesis},
  author={Brock, Andrew and Donahue, Jeff and Simonyan, Karen},
  booktitle=ICLR,
  pages={1--35},
  year={2019}
}

@article{counthallu,
  title={Counting Hallucinations in Diffusion Models},
  author={Fu, Shuai and Zhou, Jian and Chen, Qi and Jing, Huang and Nguyen, Huy Anh and Liu, Xiaohan and Zeng, Zhixiong and Ma, Lin and Zhang, Quanshi and Wu, Qi},
  journal={arXiv preprint arXiv:2510.13080},
  year={2025}
}

@inproceedings{dalle,
  title={Zero-shot text-to-image generation},
  author={Ramesh, Aditya and Pavlov, Mikhail and Goh, Gabriel and Gray, Scott and Voss, Chelsea and Radford, Alec and Chen, Mark and Sutskever, Ilya},
  booktitle=ICML,
  pages={8821--8831},
  year={2021}
}

@inproceedings{glide,
  title={GLIDE: Towards Photorealistic Image Generation and Editing with Text-Guided Diffusion Models},
  author={Nichol, Alexander Quinn and Dhariwal, Prafulla and Ramesh, Aditya and Shyam, Pranav and Mishkin, Pamela and Mcgrew, Bob and Sutskever, Ilya and Chen, Mark},
  booktitle=ICML,
  pages={16784--16804},
  year={2022},
}

@article{imagen,
  title={Photorealistic text-to-image diffusion models with deep language understanding},
  author={Saharia, Chitwan and Chan, William and Saxena, Saurabh and Li, Lala and Whang, Jay and Denton, Emily L and Ghasemipour, Kamyar and Gontijo Lopes, Raphael and Karagol Ayan, Burcu and Salimans, Tim and others},
  journal={Advances in Neural Information Processing Systems},
  volume={35},
  pages={36479--36494},
  year={2022}
}

@article{dalle-2,
  title={Hierarchical Text-Conditional Image Generation with CLIP Latents},
  author={Ramesh, Aditya and Dhariwal, Prafulla and Nichol, Alex and Chu, Casey and Chen, Mark},
  journal={arXiv preprint arXiv:2204.06125},
  year={2022}
}

@inproceedings{stackgan,
  title={Stackgan: Text to photo-realistic image synthesis with stacked generative adversarial networks},
  author={Zhang, Han and Xu, Tao and Li, Hongsheng and Zhang, Shaoting and Wang, Xiaogang and Huang, Xiaolei and Metaxas, Dimitris N},
  booktitle=ICCV,
  pages={5907--5915},
  year={2017}
}

@inproceedings{srgan,
  title={Photo-realistic single image super-resolution using a generative adversarial network},
  author={Ledig, Christian and Theis, Lucas and Husz{\'a}r, Ferenc and Caballero, Jose and Cunningham, Andrew and Acosta, Alejandro and Aitken, Andrew and Tejani, Alykhan and Totz, Johannes and Wang, Zehan and others},
  booktitle=CVPR,
  pages={4681--4690},
  year={2017}
}

@article{parti,
  title={Scaling autoregressive models for content-rich text-to-image generation},
  author={Yu, Jiahui and Xu, Yuanzhong and Koh, Jing Yu and Luong, Thang and Baid, Gunjan and Wang, Zirui and Vasudevan, Vijay and Ku, Alexander and Yang, Yinfei and Ayan, Burcu Karagol and others},
  journal={arXiv preprint arXiv:2206.10789},
  year={2022}
}

@article{emu3,
  title={Emu3: Next-token prediction is all you need},
  author={Wang, Xinlong and Zhang, Xiaosong and Luo, Zhengxiong and Sun, Quan and Cui, Yufeng and Wang, Jinsheng and Zhang, Fan and Wang, Yueze and Li, Zhen and Yu, Qiying and others},
  journal={arXiv preprint arXiv:2409.18869},
  year={2024}
}

@article{var,
  title={Visual autoregressive modeling: Scalable image generation via next-scale prediction},
  author={Tian, Keyu and Jiang, Yi and Yuan, Zehuan and Peng, Bingyue and Wang, Liwei},
  journal={Advances in Neural Information Processing Systems},
  volume={37},
  pages={84839--84865},
  year={2024}
}

@inproceedings{janus,
  title={Janus: Decoupling visual encoding for unified multimodal understanding and generation},
  author={Wu, Chengyue and Chen, Xiaokang and Wu, Zhiyu and Ma, Yiyang and Liu, Xingchao and Pan, Zizheng and Liu, Wen and Xie, Zhenda and Yu, Xingkai and Ruan, Chong and others},
  booktitle=CVPR,
  pages={12966--12977},
  year={2025}
}

@inproceedings{infinity,
  title={Infinity: Scaling bitwise autoregressive modeling for high-resolution image synthesis},
  author={Han, Jian and Liu, Jinlai and Jiang, Yi and Yan, Bin and Zhang, Yuqi and Yuan, Zehuan and Peng, Bingyue and Liu, Xiaobing},
  booktitle=CVPR,
  pages={15733--15744},
  year={2025}
}

@inproceedings{transfusion,
  title={Transfusion: Predict the Next Token and Diffuse Images with One Multi-Modal Model},
  author={Zhou, Chunting and YU, LILI and Babu, Arun and Tirumala, Kushal and Yasunaga, Michihiro and Shamis, Leonid and Kahn, Jacob and Ma, Xuezhe and Zettlemoyer, Luke and Levy, Omer},
  booktitle=ICLR,
  year={2025},
  pages={1--23},
}

@inproceedings{hart,
  title={HART: Efficient Visual Generation with Hybrid Autoregressive Transformer},
  author={Tang, Haotian and Wu, Yecheng and Yang, Shang and Xie, Enze and Chen, Junsong and Chen, Junyu and Zhang, Zhuoyang and Cai, Han and Lu, Yao and Han, Song},
  booktitle=ICLR,
  year={2025},
  pages={1--20},
}

@article{hunyuanimage,
  title={Hunyuanimage 3.0 technical report},
  author={Cao, Siyu and Chen, Hangting and Chen, Peng and Cheng, Yiji and Cui, Yutao and Deng, Xinchi and Dong, Ying and Gong, Kipper and Gu, Tianpeng and Gu, Xiusen and others},
  journal={arXiv preprint arXiv:2509.23951},
  year={2025}
}

@inproceedings{otani2023toward,
  title={Toward verifiable and reproducible human evaluation for text-to-image generation},
  author={Otani, Mayu and Togashi, Riku and Sawai, Yu and Ishigami, Ryosuke and Nakashima, Yuta and Rahtu, Esa and Heikkil{\"a}, Janne and Satoh, Shin’ichi},
  booktitle=CVPR,
  pages={14277--14286},
  year={2023}
}

@inproceedings{rahf,
  title={Rich human feedback for text-to-image generation},
  author={Liang, Youwei and He, Junfeng and Li, Gang and Li, Peizhao and Klimovskiy, Arseniy and Carolan, Nicholas and Sun, Jiao and Pont-Tuset, Jordi and Young, Sarah and Yang, Feng and others},
  booktitle=CVPR,
  pages={19401--19411},
  year={2024}
}

@inproceedings{kid,
  title={Demystifying MMD GANs},
  author={Bi{\'n}kowski, Miko{\l}aj and Sutherland, Danica J and Arbel, Michael and Gretton, Arthur},
  booktitle=ICLR,
  year={2018},
  pages={1--36},
}

@article{bagel,
  title={Emerging properties in unified multimodal pretraining},
  author={Deng, Chaorui and Zhu, Deyao and Li, Kunchang and Gou, Chenhui and Li, Feng and Wang, Zeyu and Zhong, Shu and Yu, Weihao and Nie, Xiaonan and Song, Ziang and others},
  journal={arXiv preprint arXiv:2505.14683},
  year={2025}
}

@article{coco-t2i,
  title={What you see is what you read? improving text-image alignment evaluation},
  author={Yarom, Michal and Bitton, Yonatan and Changpinyo, Soravit and Aharoni, Roee and Herzig, Jonathan and Lang, Oran and Ofek, Eran and Szpektor, Idan},
  journal={Advances in Neural Information Processing Systems},
  volume={36},
  pages={1601--1619},
  year={2023}
}

@article{genai-bench,
  title={Genai-bench: Evaluating and improving compositional text-to-visual generation},
  author={Li, Baiqi and Lin, Zhiqiu and Pathak, Deepak and Li, Jiayao and Fei, Yixin and Wu, Kewen and Ling, Tiffany and Xia, Xide and Zhang, Pengchuan and Neubig, Graham and others},
  journal={arXiv preprint arXiv:2406.13743},
  year={2024}
}

@article{gpt5,
  title={Openai gpt-5 system card},
  author={Singh, Aaditya and Fry, Adam and Perelman, Adam and Tart, Adam and Ganesh, Adi and El-Kishky, Ahmed and McLaughlin, Aidan and Low, Aiden and Ostrow, AJ and Ananthram, Akhila and others},
  journal={arXiv preprint arXiv:2601.03267},
  year={2025}
}

@misc{qwen3.5,
  title  = {{Qwen3.5}: Towards Native Multimodal Agents},
  author = {{Qwen Team}},
  month  = {February},
  year   = {2026},
  url    = {https://qwen.ai/blog?id=qwen3.5}
}

@inproceedings{siglip,
  title={Sigmoid loss for language image pre-training},
  author={Zhai, Xiaohua and Mustafa, Basil and Kolesnikov, Alexander and Beyer, Lucas},
  booktitle=ICCV,
  pages={11975--11986},
  year={2023}
}

@article{siglip2,
  title={Siglip 2: Multilingual vision-language encoders with improved semantic understanding, localization, and dense features},
  author={Tschannen, Michael and Gritsenko, Alexey and Wang, Xiao and Naeem, Muhammad Ferjad and Alabdulmohsin, Ibrahim and Parthasarathy, Nikhil and Evans, Talfan and Beyer, Lucas and Xia, Ye and Mustafa, Basil and others},
  journal={arXiv preprint arXiv:2502.14786},
  year={2025}
}

@inproceedings{mscoco,
  title={Microsoft coco: Common objects in context},
  author={Lin, Tsung-Yi and Maire, Michael and Belongie, Serge and Hays, James and Perona, Pietro and Ramanan, Deva and Doll{\'a}r, Piotr and Zitnick, C Lawrence},
  booktitle=ECCV,
  pages={740--755},
  year={2014},
  organization={Springer}
}

@article{gwetac1,
  title={Computing inter-rater reliability and its variance in the presence of high agreement},
  author={Gwet, Kilem Li},
  journal={British Journal of Mathematical and Statistical Psychology},
  volume={61},
  number={1},
  pages={29--48},
  year={2008}
}

@article{gemma4,
  title={Gemma 4 technical report},
  author={Team, Gemma and Abd, Sherif El and Aggarwal, Vaibhav and Algayres, Robin and Andreev, Alek and Bachem, Olivier and Ballantyne, Ian and Brick, Cormac and C{\u{a}}rbune, Victor and Casbon, Michelle and others},
  journal={arXiv preprint arXiv:2607.02770},
  year={2026}
}

@article{fu2023cooperative,
  title={Cooperative attention generative adversarial network for unsupervised domain adaptation},
  author={Fu, Shuai and Chen, Jing and Lei, Liang},
  journal={Knowledge-Based Systems},
  volume={261},
  pages={110196},
  year={2023}
}
\bibliographystyle{plainnat}
\newpage
\appendix
\appendix
\setcounter{table}{0}
\setcounter{figure}{0}

\renewcommand{\thetable}{A\arabic{table}}
\renewcommand{\thefigure}{A\arabic{figure}}
\section{Detailed Definitions and Criteria for TerraVis Violation Taxonomy}
\label{apx: violation_criteria}

\paratitle{Object-level.}
At the object level, the general standard is that objects and symbolic content should appear valid in the real world and be contextually appropriate for the depicted scene.
This level captures whether individual entities are plausible in themselves, independent of broader scene dynamics or inter-object relationships.
\textit{Specifically, for stylized content, we do not penalize stylistic simplifications in texture/surface appearance, high-frequency detail, or color complexity, as long as the object's identity and basic structure remain recognizable.} 
Under this level, we define seven specific object-level violation types:

\begin{itemize}
    \item \textbf{Object identity violations}: the presence of fictional or fantasy entities that do not exist in the real world (e.g., dragons, genies, elves, or unicorns), as well as cross-species feature mixtures or invalid species traits (e.g., a cat with a dog-like face).
    \item \textbf{Structural distortion}: obvious structural distortions arising from generation failures (e.g., melted-looking objects).
    \item \textbf{Biological anatomy violations}: anatomically impossible features (e.g., a human hand with six fingers or eyes looking in incompatible directions).
    \item \textbf{Non-biological structure violations}: impossible object structures or implausible part configurations (e.g.,  a bicycle with disconnected wheels or a car with misplaced wheels).
    \item \textbf{Texture/surface violations}: physically implausible textures or material patterns (e.g., skin appearing like stone or wood grain).
    \item \textbf{Symbolic-content violations}: broken, distorted, or unrecognizable text, symbols, signs, logos, or other markings.
    \item \textbf{Object-context violations}: object-scene combinations that violate common real-world knowledge (e.g., a dog working in an office), or symbolic content inappropriate to its context (e.g., menu text unrelated to food).
\end{itemize}

\textbf{Interaction-level.}
At the interaction level, the general standard is that objects, agents, and environmental elements should interact in physically plausible ways, with those interactions should be consistent with the depicted action or event.
This level focuses on whether contact, force, motion, medium response, and event-dependent behavior are visually coherent.
\textit{Similarly, for stylized images, we do not penalize simplifications in lighting, motion depiction, or medium-response details, provided they do not break the causal logic of the interaction.} 
We consider eight specific interaction-level violation types:

\begin{itemize}
    \item \textbf{Contact-state violations}: physically impossible contact relationships (i.e., object interpenetration).
    \item \textbf{Support and stability violations}: implausible hovering or levitation without support, or objects appearing stably at rest despite inadequate support, balance, or friction (e.g., a cup perched on a narrow railing without tipping).
    \item \textbf{Dynamic response violations}: missing or implausible physical responses under force (e.g., a car driving through a puddle without splash, or motion cues that contradict the implied acceleration, impact, or turning).
    \item \textbf{Medium interaction violations}: implausible interactions with water, air, snow, sand, or similar media (e.g., incorrect buoyancy, immersion depth, waterline, or air-resistance cues).
    \item \textbf{Optical effect violations}: physically implausible shadows, reflections, or refraction/transmission effects (e.g., a mirror reflection that does not match the reflected object).
    \item \textbf{Energy source violations}: visible light, heat, or other energy output without a plausible source or supporting cue (e.g., a glowing light bulb with no wires, socket, or visible power source).
    \item \textbf{Thermal response violations}: implausible responses to heat or cold (e.g., an ice cube sitting on a red-hot pan without melting).
    \item \textbf{Behavior-event interaction violations}: behaviors that do not match the depicted event (e.g., eyes not tracking the ball during a tennis shot).
\end{itemize}

\textbf{Scene-level.}
At the scene level, the general standard is that the depicted scene should exhibit a coherent global spatial configuration, with scale, depth ordering, and region continuity that are mutually consistent and plausible for a real-world scene.
This level captures failures that may not be attributable to any single object, but instead arise from inconsistencies in the global composition of the image.
We consider three specific scene-level violation types:

\begin{itemize}
    \item \textbf{Relative-scale violations}: implausible relative scale between objects and the surrounding scene, or among objects within the scene, which could be caused by inconsistent depth or perspective cues (e.g., a distant rider appearing as large as a nearby car).
    \item \textbf{Depth-order/occlusion violations}: implausible occlusion relationships or depth-order inconsistencies (e.g., overlapping elephants with incorrect occlusion that make some legs appear missing or duplicated).
    \item \textbf{Region continuity violations}: abrupt or implausible discontinuities between adjacent scene regions (e.g., a desk scene split by an abrupt lighting change).
\end{itemize}

Overall, this three-level formulation allows TerraVis to evaluate world consistency in a structured and interpretable way. 
Instead of treating inconsistency as a simple binary failure signal, TerraVis decomposes it into violations at the object, interaction, and scene levels, covering 7, 8, and 3 specific violation types, respectively. 
This design supports finer-grained analysis of model behavior and provides greater insight into the failure modes underlying world-consistency errors.

\section{Prompts Used in TerraVis}
\label{apx: prompts}

This appendix provides the full  prompts used in the TerraVis evaluation pipeline.
All prompts are fixed across models and samples unless otherwise specified.

\subsection{Eligibility Checking}
\label{apx: prompt_eligibility}

\begin{promptbox}{Eligibility Checking Prompt}
\begin{Verbatim}[
    fontsize=\scriptsize,
    breaklines=true,
    breakanywhere=true,
    breaksymbolleft={},
    breaksymbolright={},
    xleftmargin=0pt,
    baselinestretch=0.95
]
### Task
You are an image eligibility classifier. Determine whether the given image is eligible for world-consistency evaluation.

### Rules

Mark as "Eligible" if the image is presented as a single-frame representational image depicting identifiable macroscopic objects or scenes. Photorealistic and stylized images both count.

Mark as "N/A" if the image's primary content belongs to any of the following categories:
- Abstract content: (e.g., color blocks, geometric patterns, noise textures, abstract shapes, abstract illustrations)
- Symbolic content: (e.g., logos, icons, emblems, flags, emojis, QR codes)
- Informational content: (e.g., charts, diagrams, maps, tables, flowcharts, infographics, timelines, mind maps, menus)
- Interface content: (e.g., UI mockups, app/website screenshots, software interfaces, operating-system windows, video game menus, control panels)
- Layout-based compositions: (e.g., posters, advertisements, book covers, storyboards, collages, multi-panel layouts)
- Microscopic content: (e.g., cell structures, material surfaces, mineral textures, crystal structures)

### Priority Rules
- If the image is clearly composed of multiple independent panels, frames, or views, mark it as "N/A".
- If the image shows a physical object that happens to contain text, a map, a menu, a cover design, or other printed information, mark it as "Eligible" unless the image is only a flat graphic/layout with no physical-object presentation.

### Output Format
Output strictly in the following format, with no additional text:

Answer: <Eligible or N/A>
Reason: <One sentence describing the reason.>
\end{Verbatim}
\end{promptbox}

\subsection{Violation Detection}
\label{apx: prompt_vio_detect}

\begin{promptbox}{Violation Detection Prompt Template}
\begin{Verbatim}[
    fontsize=\scriptsize,
    breaklines=true,
    breakanywhere=true,
    breaksymbolleft={},
    breaksymbolright={},
    xleftmargin=0pt,
    baselinestretch=0.95
]
### Task
You are an image issue detector. Determine whether the following issue is present in the image.

### Rules
- Do not penalize intentional reduced photorealism in stylized images (e.g., simplified texture, color, lighting, motion depiction, medium-response details).
- If the issue is present, report the single most salient instance only.
- Focus only on visible evidence.
- Do not rely on imagination.
- Use a strict standard, as some violations may be subtle.

### Question:
{QUESTION}

### Output Format
Output strictly in the following format, with no additional text:

Answer: <Yes or No>.
Affected aspect: <name of the affected object, interaction, or scene element; or empty if No>
Evidence: <one sentence describing the visible evidence; or empty if No>
\end{Verbatim}
\end{promptbox}
The above template is shared across all violation detections.
For each violation type, we instantiate the placeholder \texttt{\{QUESTION\}} with a violation-specific question derived from the detailed definitions, criteria, and examples in \apxref{apx: violation_criteria}.
We found that directly asking the model whether a named violation category is present yields limited effectiveness, so we formulate each question around the concrete visual conditions to be checked.
This design reduces ambiguity and encourages the model to ground its decision in visible evidence rather than in its interpretation of a taxonomy label.
The full set of violation-specific questions is shown in \tabref{tab: violation_questions}.

\begin{promptbox}{Violation Severity Classification Prompt Template}
\begin{Verbatim}[
    fontsize=\scriptsize,
    breaklines=true,
    breakanywhere=true,
    breaksymbolleft={},
    breaksymbolright={},
    xleftmargin=0pt,
    baselinestretch=0.95
]
### Task
You are an image issue severity classifier. Determine whether the detected issue is a major or minor world-consistency violation.

### Rules
- Major violation: the violation affects the main subject, a foreground object, the depicted action/event, or another semantically central element in the image, even if the visible anomaly is small.
- Minor violation: the violation affects a background, peripheral, or semantically non-central element.
- Use only the detected issue and visible evidence. Do not introduce new issues.

### Detected Issue
Violation type: {VIOLATION_TYPE}
Question: {QUESTION}
Affected aspect: {AFFECTED_ASPECT}
Evidence: {EVIDENCE}

### Output Format
Output strictly in the following format, with no additional text:

Severity: <Major or Minor>.
Reason: <One sentence explaining whether the affected aspect is semantically central or peripheral.>
\end{Verbatim}
\end{promptbox}
This prompt is used for severity classification after a world-consistency violation has been detected. 
Given the violation type, diagnostic question, affected aspect, and visible evidence provided by the previous stage, the classifier determines whether the issue should be treated as a major or minor violation. 
A violation is classified as major if it affects the main subject, foreground object, depicted action, or another semantically central element; otherwise, it is classified as minor when the affected aspect is peripheral or background-related. 
This design separates violation detection from severity assessment and encourages the model to judge severity based only on the detected issue and its visual evidence.

\begin{longtable}{p{0.25\linewidth} p{0.68\linewidth}}
\caption{Violation-specific questions used to instantiate the prompt template in \secref{apx: prompt_vio_detect}.}
\label{tab: violation_questions} \\
\toprule
\textbf{Violation type} & \textbf{Question inserted into \texttt{\{QUESTION\}}} \\
\midrule
\endfirsthead

\toprule
\textbf{Violation type} & \textbf{Question inserted into \texttt{\{QUESTION\}}} \\
\midrule
\endhead

Object identity
& Ignoring embedded depictions such as murals, paintings, screens, or printed images, does the image contain any physically present entity that is presented as a real object or living being but has impossible identity traits, such as a living dragon or an impossible animal/object hybrid? \\

Structural distortion 
& Does any object appear visibly malformed due to generation failure, for example melted, broken in shape, or geometrically distorted in an implausible way? \\

Biological anatomy
& Does any human or animal body show anatomically impossible features, such as an impossible limb orientation, or eyes looking in incompatible directions? \\

Non-biological structure 
& Does any non-living object have an impossible structure or part arrangement, such as disconnected wheels, misplaced parts, physically invalid construction, or parts whose geometry or alignment is inconsistent under the apparent viewpoint? \\

Texture/surface
& Does any object show a physically implausible material or surface pattern, such as skin that looks like wood or stone without scene justification? \\

Symbolic content
& Does any visible text, sign, symbol, logo, or marking appear broken, distorted, or unrecognizable in a way that violates real-world plausibility? \\

Object-context 
& Is there any object or symbolic content that is clearly inappropriate for the scene context
according to common real-world knowledge, such as a dog working in an office? \\

\midrule

Contact state 
& Do any two objects appear to pass through each other or interpenetrate when they should be in normal physical contact? \\

Support and stability 
& Does any person or object appear to float, hover, or remain stably at rest without adequate support, balance, or friction, and without plausible motion context? \\

Dynamic response 
& Is there any missing or implausible physical response to force, motion, impact, or turning, such as no splash, no deformation, or contradictory motion cues? \\

Medium interaction 
& Is there any implausible interaction with water, air, snow, sand, or similar media, such as incorrect buoyancy, immersion depth, wake, or air-resistance cues? \\

Optical effect
& Is there any physically implausible shadow, reflection, refraction, or transmission effect? \\

Energy source 
& Is there visible light, heat, glow, or other energy output without a plausible source or supporting cue? \\

Thermal response 
& Is there any implausible response to heat or cold, such as missing or inappropriate melting, burning, boiling, steam emission, freezing, or condensation? \\

Behavior-event
& Does any person or animal behave in a way that contradicts the depicted event or action? \\

\midrule

Relative scale
& Do objects or subjects appear to have unrealistic relative sizes compared to each other or to the scene? \\

Depth-order/occlusion 
& Are there any implausible occlusion relationships or depth-order inconsistencies, such as overlapping subjects causing parts to appear implausibly missing, duplicated, or incorrectly occluded? \\

Region continuity
& Does the image show unnatural transitions between adjacent regions, such as visible seams, inconsistent lighting across regions, mismatched perspective, or spatial discontinuities at region boundaries? \\

\bottomrule
\end{longtable}


\section{Human Annotation}
\label{apx: human_annotation}

\begin{figure}[t]
    \centering
    \includegraphics[width=\textwidth]{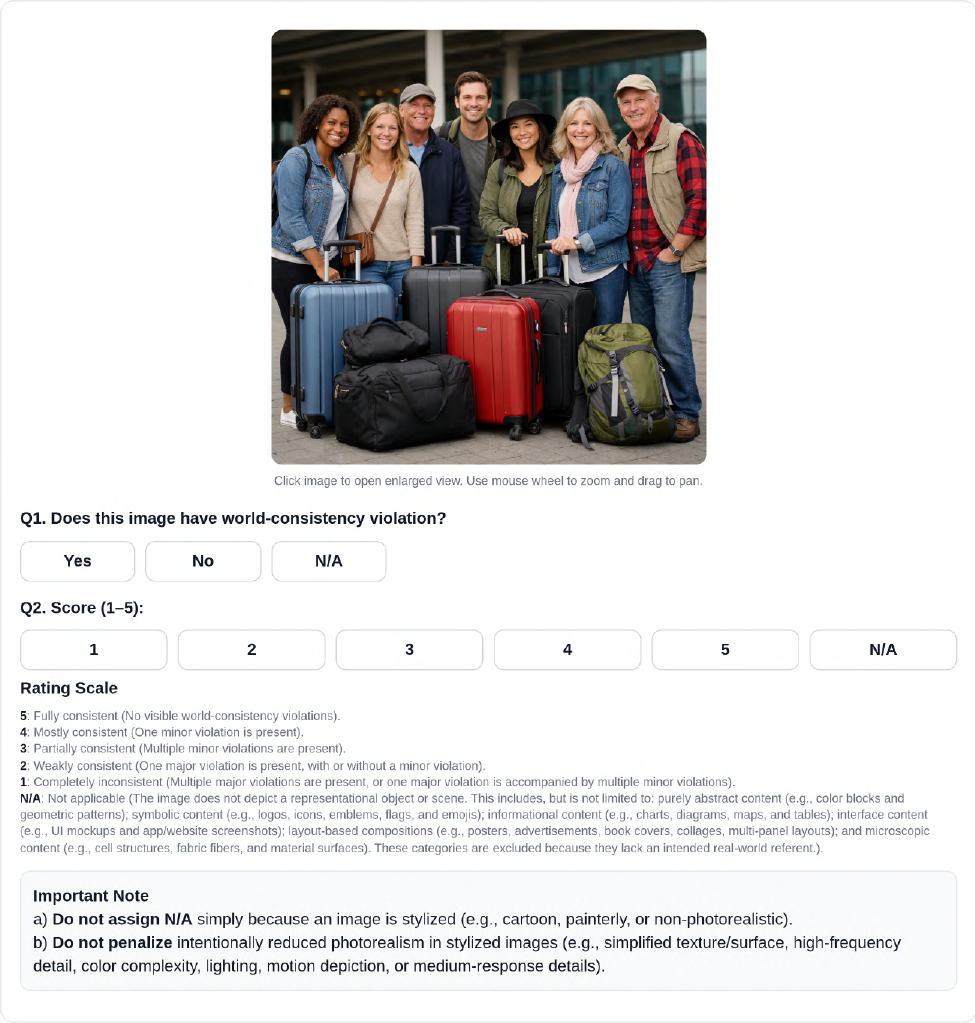}
    \caption{
    AMT interface for rating world-grounded visual consistency.
    Annotators first read the descriptions of taxonomy-guided criteria covering object-level,
    interaction-level, and scene-level violations, and then answer whether the image contains a world-consistency violation and assign a score from 1 to 5.
    "N/A" instruction is also added in the rating scale section.
    }
    \label{fig: amt_interface}
\end{figure}

Similar to common human annotation protocols in text-to-image evaluation~\cite{otani2023toward}, 
we ask three independent annotators on Amazon Mechanical Turk (AMT) to provide an overall world-consistency rating for each generated image. 
For five text-to-image models across two benchmarks, this results in 
$3 \times 5 \times (100 + 800) = 13{,}500$ annotations in total. 
Annotators are required to be AMT Masters, over 18 years old, and willing to review potentially offensive content. 
We pay approximately \$0.08 for each annotation. 
\figref{fig: amt_interface} shows the interface used for human evaluation. 
Annotators rate the world consistency of each image on a scale from 1 to 5, with an additional “N/A” option for images outside the scope of world-consistency evaluation. The rating criteria are as follows:

\begin{description}[leftmargin=0em, labelsep=0.6em, itemsep=0em]
    \item[5:] Fully consistent (with no visible world-consistency violation).
    \item[4:] Mostly consistent (with only one minor violation).
    \item[3:] Partially consistent (with multiple minor violations).
    \item[2:] Weakly consistent (with one major violation with or without an additional minor violation).
    \item[1:] Completely inconsistent (with multiple major violations, or one major violation accompanied by multiple minor violations).
    \item[N/A:] Not applicable. Images that do not contain representational content at the macroscopic object or scene level.
\end{description}

To improve inter-annotator agreement (IAA), we additionally include a binary world-consistency question before the 5-point rating, asking annotators to first judge whether the image contains any perceptible world-consistency violation.
Our collected human ratings show a high level of inter-rater agreement: Krippendorff's $\alpha$ reaches 0.686 and 0.513 on the 5-point world-consistency ratings for COCO-T2I and GenAI-Bench, respectively.
For the binary violation labels, annotators achieve Gwet's AC1~\cite{gwetac1} scores of 0.603 and 0.555 on COCO-T2I and GenAI-Bench, respectively, indicating moderate-to-substantial agreement beyond chance.

\section{Additional Experimental Results}
\label{apx: exp}

\subsection{Holistic versus Structured Violation Detection}
\label{apx: exp_holistic_vs_structured}

\begin{table}[t]
    \centering
    \caption{
        Positive rates and world-consistency violation detection performance on COCO-T2I and GenAI-Bench. Precision and recall are computed against human majority-vote labels. The highest precision and recall within each benchmark are bolded.
    }
    \label{tab: violation_detection_performance}
    \small
    \begin{tabular}{llccc}
        \toprule
        Benchmark & Evaluator & Positive rate (\%)
        & Precision (\%) $\uparrow$ & Recall (\%) $\uparrow$ \\
        \midrule
        \multirow{3}{*}{COCO-T2I}
        & Human majority vote & 48.6 & -- & -- \\
        & TerraVis-H & 8.7 & \textbf{86.0} & 15.4 \\
        & TerraVis & 33.3 & 75.8 & \textbf{51.9} \\
        \midrule
        \multirow{3}{*}{GenAI-Bench}
        & Human majority vote & 69.7 & -- & -- \\
        & TerraVis-H & 14.8 & \textbf{93.3} & 19.8 \\
        & TerraVis & 37.8 & 89.1 & \textbf{48.3} \\
        \bottomrule
    \end{tabular}
\end{table}

GPTScore, TerraVis-H, and TerraVis form a progression from generic holistic scoring to task-specific holistic scoring and, finally, structured evaluation.
GPTScore uses a generic instruction to produce a holistic score, whereas TerraVis-H uses the full instructions provided to human annotators while retaining holistic scoring.
TerraVis further decomposes the assessment into explicit checks and aggregates detected violations according to severity.

The improvement from the near-zero correlations of GPTScore to the higher correlations of TerraVis-H suggests that task-specific instructions help align evaluation with human criteria.
To examine the additional contribution of structured evaluation, we compare TerraVis-H and TerraVis on image-level world-consistency violation detection.
Human reference positives are images judged to contain at least one world-consistency violation by a majority of annotators (at least two of three).
\tabref{tab: violation_detection_performance} reports the positive rates for the human reference and both evaluators, together with precision and recall computed against these reference labels.

We observe TerraVis-H exhibits a conservative detection pattern as its positive rates are substantially lower than those of the human reference on both benchmarks, with high precision but low recall, while TerraVis achieves $3.37\times$ and $2.44\times$ the recall of TerraVis-H on COCO-T2I and GenAI-Bench, respectively.
These gains are accompanied by precision decreases of $10.2$ and $4.2$ percentage points, to $75.8\%$ and $89.1\%$, respectively.
The low recall of TerraVis-H indicates that even with detailed task instructions, holistic scoring can leave many images containing human-identified violations undetected.
TerraVis's substantially higher recall suggests that decomposing the assessment into explicit checks helps identify more of these images, although at some cost to precision.
This improved detection provides a possible explanation for TerraVis's stronger correlations with human judgments.

\begin{table}[t]
    \centering
    \caption{
    Ablation of individual violation types in TerraVis, with results aggregated across COCO-T2I and GenAI-Bench.
    \textsc{Only} reports Spearman’s $\rho$ with human ratings using each type alone; 
    \textsc{LOO} reports the change in $\rho$ when that type is removed.
    Underline highlights the individual violation type with the largest marginal contribution to correlation with human ratings.
    }
    \label{tab: ablation_violation_types}
    \setlength{\tabcolsep}{6pt}
    \renewcommand{\arraystretch}{1.08}
    \setlength{\fboxsep}{1pt}
    \begin{tabular}{@{}lrr@{}}
        \toprule
        Violation type
        & \multicolumn{1}{c}{\textsc{Only}: $\rho$}
        & \multicolumn{1}{c}{\textsc{LOO}: $\Delta$} \\
        \midrule

        \textbf{Object (all 7)}
        & \textbf{0.384} & \textbf{\textminus 0.1488} \\
        \quad Object identity
        & 0.275 & \textminus 0.0330 \\
        \quad Structural distortion
        & 0.241 & \underline{\textminus 0.0419} \\
        \quad Biological anatomy
        & 0.109 & \textminus 0.0012 \\
        \quad Non-biological structure
        & 0.124 & \textminus 0.0044 \\
        \quad Texture/surface
        & 0.068 & \textminus 0.0026 \\
        \quad Symbolic content
        & 0.114 & \textminus 0.0084 \\
        \quad Object-context
        & 0.216 & \textminus 0.0121 \\

        \addlinespace[3pt]
        \textbf{Interaction (all 8)}
        & \textbf{0.253} & \textbf{\textminus 0.0327} \\
        \quad Contact state
        & 0.104 & \textminus 0.0016 \\
        \quad Support and stability
        & 0.077 & \textminus 0.0014 \\
        \quad Dynamic response
        & 0.161 & \textminus 0.0032 \\
        \quad Medium interaction
        & 0.146 & \textminus 0.0027 \\
        \quad Optical effect
        & 0.115 & \textminus 0.0021 \\
        \quad Energy source
        & 0.065 & \textminus 0.0012 \\
        \quad Thermal response
        & 0.093 & \textminus 0.0021 \\
        \quad Behavior-event
        & 0.147 & \textminus 0.0090 \\

        \addlinespace[3pt]
        \textbf{Scene (all 3)}
        & \textbf{0.174} & \textbf{\textminus 0.0164} \\
        \quad Relative scale
        & 0.150 & \textminus 0.0058 \\
        \quad Depth-order/occlusion
        & 0.069 & \textminus 0.0051 \\
        \quad Region continuity
        & 0.099 & \textminus 0.0055 \\

        \midrule
        \textbf{Full (18)}
        & \textbf{0.428} & -- \\
        \bottomrule
    \end{tabular}
    \vspace{-3mm}
\end{table}

\subsection{Ablation of Violation Types}
\label{apx: exp_ablation_types}

\tabref{tab: ablation_violation_types} reports ablations of individual violation types in TerraVis.
The best individual check achieves $\rho=0.275$, compared with $0.428$ for the full framework, suggesting that performance benefits from broad coverage across violation types.
Removing the check for structural distortion yields the largest decrease in correlation ($\Delta\rho=-0.0419$).
Although individual LOO effects are modest, they are negative for all 18 checks, suggesting that each provides additional signal.
These findings support retaining fine-grained checks to capture complementary aspects of world consistency.

\begin{table}[t]
    \centering
    \caption{
        Sensitivity analysis of TerraVis aggregation hyperparameters.
        Rank correlations remain stable across a broad range of $\lambda$ and $\alpha$.
        The default setting $(\lambda=1,\alpha=0.5)$ achieves near-optimal performance.
    }
    \label{tab: ablation_lamda_and_alpha}
    \resizebox{0.8\linewidth}{!}{%
    \begin{tabular}{lccccc}
    \toprule
    Setting & $\lambda$ & $\alpha$ & Pearson's $r$ & Spearman's $\rho$ & Kendall's $\tau$ \\
    \midrule
    Best Pearson  & 0.9 & 0.5 & \textbf{0.4230} & 0.4281          & 0.3471 \\
    Best Spearman & --  & 0.3 & 0.3826          & \textbf{0.4284} & 0.3478 \\
    Best Kendall  & --  & 1.0 & 0.3840	        & 0.4264          & \textbf{0.3481} \\
    Default       & 1.0 & 0.5 & 0.4229          & 0.4281          & 0.3477 \\
    \midrule
    Range over grid & -- & -- & 0.3797--0.4230  & 0.4234--0.4284  & 0.3463–0.3481 \\
    \bottomrule
    \end{tabular}}
\end{table}

\subsection{Sensitivity to Aggregation Hyperparameters}
\label{apx: sensitivity_lamda_alpha}

We evaluate the sensitivity of TerraVis to the aggregation hyperparameters $\lambda$ and $\alpha$ in \eqref{eq.1} on the COCO-T2I benchmark.
For a fixed $\alpha$, varying $\lambda>0$ only induces a monotonic rescaling of the score, so rank-based correlations such as Spearman's $\rho$ and Kendall's $\tau$ remain unchanged.
Pearson correlation, however, can vary because it is sensitive to nonlinear transformations of the score range.
While changing $\alpha$ adjusts the relative weight of minor violations and can affect the ranking, TerraVis remains stable across a broad range of $\alpha$, showing that the aggregation is robust to this hyperparameter choice.
As shown in \tabref{tab: ablation_lamda_and_alpha}, the default setting $(\lambda=1,\alpha=0.5)$ achieves near-optimal correlations with human judgments.

\begin{table}[t]
    \centering
    \caption{
        Pairwise Kendall's $\tau$ between T2I model rankings produced by TerraVis using different MLLM evaluators.
        \textit{Group 1} comprises GPT-5.5 and Gemma 4 31B, while \textit{Group 2} comprises the remaining three evaluators.
        Shaded rows highlight the agreement within \textit{Group 1} and the mean pairwise agreement within \textit{Group 2}.
    }
    \label{tab: evaluator_ranking_agreement}
    \resizebox{\linewidth}{!}{%
    \begin{tabular}{lcc}
        \toprule
        Evaluator pair
        & COCO-T2I ($\tau$)
        & GenAI-Bench ($\tau$) \\
        \midrule

        \multicolumn{3}{l}{\textit{Group 1}} \\
        \rowcolor{gray!12}
        GPT-5.5 -- Gemma 4 31B
        & \textbf{0.80} & \textbf{0.80} \\

        \midrule
        \multicolumn{3}{l}{\textit{Group 2}} \\
        Gemma 4 26B-A4B -- Qwen3-VL-30B-A3B-Instruct
        & 1.00 & 0.80 \\
        Gemma 4 26B-A4B -- Qwen3-VL-32B-Instruct
        & 0.80 & 0.80 \\
        Qwen3-VL-30B-A3B-Instruct -- Qwen3-VL-32B-Instruct
        & 0.80 & 1.00 \\
        \rowcolor{gray!12}
        Group mean (3 pairs)
        & \textbf{0.87} & \textbf{0.87} \\

        \midrule
        \multicolumn{3}{l}{\textit{Between groups}} \\
        GPT-5.5 -- Gemma 4 26B-A4B
        & 0.60 & 0.80 \\
        GPT-5.5 -- Qwen3-VL-30B-A3B-Instruct
        & 0.60 & 0.60 \\
        GPT-5.5 -- Qwen3-VL-32B-Instruct
        & 0.60 & 0.60 \\
        Gemma 4 31B -- Gemma 4 26B-A4B
        & 0.60 & 0.60 \\
        Gemma 4 31B -- Qwen3-VL-30B-A3B-Instruct
        & 0.40 & 0.40 \\
        Gemma 4 31B -- Qwen3-VL-32B-Instruct
        & 0.20 & 0.40 \\

        \midrule
        Overall mean (10 pairs)
        & 0.64 & 0.68 \\
        \bottomrule
    \end{tabular}}
    \vspace{-3mm}
\end{table}

\subsection{Ranking Agreement across Evaluators}
\label{apx: exp_ranking_agreement}

\tabref{tab: evaluator_ranking_agreement} reports pairwise agreement between T2I model rankings produced by TerraVis using different MLLM evaluators.
Mean Kendall's $\tau$ across all ten evaluator pairs is 0.64 on COCO-T2I and 0.68 on GenAI-Bench.
Agreement is higher within groups defined by evaluators' correlations with human judgments.
GPT-5.5 and Gemma 4 31B, the two evaluators most closely aligned with human judgments, achieve $\tau=0.80$ on both benchmarks.
The remaining three evaluators also agree closely among themselves, with a mean pairwise $\tau=0.87$ on both benchmarks.
Between-group agreement is lower, averaging $\tau=0.50$ on COCO-T2I and $\tau=0.57$ on GenAI-Bench.
These results support the consistency of TerraVis rankings within each evaluator group, while suggesting that differences in alignment with human judgments may contribute to the remaining ranking variation.

\begin{table}[t]
    \centering
    \caption{
        Computational efficiency on 200 randomly sampled images using a single NVIDIA H100 GPU.
        Average time per image is the reciprocal of throughput; relative time is normalized to RAHF, the fastest method.
        Higher throughput and lower processing time indicate better efficiency.
    }
    \label{tab:computational_efficiency}
    \resizebox{\linewidth}{!}{%
    \begin{tabular}{llrrr}
        \toprule
        Metric
        & Judge / backbone
        & \shortstack{Throughput\\(images/s) $\uparrow$}
        & \shortstack{Average time\\(s/image) $\downarrow$}
        & \shortstack{Relative time\\$\downarrow$} \\
        \midrule
        RAHF
        & ViT-L + T5 multi-head
        & 43.06 & 0.023 & $1.0\times$ \\
        HPSv3
        & Qwen2-VL 7B
        & 15.35 & 0.065 & $2.8\times$ \\
        VQAScore
        & Gemma 4 31B
        & 13.35 & 0.075 & $3.2\times$ \\
        \textbf{TerraVis (ours)}
        & Gemma 4 31B
        & 0.855 & 1.17 & $50.4\times$ \\
        TIFA
        & Gemma 4 31B
        & 0.799 & 1.25 & $53.9\times$ \\
        \bottomrule
    \end{tabular}
    }
\end{table}

\subsection{Computational Efficiency Analysis}
\label{apx: exp_efficiency}

We compare the computational efficiency of TerraVis with four baseline metrics on 200 randomly sampled images using a single NVIDIA H100 GPU.
We use Gemma 4 31B as the evaluation backbone for TerraVis, VQAScore, and TIFA.
Table~\ref{tab:computational_efficiency} reports throughput, average processing time per image, and processing time relative to the fastest method.

TerraVis achieves a throughput of 0.855 images/s, corresponding to an average processing time of 1.17 seconds per image.
Its multi-stage evaluation requires approximately $15.6\times$ and $18.0\times$ the processing time of the single-pass MLLM-based metrics VQAScore and HPSv3, respectively, and $50.4\times$ that of RAHF.
TerraVis is nevertheless slightly faster than TIFA, another multi-stage method that generates question--answer pairs from the prompt before performing visual question answering.
These results highlight the computational cost of structured evaluation, while the absolute throughput of approximately one image per second suggests that TerraVis remains practical for benchmark-scale evaluation under this experimental setting.

\subsection{Distribution of Detected Violations}
\label{apx: exp_violation_distribution}

\begin{figure}[p]
    \centering
    \includegraphics[width=\linewidth]{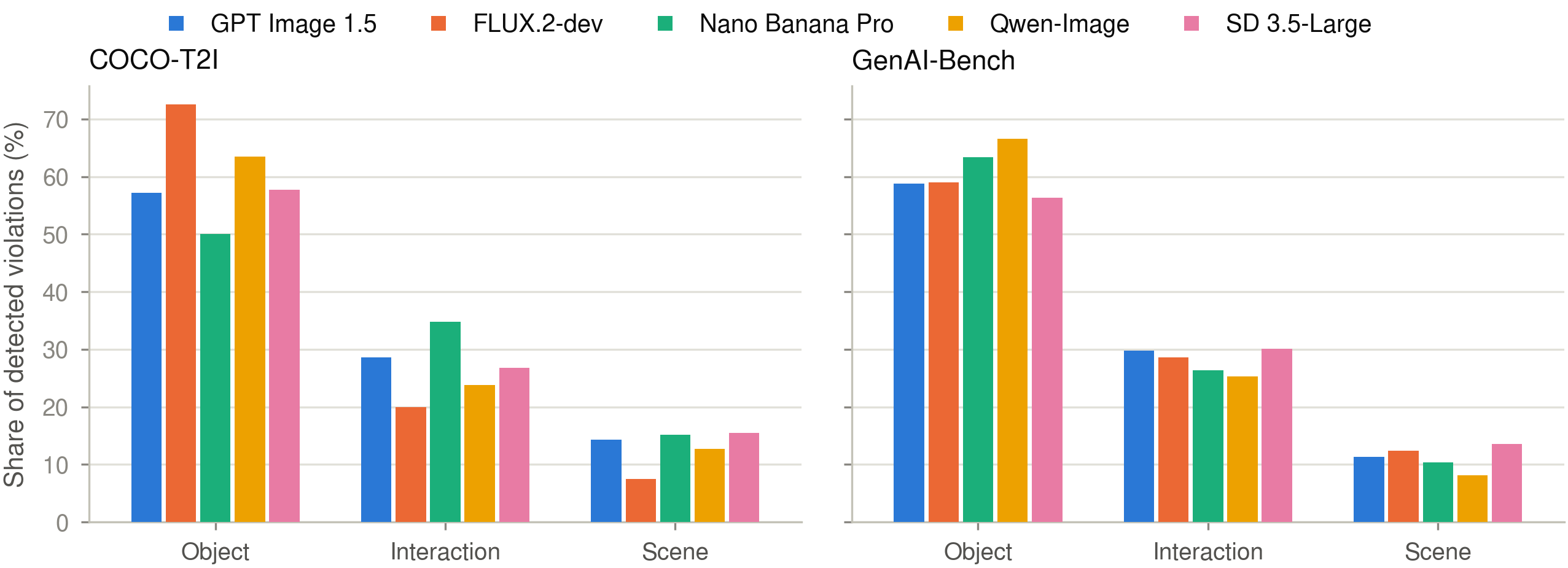}
    \caption{
        \textbf{Share of detected violations by level.}
        For each model, bars give the percentage of its TerraVis-detected violations that fall at the object, interaction, and scene levels on COCO-T2I (left) and GenAI-Bench (right); shares sum to 100\% for each model and benchmark. 
        Object-level violations dominate for every model, yet the interaction and scene levels together still account for 27.5--50.0\% of all violations.
    }
    \label{fig: violation_share_level}
\end{figure}

\begin{figure}[p]
    \centering
    \includegraphics[width=\linewidth]{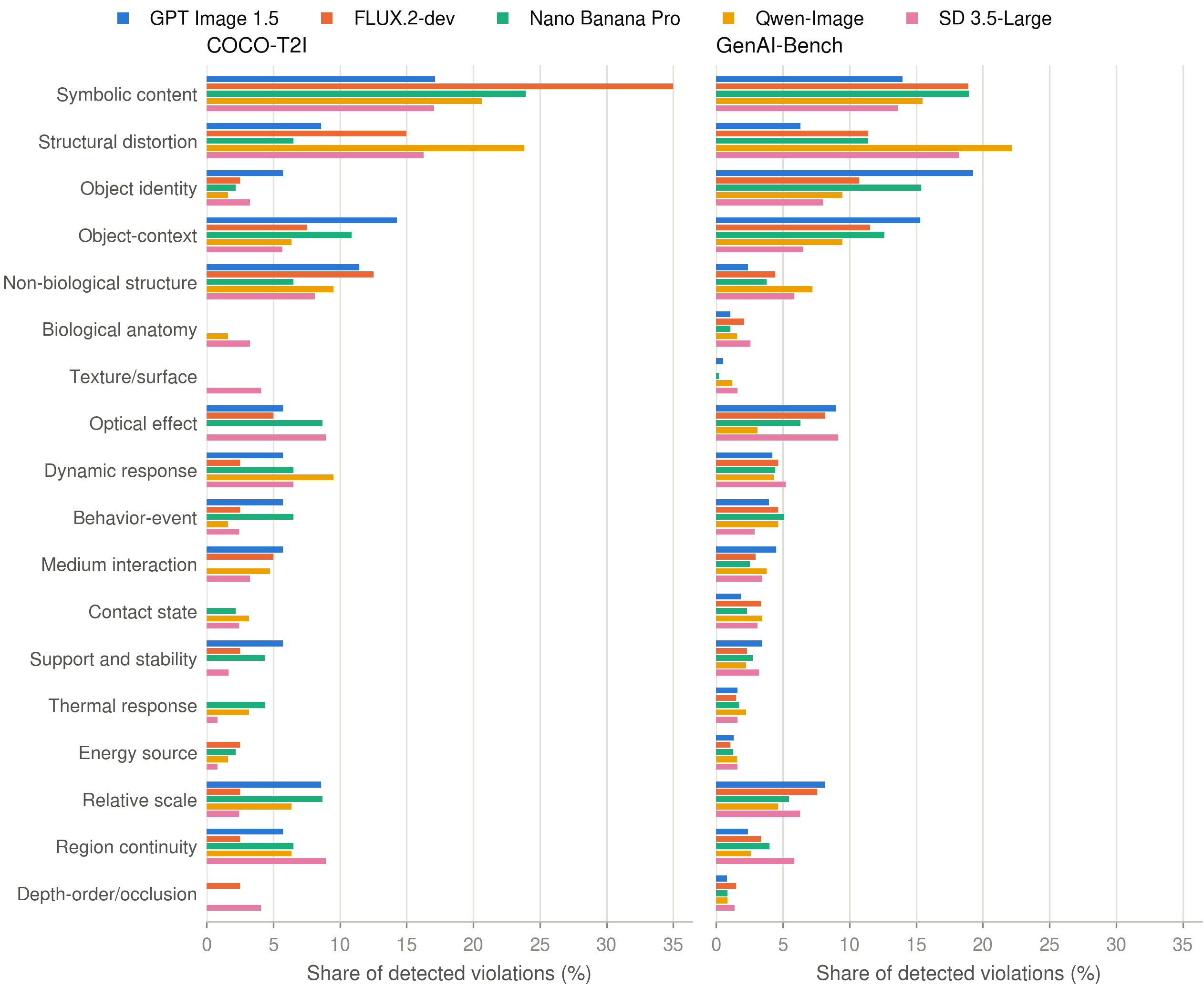}
    \caption{
        \textbf{Share of detected violations by type.}
        The shares in \figref{fig: violation_share_level} broken down into the 18 violation types. 
        The top seven rows are object-level types, the next eight interaction-level, and the bottom three scene-level; shares sum to 100\% for each model and benchmark. 
        The breakdown exposes model-specific profiles that recur on both benchmarks, e.g., \textit{structural distortion} for Qwen-Image and \textit{object-context} and \textit{relative scale} for GPT Image 1.5.
    }
    \label{fig: violation_share_type}
\end{figure}

We break down the violations detected by TerraVis for each model at two granularities: the three levels in \figref{fig: violation_share_level} and the 18 violation types in \figref{fig: violation_share_type}. 
Each bar is the share of a model's detected violations on a benchmark that falls into a given level or type.
Because shares are normalized per model, they describe the composition of a model's violations, not how often it violates world-consistency, which is reflected in its overall TerraVis score.
COCO-T2I also yields far fewer detected violations per model than GenAI-Bench, so its type-level shares are coarser; we emphasize patterns that hold on both benchmarks.

\paragraph{Violations extend beyond the object level.}
Object-level violations are the most common for every model on both benchmarks (around 50.0\% to 72.0\% on COCO-T2I), and the ordering object $>$ interaction $>$ scene holds in all ten model--benchmark pairs.
Yet the interaction and scene levels together account for approximately 25\%--50\% of violations, which would be overlooked by evaluations that inspect objects in isolation. 
Consistent with this finding, removing either level leads to lower agreement with human ratings, as shown in \tabref{tab: ablation_violation_types}. 
At this coarse granularity, however, the models appear similar, particularly on GenAI-Bench, with object-, interaction-, and scene-level shares of approximately 6:3:1, respectively.

\paragraph{Fine-grained types reveal model-specific profiles.}
The type-level breakdown further distinguishes models that appear similar at a coarser level of granularity:
\begin{itemize}
    \item \textbf{GPT Image 1.5 and Nano Banana Pro} exhibit substantially higher violation rates than the other models in the \textit{object-context} category. 
    
    \item \textbf{Qwen-Image and SD 3.5-Large} show the opposite pattern for \textit{structural distortion}, whose share exceeds the combined shares of \textit{object identity} and \textit{object-context} violations on both benchmarks. 
    
    \item \textbf{FLUX.2-dev} is characterized by a high prevalence of \textit{symbolic content} violations, which constitute its most frequent type on both benchmarks. 
    On COCO-T2I, this category accounts for 35.0\% of violations, the largest single share shown in the figure.
    
    \item \textbf{SD~3.5-Large}, the oldest model in our comparison, exhibits the largest shares of \textit{biological anatomy}, \textit{texture/surface}, \textit{optical effect}, and \textit{region continuity} on both benchmarks. 
\end{itemize}

Notably, the two violation types exhibiting the largest cross-model variation on both benchmarks, \textit{structural distortion} and \textit{object identity}, are also the two types whose removal leads to the largest reductions in agreement with human ratings, as shown in \tabref{tab: ablation_violation_types}. 
This correspondence suggests that the dimensions along which TerraVis most clearly distinguishes models are also among those most strongly associated with human judgments.

\paragraph{Common artifacts, such as \textit{biological anatomy} and \textit{texture/surface}, have been substantially mitigated in newer models.}
These issues were long-standing failure modes of earlier image generators, including SD 3.5-Large and Qwen-Image, yet neither type of violation is detected in the remaining models. 
In contrast, \textit{symbolic content} remains a pervasive failure mode: it ranks among the three most frequent violation types for every model on both benchmarks and is the most frequent type in six of the ten model--benchmark pairs.
Notably, \textit{contact state}, \textit{support and stability}, \textit{thermal response}, and \textit{energy source} each account for less than 4\% of all violations. 
However, these low violation rates should not necessarily be interpreted as evidence that models have mastered these phenomena. 
They may instead be partly attributable to the relative rarity of prompts involving such phenomena, which limits the extent to which their corresponding capabilities are tested.
 
\paragraph{Benchmarks shift the model profiles uniformly.}
Since TerraVis is an image-level, prompt-independent evaluation method, the prompt distribution can influence the distribution of generated images and, consequently, the proportion of detected violations across models. 
For example, from COCO-T2I to GenAI-Bench, the proportion of \textit{object identity} violations increases consistently across all five models. 
This shift can be attributed to the larger proportion of prompts in GenAI-Bench, which often require models to render fictional or implausible objects and scenes, such as live dragons and anthropomorphic animals.
In other words, while the benchmark changes the overall prevalence of certain violation categories, it does not substantially alter the relative profiles of the models. 
This suggests that benchmark-specific prompt composition affects the overall violation rates, whereas persistent differences across models are likely to reflect systematic differences in the generators.

\section{Distinguishing World Consistency from Photorealism, Faithfulness, and Aesthetics}
\label{apx: discussion_wc}

A key motivation for \textbf{TerraVis} is that existing evaluation dimensions for text-to-image generation capture different aspects of image quality, but do not explicitly measure whether an image remains visually plausible with respect to the real world. In this work, we distinguish four related but fundamentally different concepts: \textit{photorealism}, \textit{faithfulness}, \textit{aesthetics}, and \textit{world consistency}. \textit{Photorealism} refers to the extent to which an image visually resembles a natural photograph or real visual observation, emphasizing photographic appearance, texture fidelity, lighting, and perceptual detail. \textit{Faithfulness} (or text-image alignment) measures whether the generated image semantically matches the conditioning prompt, regardless of whether the depicted content is physically or structurally plausible. \textit{Aesthetics} evaluates subjective visual appeal, including composition, color harmony, artistic quality, and overall attractiveness to human observers. In contrast, \textit{world consistency} evaluates whether the depicted objects, interactions, and scenes conform to real-world structural, spatial, biological, and physical constraints. Importantly, these dimensions are neither equivalent nor mutually implied. For example, an image may appear highly realistic and aesthetically pleasing while still containing anatomically impossible hands, inconsistent reflections, or implausible object interactions. Similarly, an image may faithfully follow a prompt describing an impossible event while remaining world-inconsistent. \textbf{TerraVis} therefore treats world consistency as a complementary evaluation dimension that is orthogonal to photorealism, aesthetics, and prompt faithfulness, focusing specifically on perceptible violations of real-world plausibility.

\section{Additional Case Studies}
\label{apx: case_studies}

\begin{figure}[p]
    \centering

    \casegroup{(a) Human Oversights}
          {TerraVis identifies genuine violations that are overlooked by human evaluators}

    \casepanelwide
      {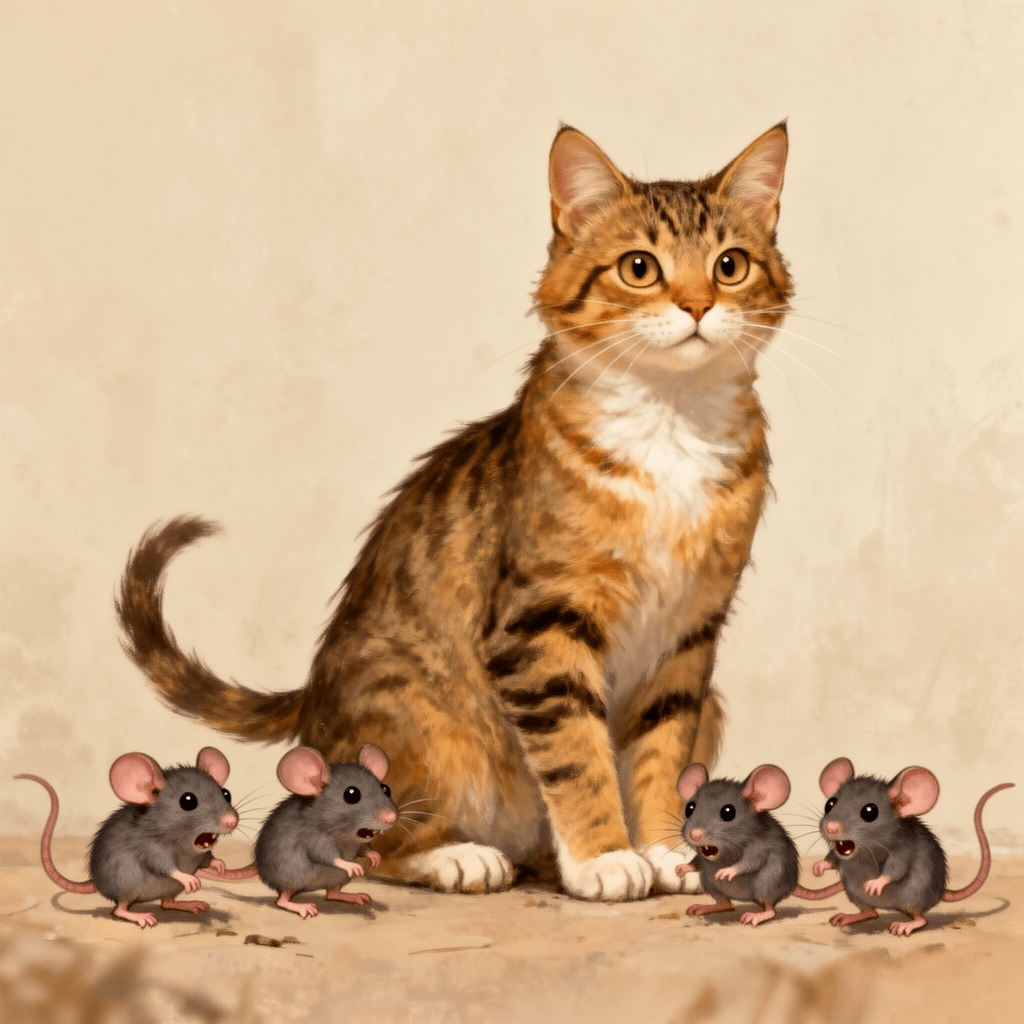}
      {0.368}
      {1.0\,(5,5,5)}
      {FLUX.2-dev / GenAI-Bench-0153}
      {
        \textbf{Behavior-event}\par
        {\fontsize{7}{9}\selectfont\color{darkgray}
          Major $\cdot$ interaction-level
        }\par\vspace{3pt}
    
        \textit{Affected aspect:} mice around the cat.\par
        \textit{Evidence:} the mice stand calmly next to the cat instead of fleeing or showing fear despite being beside a predator.\par\vspace{3pt}

      }
    \hfill
    \casepanelwide
      {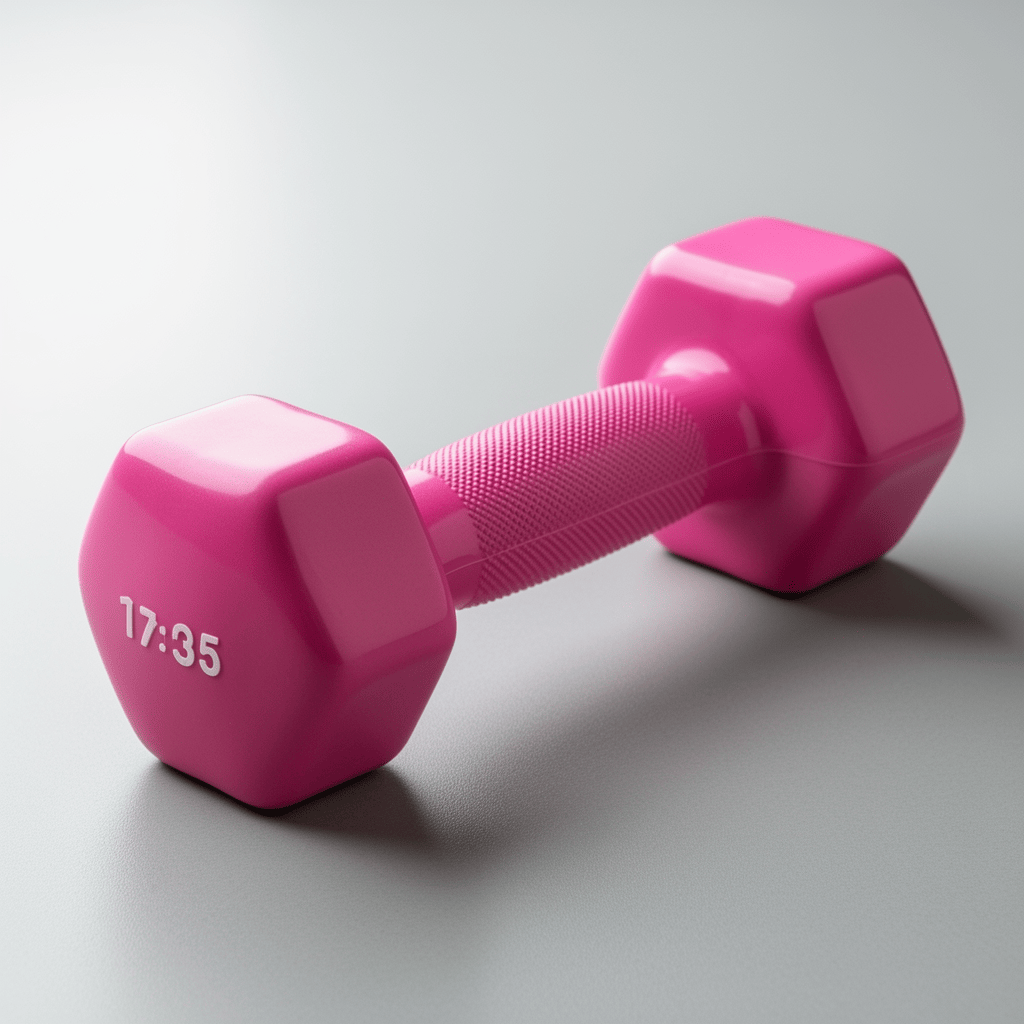}
      {0.368}
      {1.00\,(5,5,5)}
      {FLUX.2-dev / GenAI-Bench-0731}
      {
        \textbf{Symbolic content}\par
        {\fontsize{7}{9}\selectfont\color{darkgray}
          Major $\cdot$ object-level
        }\par\vspace{3pt}
    
        \textit{Affected aspect:} time label on dumbbell.\par
        \textit{Evidence:} the dumbbell has the text “17:35” printed on it, which resembles a time rather than an appropriate weight marking.\par\vspace{3pt}

      }
    \par\vspace{8pt}

    \casegroup{(b) False Positives}
          {TerraVis produces false-positive violations that are not recognized by human evaluators}

    \casepanelwide
      {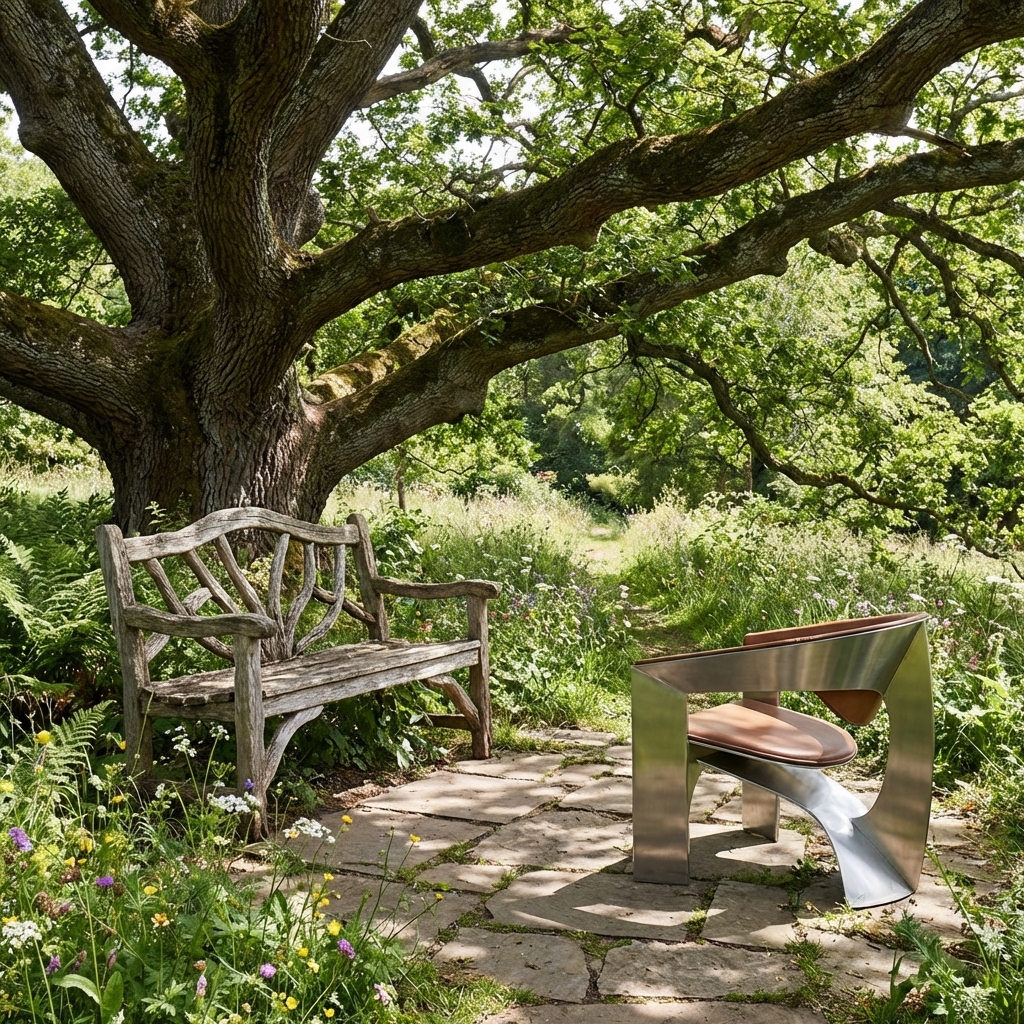}
      {0.018}
      {1.00\,(5,5,5)}
      {Nano Banana Pro / GenAI-Bench-0401}
      {%
        \textbf{Region continuity}\par
        {\fontsize{7}{9}\selectfont\color{darkgray}
          Major $\cdot$ scene-level
        }\par\vspace{3pt}
    
        \textit{Affected aspect:} metallic chair on the right.\par
        \textit{Evidence:} the metallic chair appears poorly integrated with the stone path, with mismatched reflections and weak contact alignment at its base compared to the surrounding scene.\par\vspace{3pt}
    
        {\fontsize{7}{9}\selectfont
          \textbf{Other detected types:} structural distortion, non-biological structure, and contact state.
        }
      }
    \hfill
    \casepanelwide
      {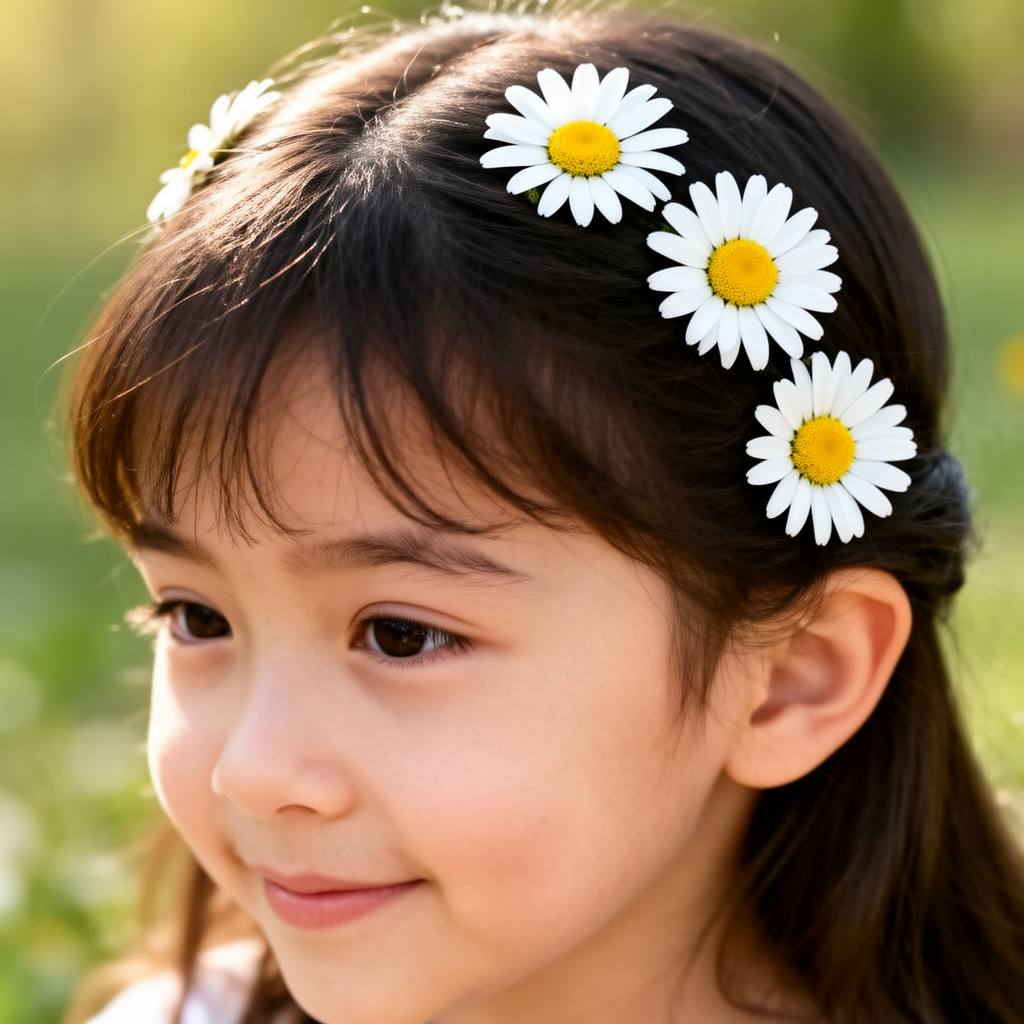}
      {0.135}
      {1.00\,(5,5,5)}
      {FLUX.2-dev / GenAI-Bench-1580}
      {%
        \textbf{Optical effect}\par
        {\fontsize{7}{9}\selectfont\color{darkgray}
          Major $\cdot$ interaction-level
        }\par\vspace{3pt}
    
        \textit{Affected aspect:} daisies on the girl's hair.\par
        \textit{Evidence:} the flowers appear pasted onto the dark hair with little to no contact shadow beneath the petals, making the interaction look physically implausible.\par\vspace{3pt}
    
        {\fontsize{7}{9}\selectfont
          \textbf{Other detected types:} region continuity.
        }
      }
    \par\vspace{8pt}

    \casegroup{(c) Missed Violations}
          {TerraVis misses genuine violations identified by human evaluators}

    \casepanelwide
      {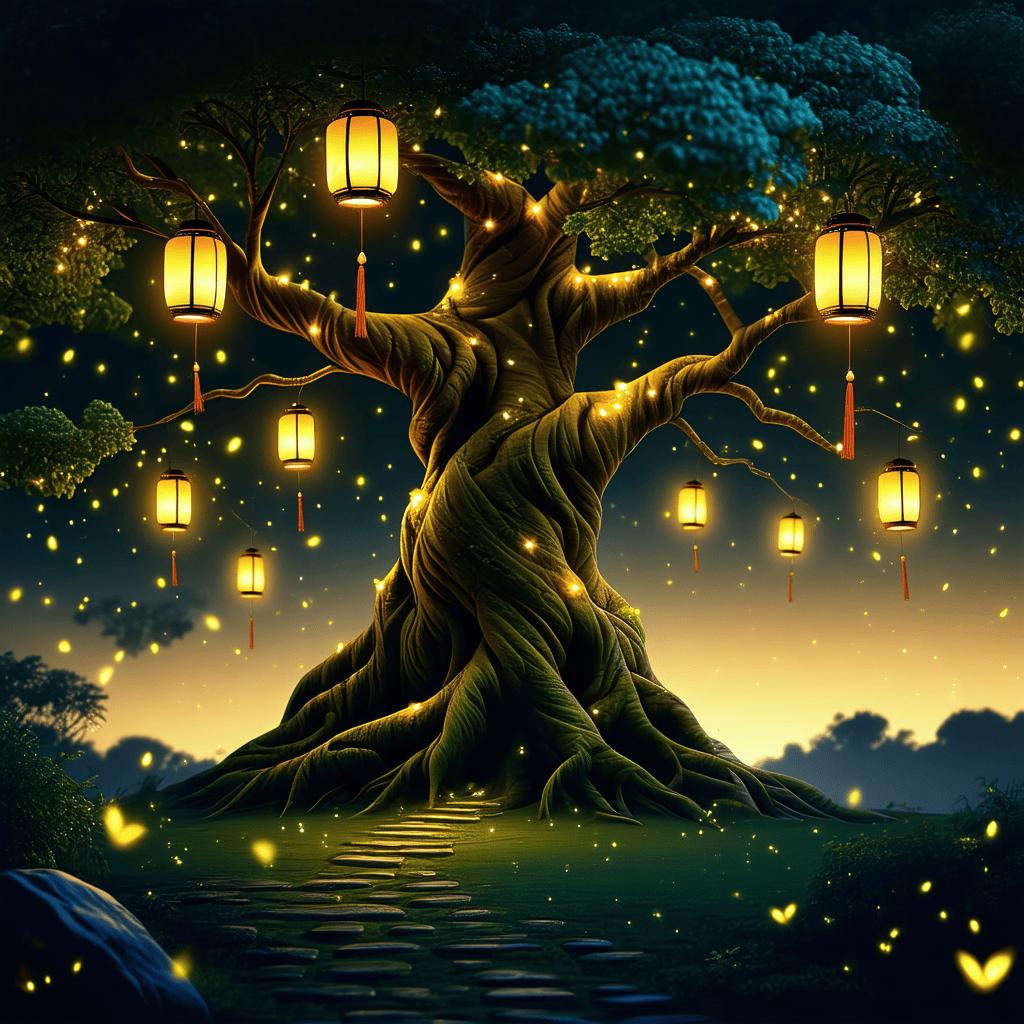}
      {1.000}
      {0.00\,(1,1,1)}
      {SD 3.5-Large / GenAI-Bench-0249}
      {%
        \textcolor{TV}{\textbf{No violation detected}}\par
        \textbf{Disagreement:} TerraVis detects no violations and assigns the maximum world-consistency score of 1.0, whereas all three human raters assign the lowest rating (1/5). Our manual review confirms that several lanterns appear to float below the branches without visible support, a support violation overlooked by TerraVis.
      }
    \hfill
    \casepanelwide
      {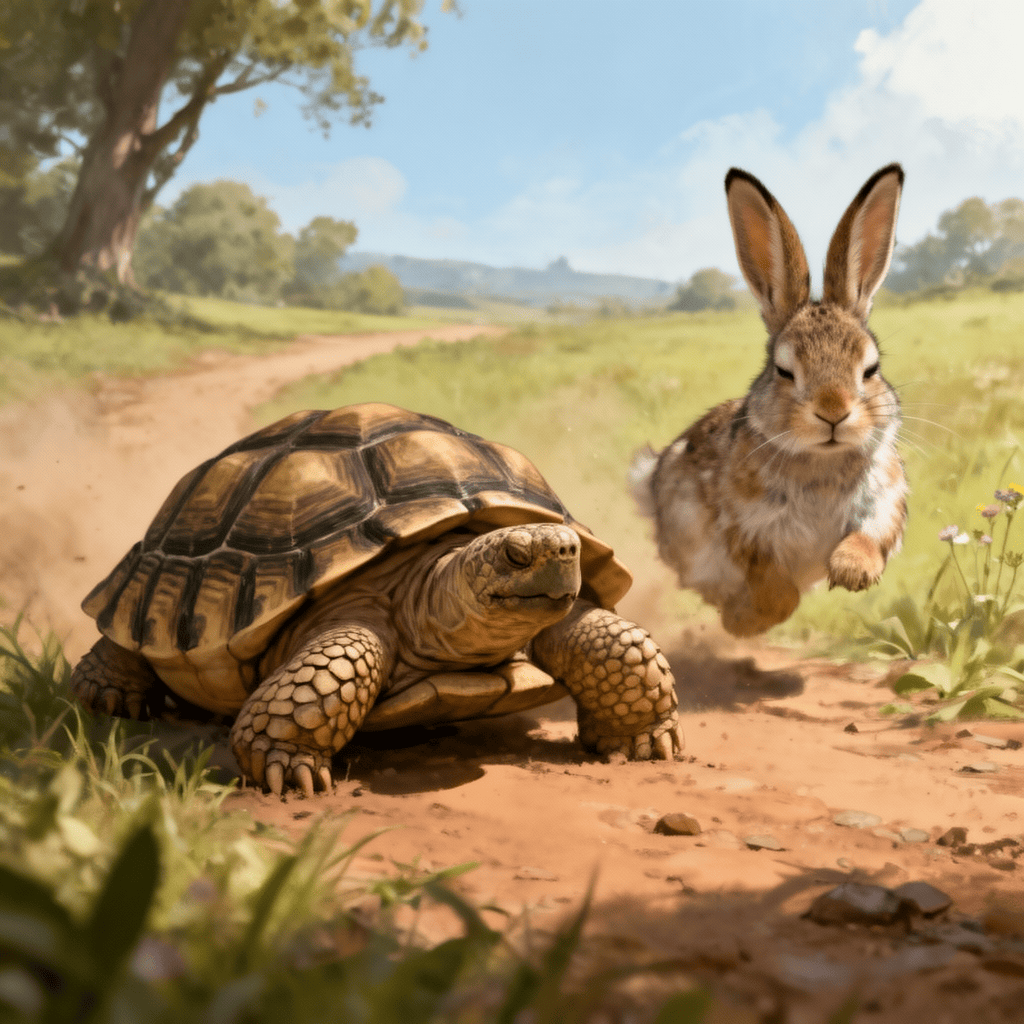}
      {1.000}
      {0.17\,(2,2,1)}
      {FLUX.2-dev / GenAI-Bench-0204}
      {%
        \textcolor{TV}{\textbf{No violation detected}}\par
        \textbf{Disagreement:} TerraVis detects no violations, whereas human raters assign low world-consistency ratings. Our manual review identifies apparent anatomical distortions in the rabbit's lower limbs, with poorly defined paws that appear fused with the body, an anatomy violation overlooked by TerraVis.
      }
    \par\vspace{8pt}

    \casegroup{(d) Rater Disagreement}
          {Human evaluators have inconsistent judgments on the presence or severity of violations}

    \casepanelwide
      {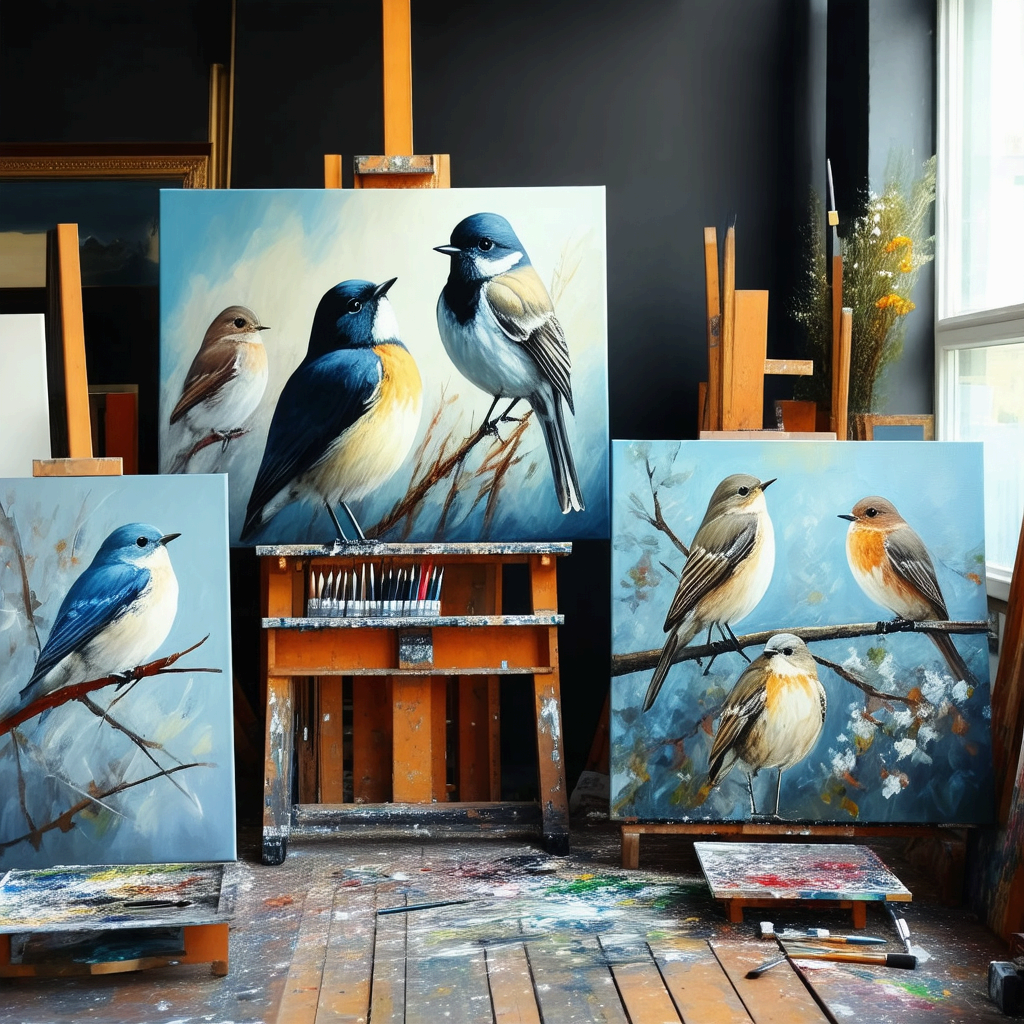}
      {1.00}
      {0.50\,(3,5,1)}
      {SD 3.5-Large / GenAI-Bench-1274}
      {%
        \textcolor{Human}{\textbf{Human Judgements}}\par

        \textbf{Rater 1 (3/5):} Partially consistent.\par\vspace{3pt}
        \textbf{Rater 2 (5/5):} Fully consistent.\par\vspace{3pt}
        \textbf{Rater 3 (1/5):} Completely inconsistent.\par\vspace{4pt}
        \textbf{Disagreement:} Ratings span the full 1--5 scale for the same image, indicating substantial inter-rater disagreement.
      }
    \hfill
    \casepanelwide
      {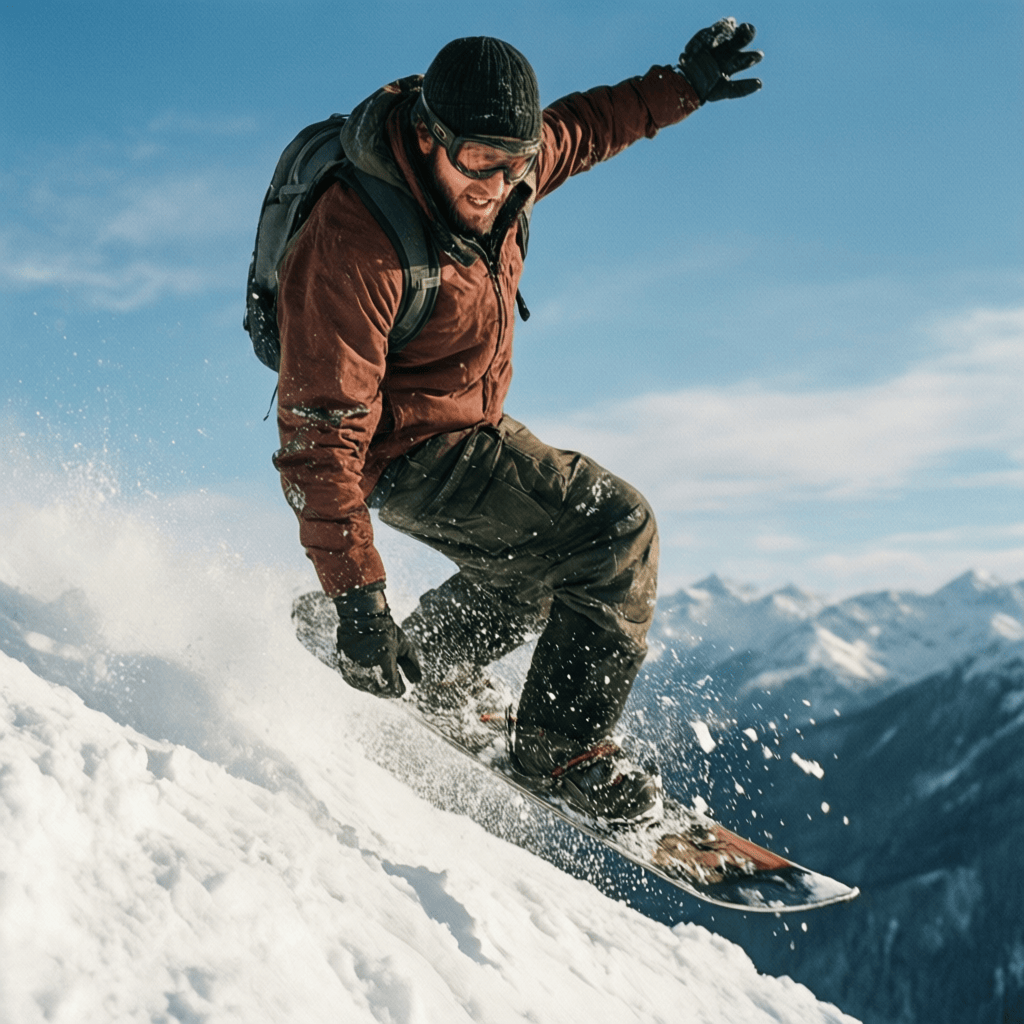}
      {1.00}
      {0.50\,(2,5,2)}
      {Qwen-Image / COCO-T2I-008}
      {%
        \textcolor{Human}{\textbf{Human Judgements}}\par
        
        \textbf{Rater 1 (2/5):} Weakly consistent.\par\vspace{3pt}
        \textbf{Rater 2 (5/5):} Fully consistent.\par\vspace{3pt}
        \textbf{Rater 3 (2/5):} Weakly consistent.\par\vspace{4pt}
        \textbf{Disagreement:} Ratings span from 2 to 5 for the same image, indicating substantial inter-rater disagreement.

      }
    \par\vspace{8pt}
    \caption{
        \textbf{Four patterns of disagreement in world-consistency evaluation.}
        (a) TerraVis detects genuine violations overlooked by human raters. 
        (b) TerraVis incorrectly detects violations in images that all human raters judge fully consistent. 
        (c) TerraVis fails to detect violations that human raters identify and our manual review confirms. 
        (d) Human raters assign substantially different scores to the same image.
        The violation descriptions in (a) and (b) are reproduced verbatim from TerraVis outputs.
    }
    \label{fig: apx_disagreement}
\end{figure}

\begin{figure}[p]
    \centering
    \casegroup{(a) High Alignment but Low Human World-Consistency Scores}{}%
    \baselinecase
        {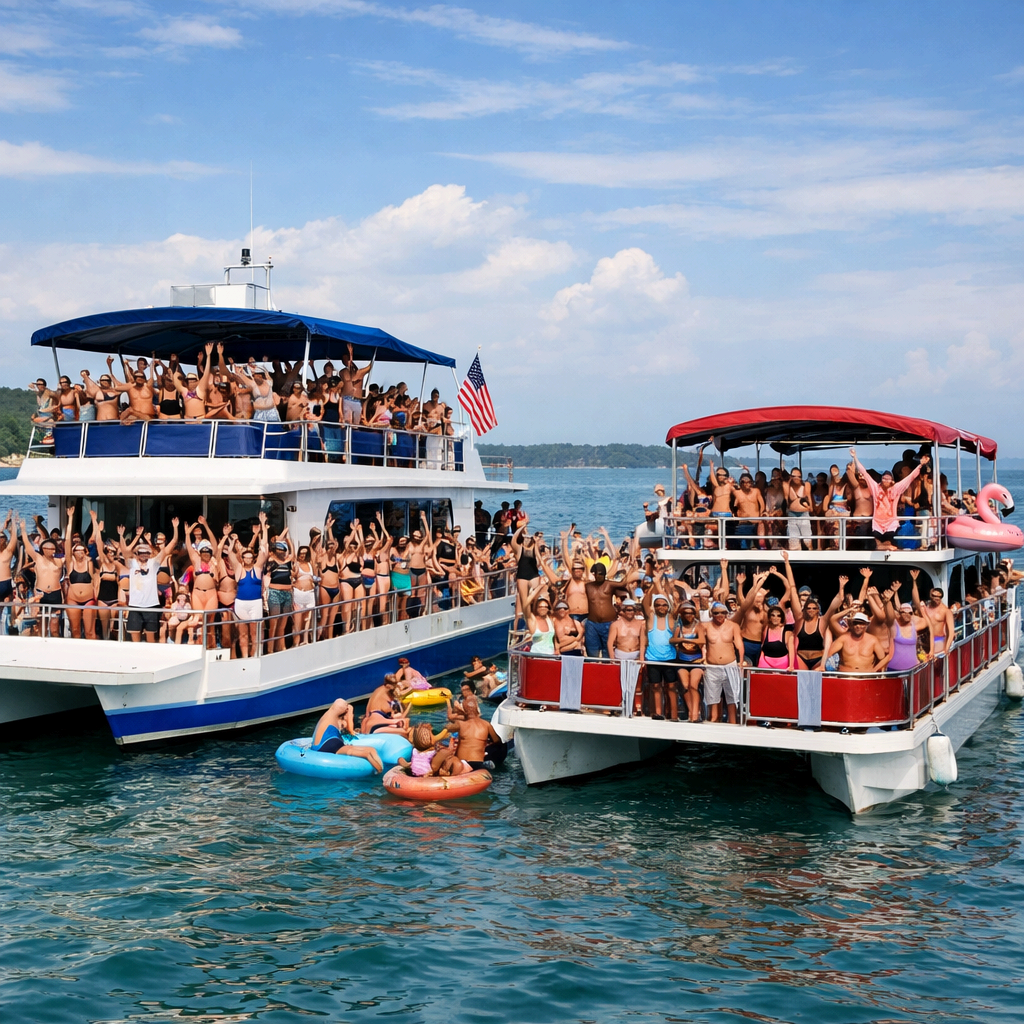}
        {GPT-Image-1.5 / COCO-T2I-051}
        {\textbf{TIFA}~1.00\hfill\textbf{VQAScore}~0.71}
        {0.37}{0.00}
        {A large group of people on a couple boats in the water.}
        {The crowd includes fused arms, although the depicted people, boats, and water match the prompt.}%
    \hfill
    \baselinecase
        {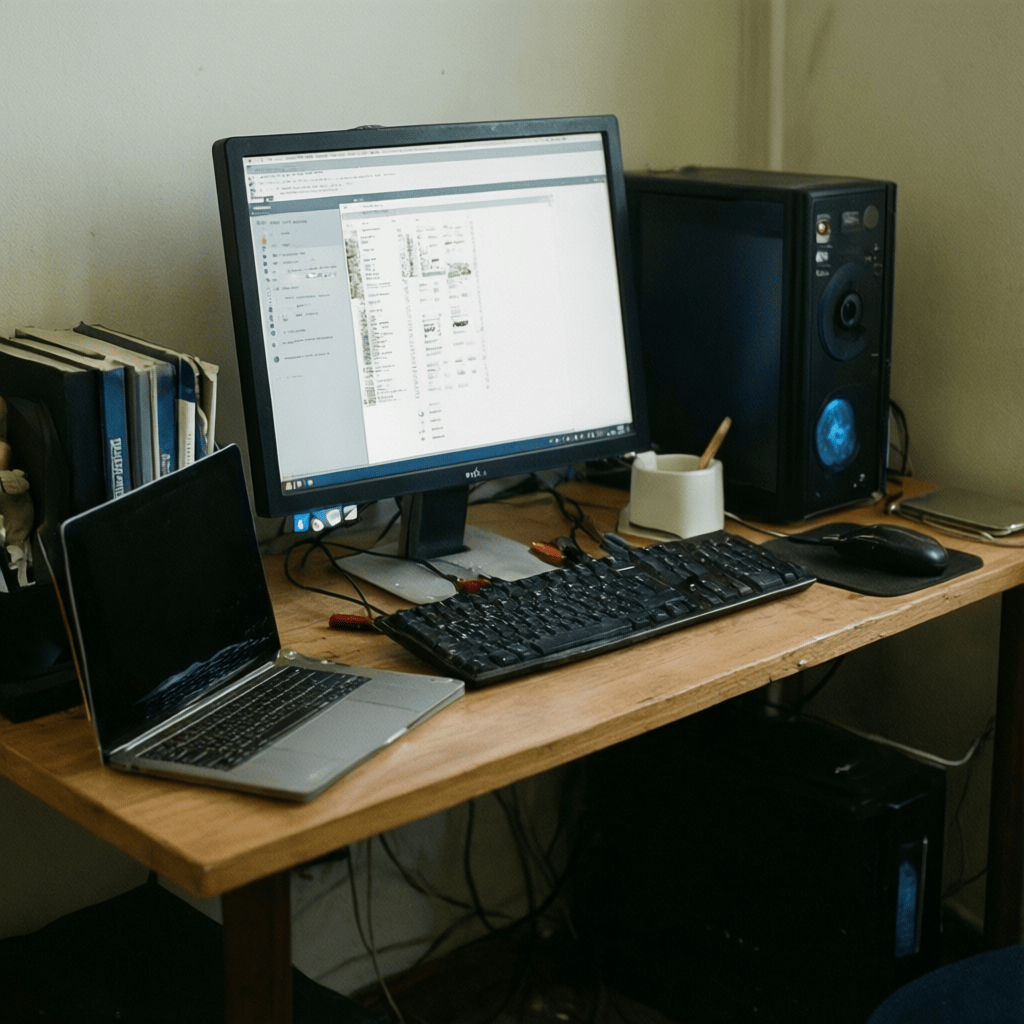}
        {Qwen-Image / COCO-T2I-069}
        {\textbf{TIFA}~1.00\hfill\textbf{VQAScore}~0.64}
        {0.14}{0.00}
        {A computer desk topped with a desktop computer and a laptop.}
        {The laptop has a distorted, melted appearance despite the correct desk and computer content.}
    \par\vspace{7pt}\noindent%
    \baselinecase
        {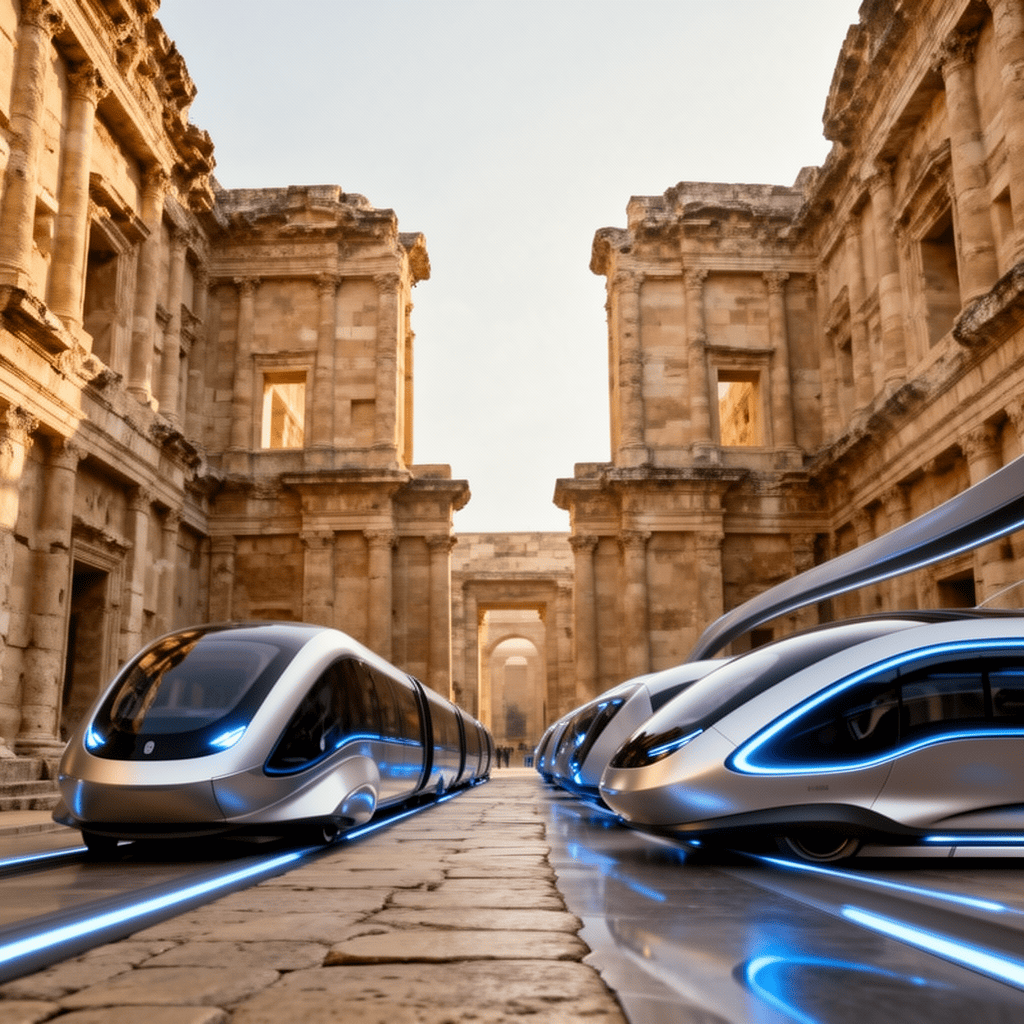}
        {FLUX.2-dev / GenAI-Bench-0051}
        {\textbf{TIFA}~1.00\hfill\textbf{VQAScore}~0.92}
        {0.37}{0.00}
        {Ancient buildings juxtaposed with sleek, futuristic transports.}
        {The requested juxtaposition receives high alignment scores, while both world-consistency assessments are low.}%
    \hfill
    \baselinecase
        {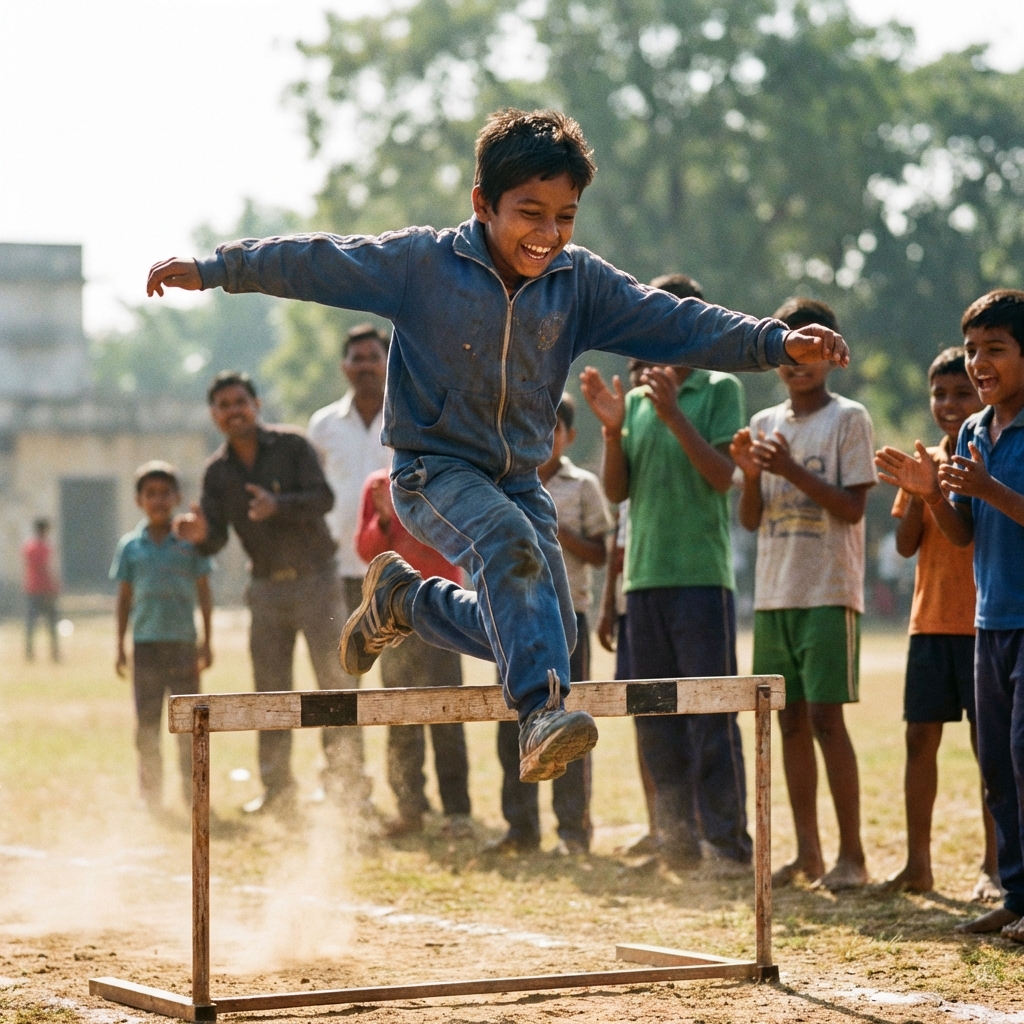}
        {Nano Banana Pro / GenAI-Bench-0028}
        {\textbf{TIFA}~1.00\hfill\textbf{VQAScore}~0.91}
        {0.37}{0.00}
        {A boy leaps over a hurdle.}
        {The action matches the prompt, but the boy's left hand has distorted fingers.}
    \par\vspace{8pt}
    \casegroup{(b) Low Alignment but High Human World-Consistency Scores}{}%
    \baselinecase
        {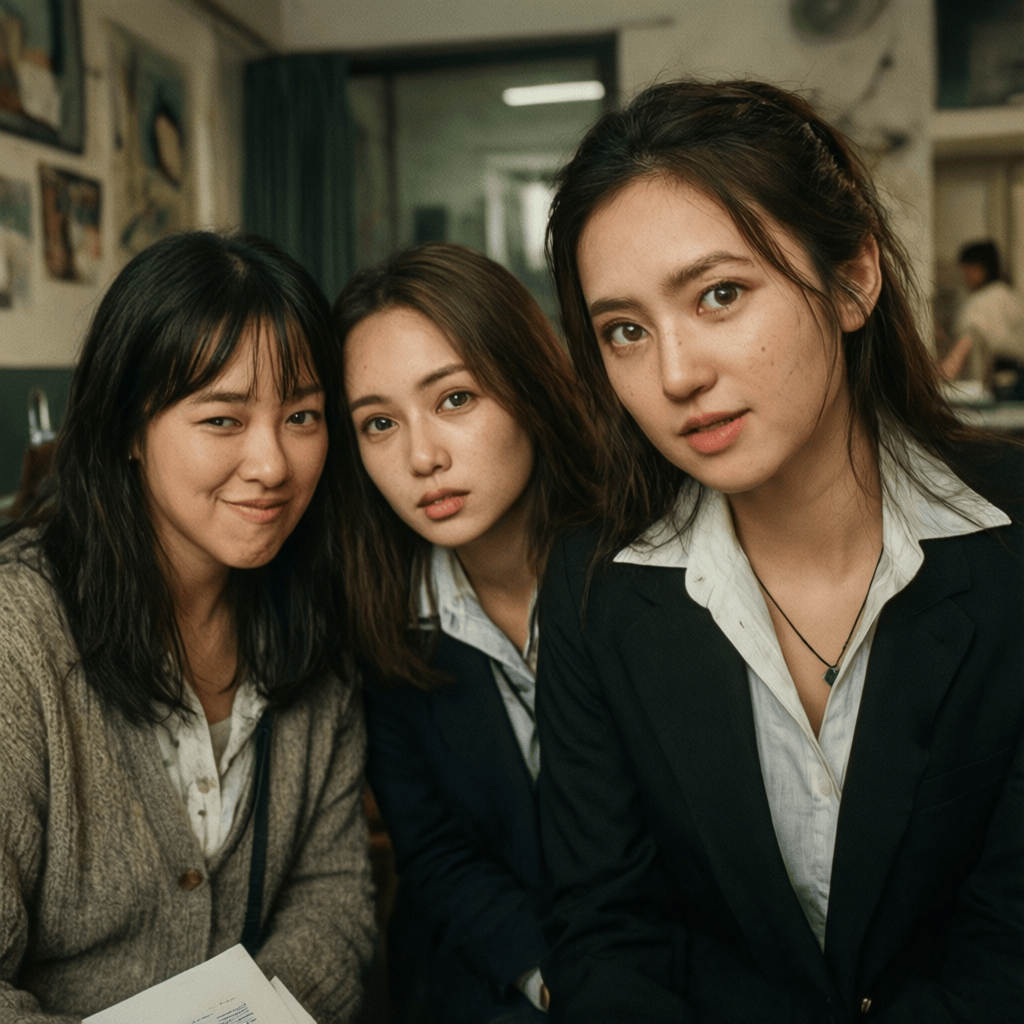}
        {Qwen-Image / GenAI-Bench-1233}
        {\textbf{TIFA}~0.17\hfill\textbf{VQAScore}~0.03}
        {1.00}{1.00}
        {Three jealous teachers.}
        {Jealousy is not clearly conveyed; the people themselves appear visually plausible.}%
    \hfill
    \baselinecase
        {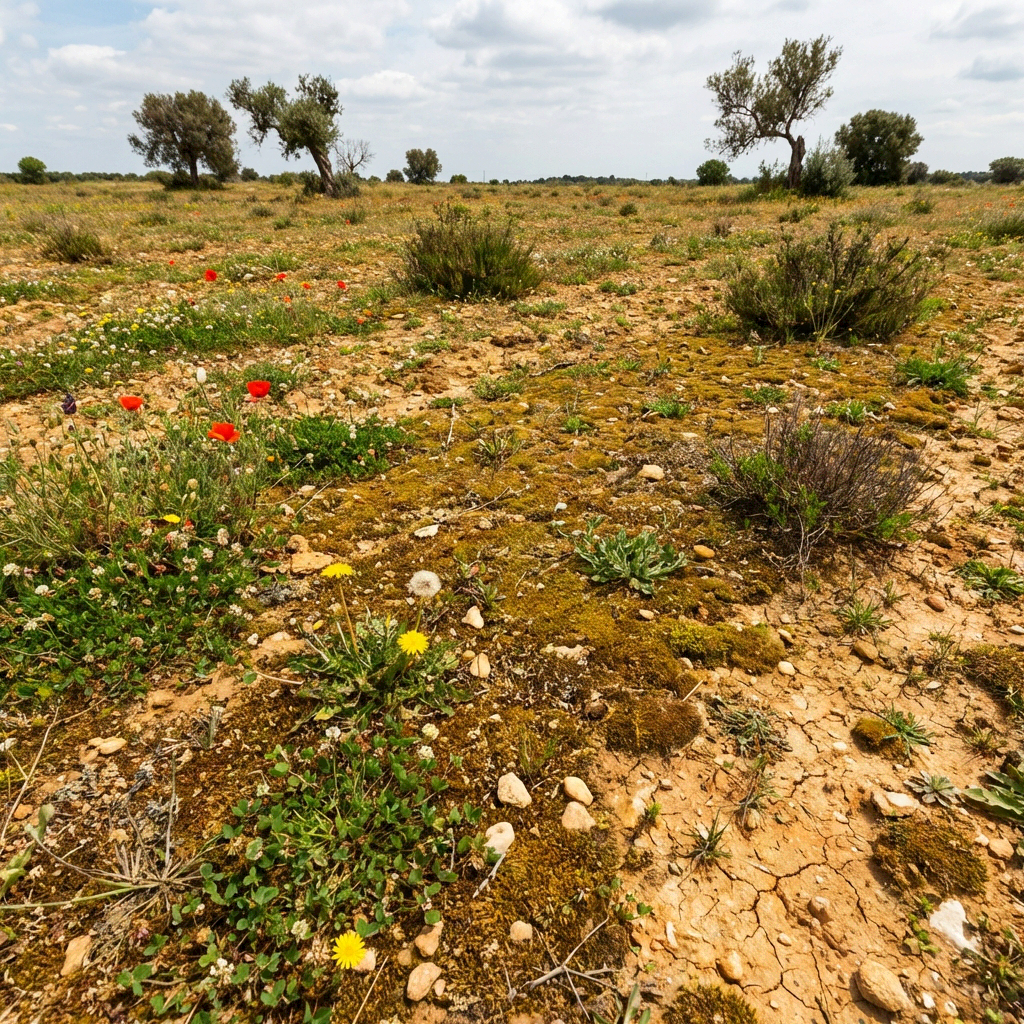}
        {Nano Banana Pro / GenAI-Bench-0459}
        {\textbf{TIFA}~0.33\hfill\textbf{VQAScore}~0.03}
        {1.00}{1.00}
        {A field without a single blade of grass.}
        {Grass remains visible despite the negation in the prompt; its presence is physically plausible.}
    \par\vspace{7pt}\noindent%
    \baselinecase
        {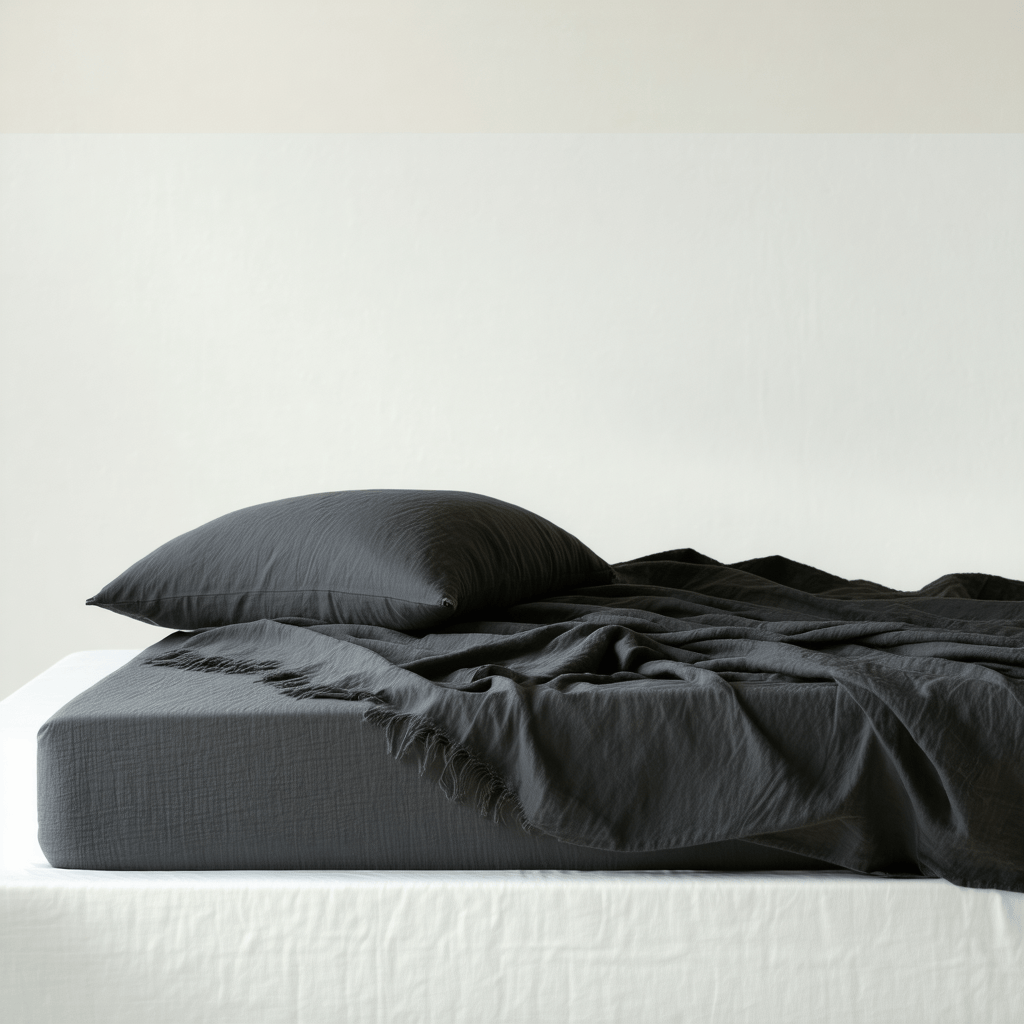}
        {SD 3.5-Large / GenAI-Bench-0443}
        {\textbf{TIFA}~0.44\hfill\textbf{VQAScore}~0.01}
        {1.00}{1.00}
        {A bed with no pillows, only a folded blanket.}
        {The visible pillow contradicts the prompt, while the bedroom remains world consistent.}%
    \hfill
    \baselinecase
        {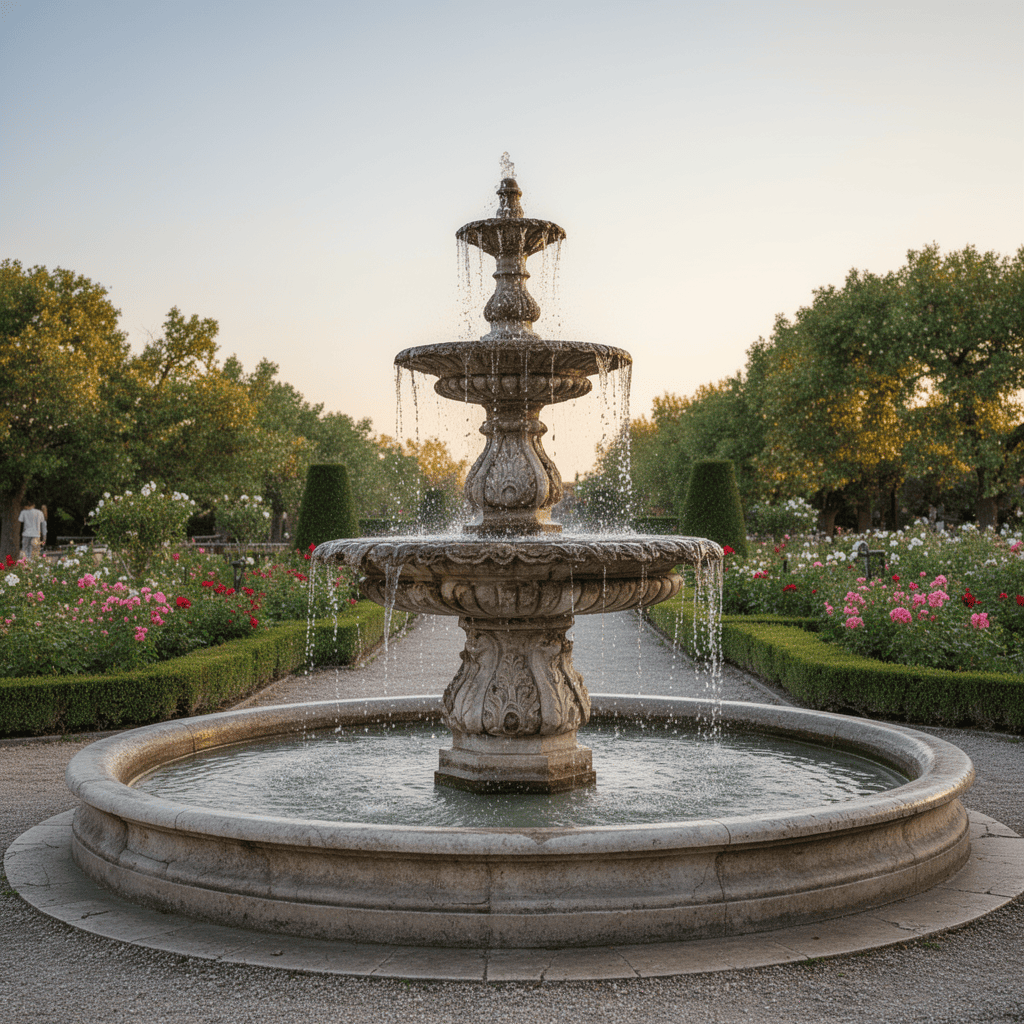}
        {FLUX.2-dev / GenAI-Bench-0490}
        {\textbf{TIFA}~0.43\hfill\textbf{VQAScore}~0.01}
        {1.00}{1.00}
        {A fountain with no water flowing.}
        {Water flows despite the prompt requesting none; the depicted fountain remains physically plausible.}
    \caption{
        \textbf{Prompt alignment versus world consistency.}
        TIFA and VQAScore assess prompt alignment, which does not necessarily reflect world consistency. 
        (a) Images receive high alignment scores despite low human world-consistency ratings. 
        (b) Images that deviate from their prompts receive low alignment scores despite high human world-consistency ratings. 
        These examples highlight the distinction between prompt alignment and world consistency.
        For example, the third image in (a) matches the requested juxtaposition of ancient buildings and futuristic transport, yet receives low world-consistency scores, illustrating that prompt faithfulness does not guarantee world plausibility.
    }
    \label{fig: apx_alignment_vs_wc}
\end{figure}

\begin{figure}[p]
    \centering
    \casegroup{(a) High HPSv3 but Low Human World-Consistency Scores}{}%
    \baselinecase
        {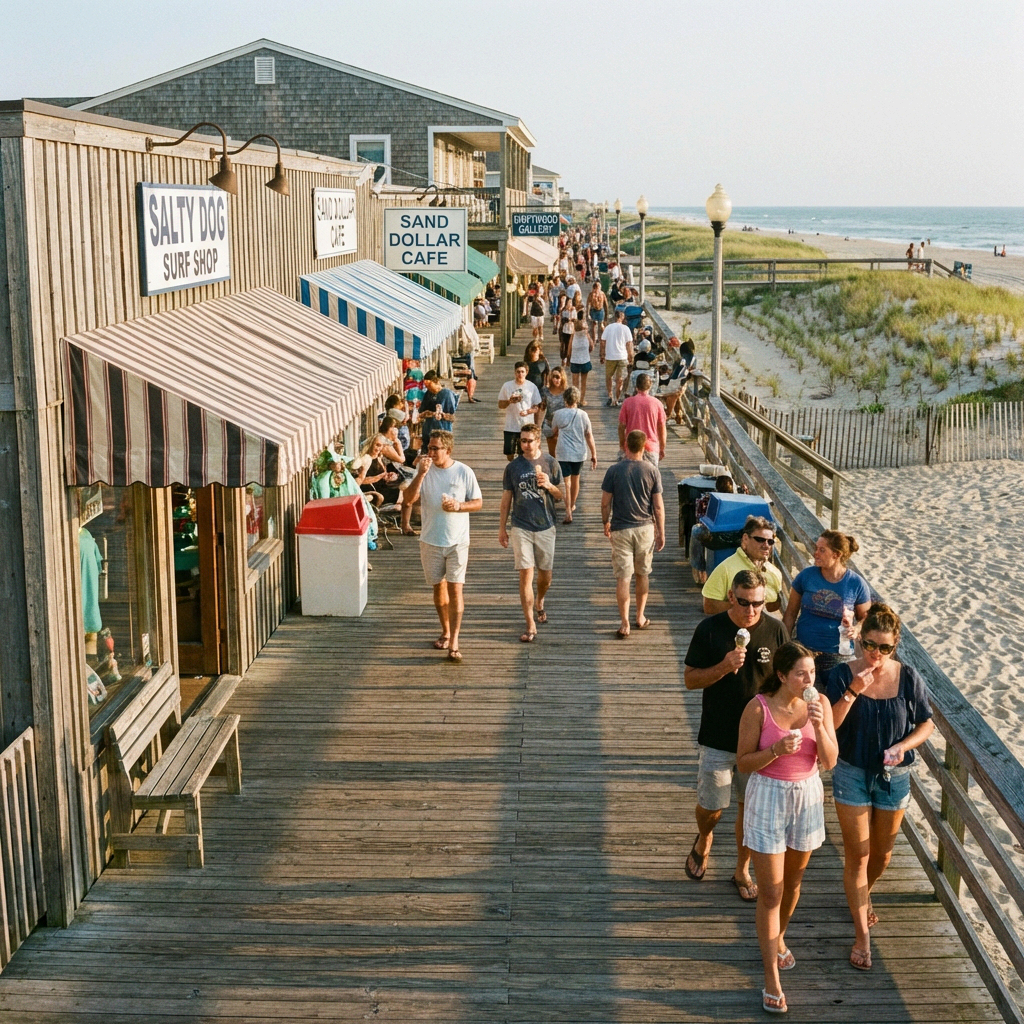}
        {Nano Banana Pro / GenAI-Bench-0791}
        {\textbf{HPSv3}~13.03}
        {0.61}{0.00}
        {The wooden boardwalk along the beach is lined with shops.}
        {The detailed boardwalk receives a high preference score, while TerraVis and all three raters give lower world-consistency scores.}%
    \hfill
    \baselinecase
        {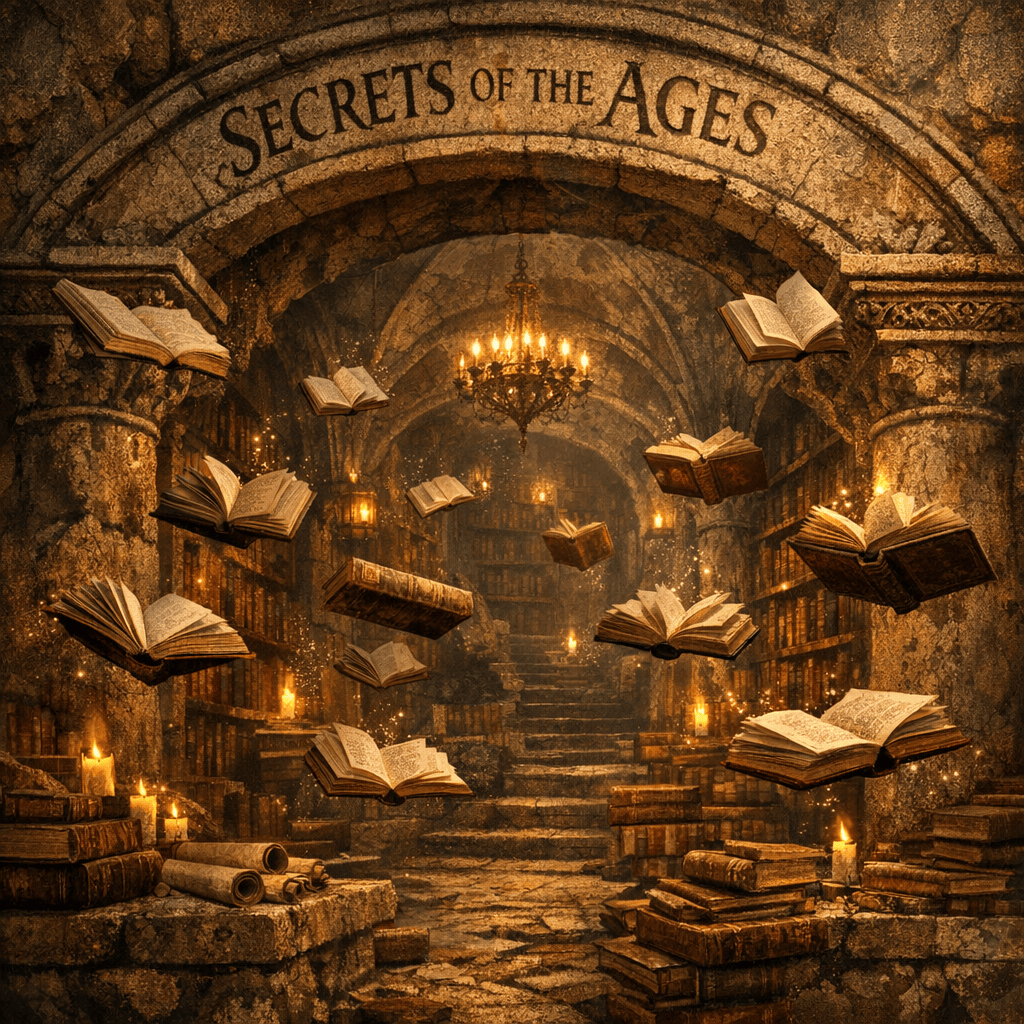}
        {GPT-Image-1.5 / GenAI-Bench-0237}
        {\textbf{HPSv3}~12.99}
        {0.01}{0.00}
        {An ancient library hidden beneath the earth, `Secrets of the Ages' inscribed on the archway, books floating around as if by magic.}
        {Books float without visible support in an otherwise richly detailed, dramatically lit library.}
    \par\vspace{7pt}\noindent%
    \baselinecase
        {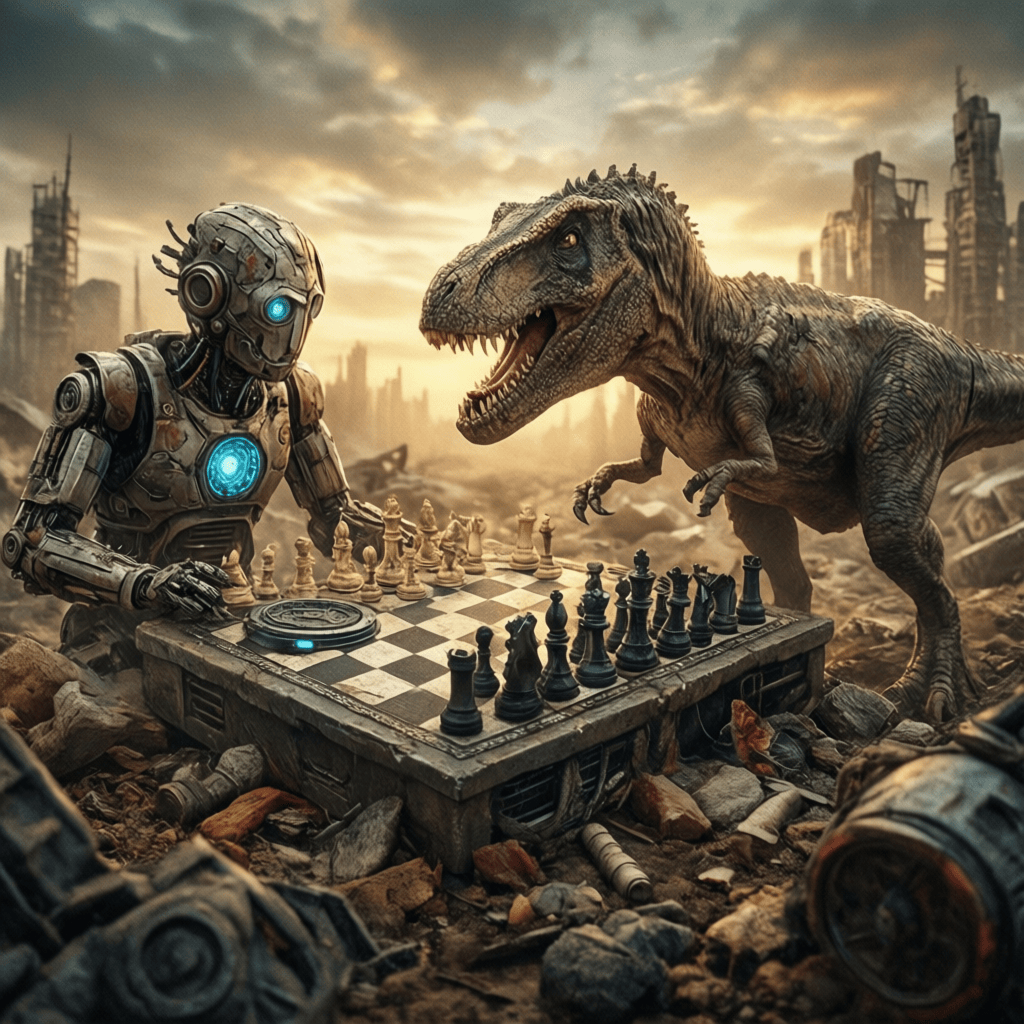}
        {Qwen-Image / GenAI-Bench-0568}
        {\textbf{HPSv3}~12.19}
        {0.02}{0.00}
        {A robot and a dinosaur play chess in a post-apocalyptic landscape.}
        {The polished fantasy scene receives a high preference score despite its implausible participants and interaction.}%
    \hfill
    \baselinecase
        {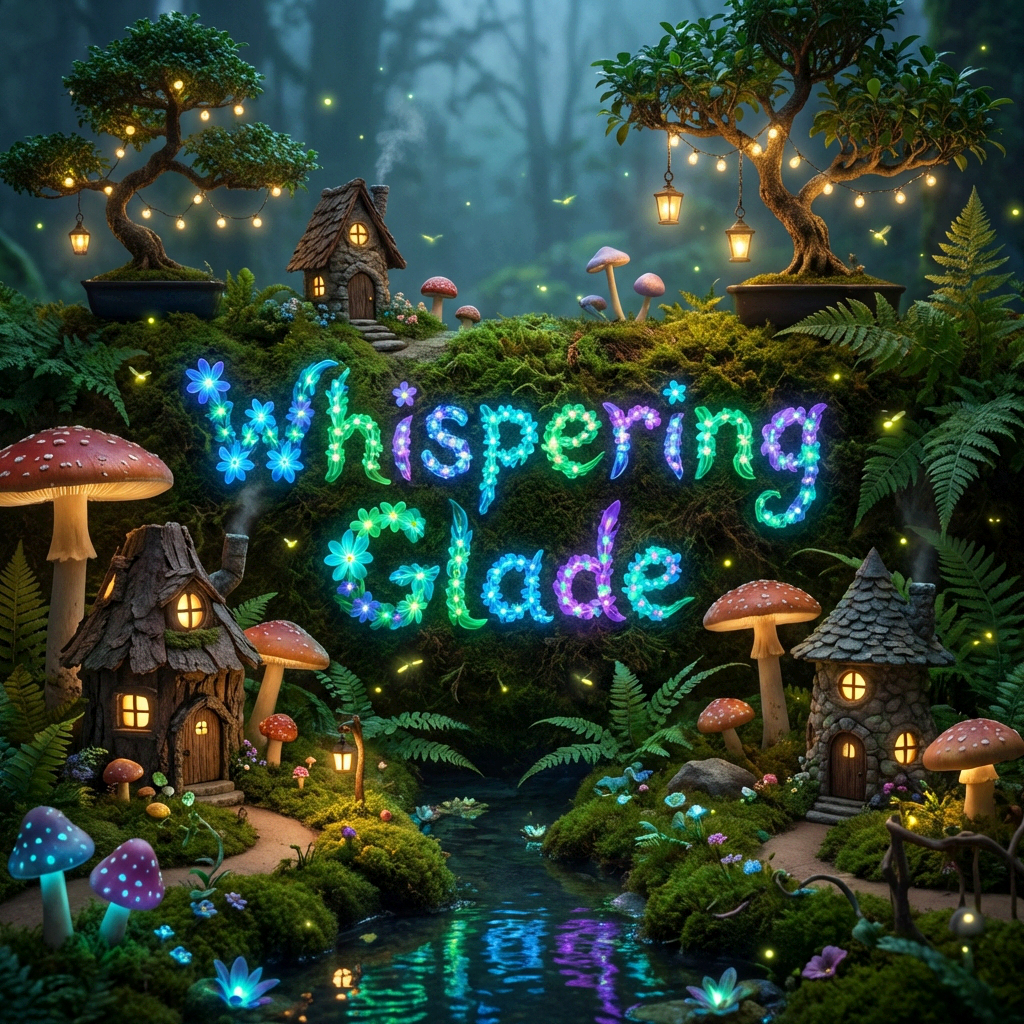}
        {Nano Banana Pro / GenAI-Bench-0673}
        {\textbf{HPSv3}~12.77}
        {0.37}{0.00}
        {A fairy garden at twilight, `Whispering Glade' spelled out in luminescent flowers.}
        {Glowing floral lettering and miniature fantasy dwellings receive a high preference score but low world-consistency scores.}
    \par\vspace{8pt}
    \casegroup{(b) Low HPSv3 but High Human World-Consistency Scores}{}%
    \baselinecase
        {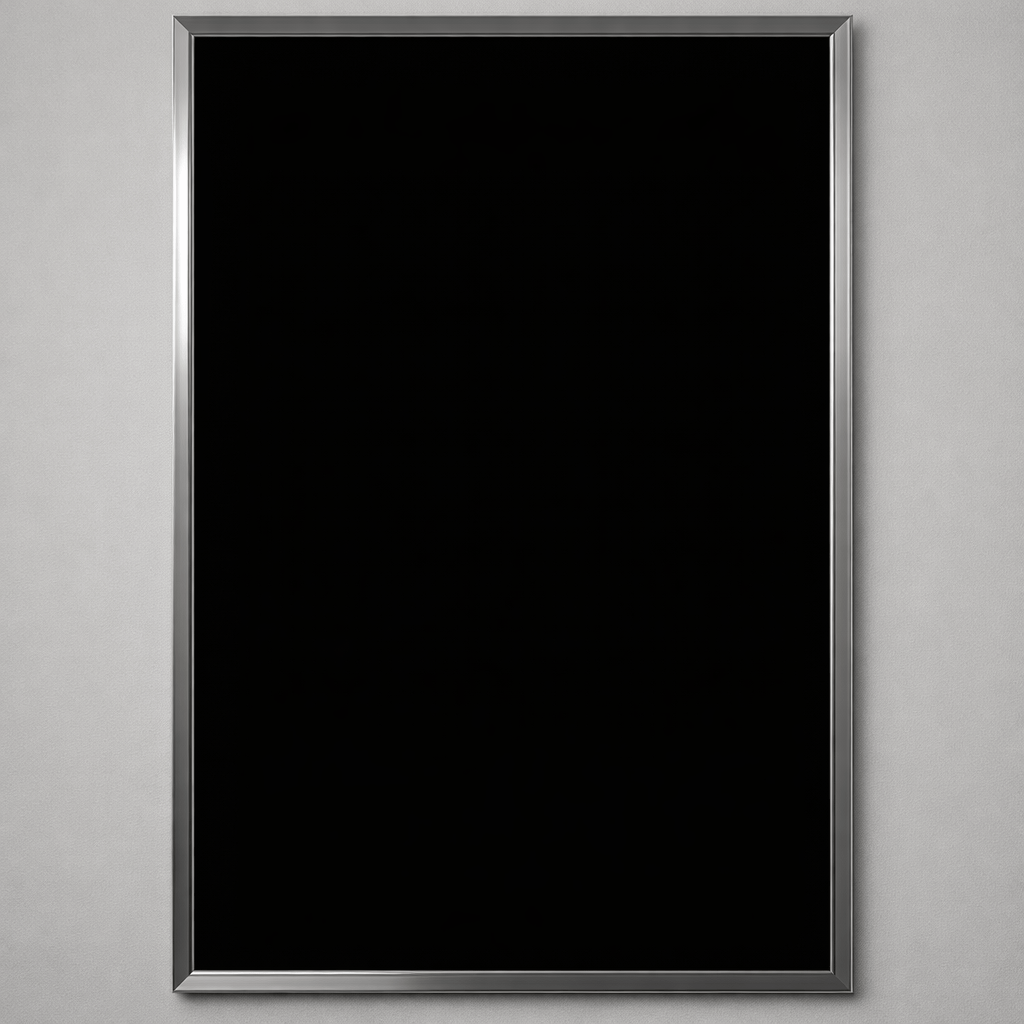}
        {GPT-Image-1.5 / GenAI-Bench-0486}
        {\textbf{HPSv3}~1.91}
        {1.00}{1.00}
        {A mirror without a reflection of light.}
        {The sparse composition receives a low preference score, while TerraVis and all raters assign maximum scores.}%
    \hfill
    \baselinecase
        {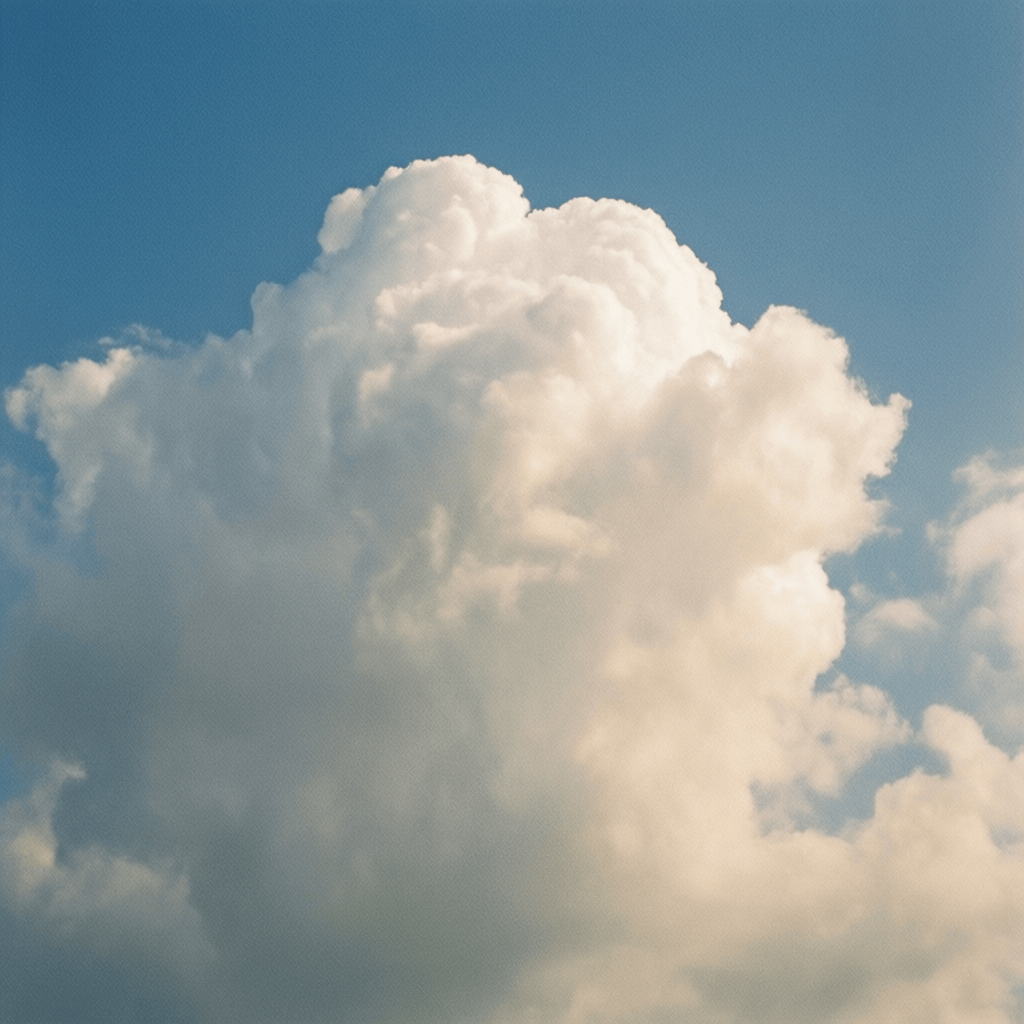}
        {Qwen-Image / GenAI-Bench-0451}
        {\textbf{HPSv3}~1.92}
        {1.00}{1.00}
        {A sky without a single cloud.}
        {The cloud contradicts the prompt but is physically plausible; preference and world-consistency scores diverge.}
    \par\vspace{7pt}\noindent%
    \baselinecase
        {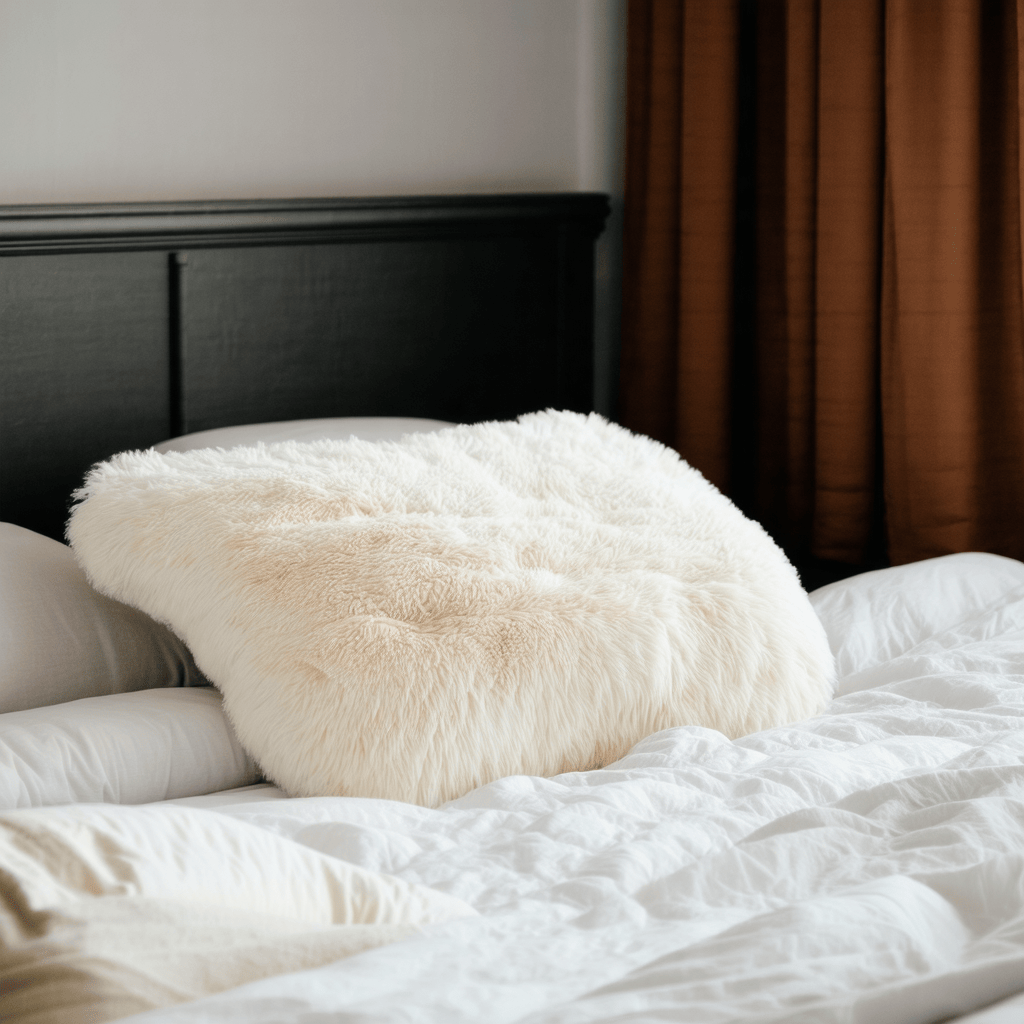}
        {SD 3.5-Large / GenAI-Bench-1312}
        {\textbf{HPSv3}~2.72}
        {1.00}{1.00}
        {A cozy bedroom without a fluffy pillow on the bed.}
        {The pillow contradicts the prompt but remains physically plausible; the preference score is low.}%
    \hfill
    \baselinecase
        {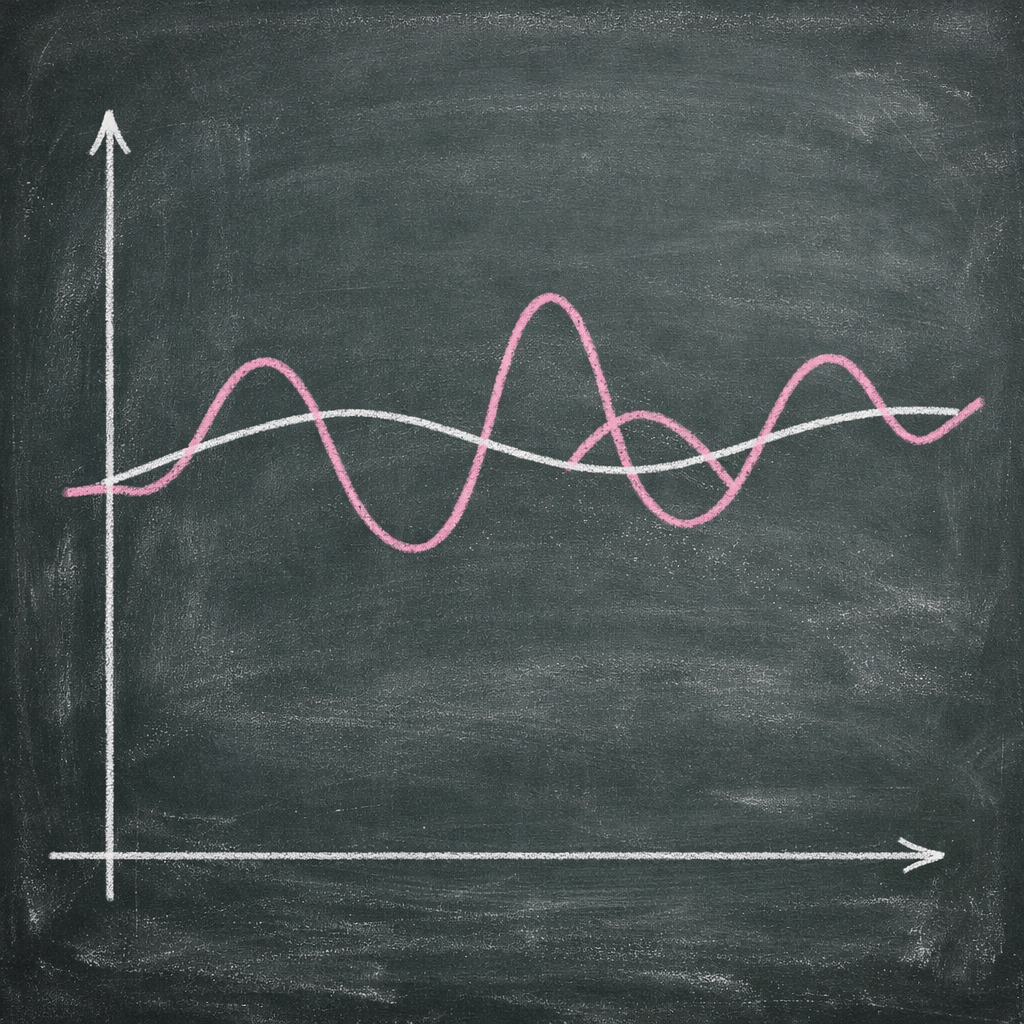}
        {GPT-Image-1.5 / GenAI-Bench-1567}
        {\textbf{HPSv3}~5.36}
        {N/A}{1.00}
        {A blackboard displays two curves; the pink one undulates more than the white one.}
        {Humans give maximum ratings; TerraVis marks the graph as out of scope and returns N/A.}
    \caption{\textbf{HPSv3 versus world consistency.}
        (a) Images receive high preference scores despite low human world-consistency ratings.
        (b) Images receive low preference scores despite high human world-consistency ratings.
        HPSv3 is shown on its original scale; TerraVis and normalized human scores use $[0,1]$.
        These examples highlight the distinction between human preference and world consistency.
        For instance, the third example in (a), depicting an ancient library with floating books and ``Secrets of the Ages'' inscribed on the archway, receives a high HPSv3 preference score despite low world-consistency scores from both TerraVis and human raters.
        }
    \label{fig: apx_hpsv3_vs_wc}
\end{figure}

\begin{figure}[p]
    \centering
    \casegroup{(a) High RAHF but Low Human World-Consistency Scores}{}%
    \baselinecase
        {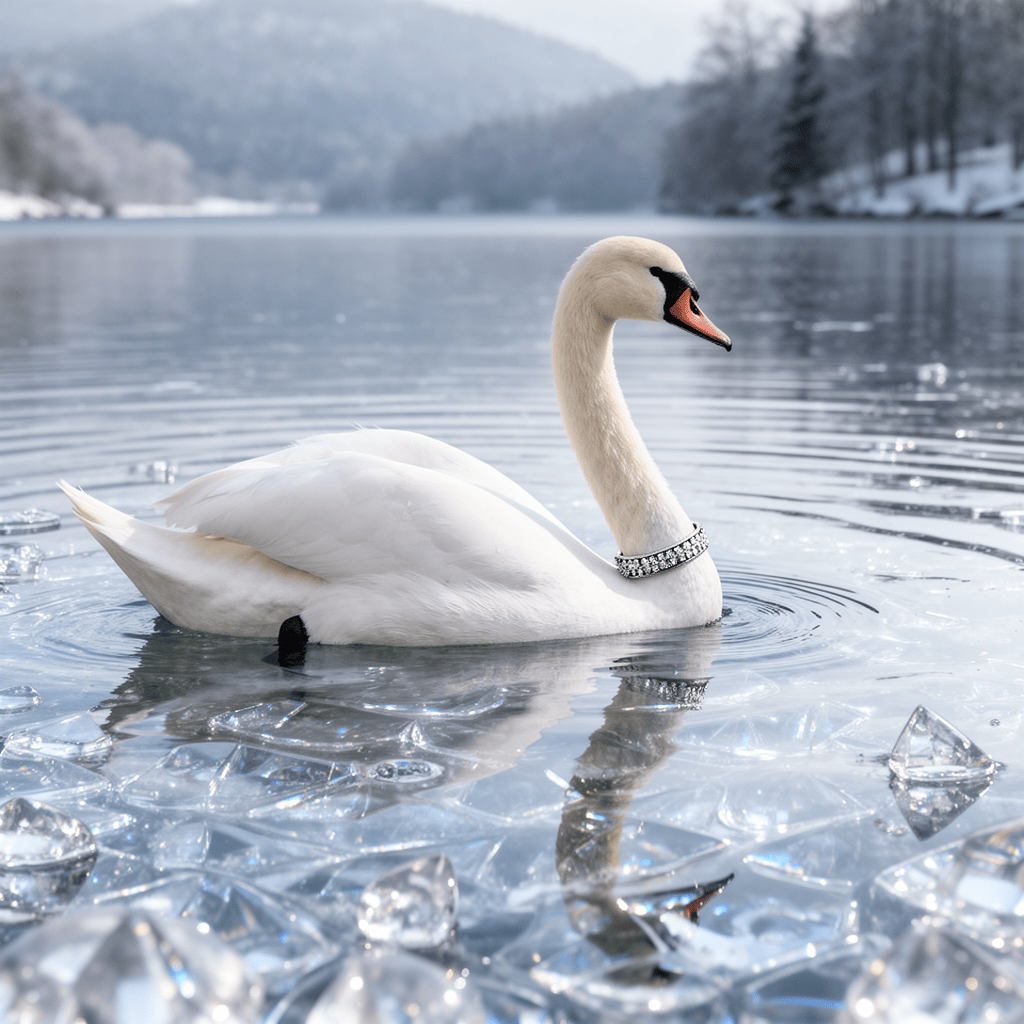}
        {FLUX.2-dev / GenAI-Bench-0064}
        {\rahfscores{0.72}{0.79}{0.75}{0.77}}
        {0.01}{0.00}
        {A swan with a silver anklet on a crystal lake.}
        {The lake surface mixes water and crystal-like solids, while the requested anklet is placed around the neck.}%
    \hfill
    \baselinecase
        {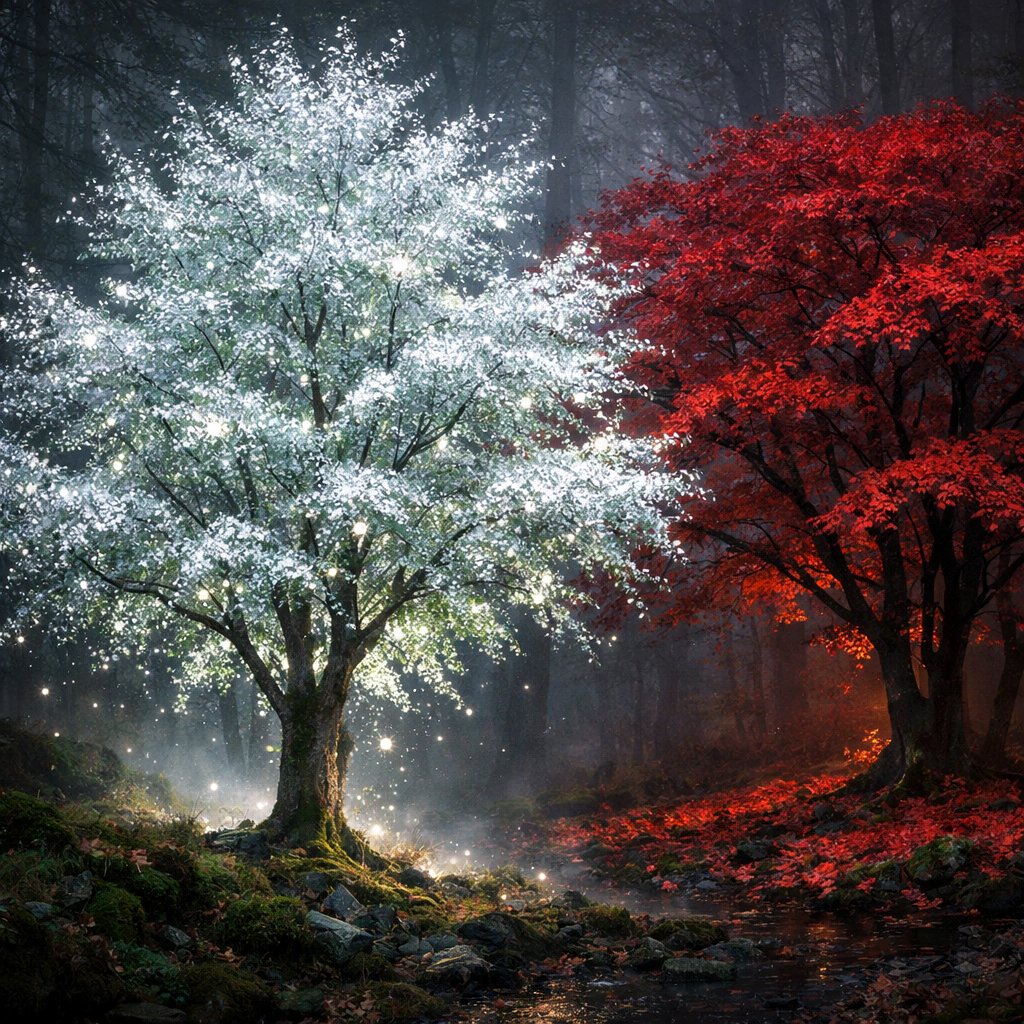}
        {GPT-Image-1.5 / GenAI-Bench-0243}
        {\rahfscores{0.66}{0.93}{0.52}{0.77}}
        {0.14}{0.08}
        {In a mysterious forest, the leaves of a green tree shimmer with a silvery glow, contrasting the dim leaves of a nearby red tree.}
        {The tree appears to emit a silvery glow despite its high RAHF plausibility score.}
    \par\vspace{7pt}\noindent%
    \baselinecase
        {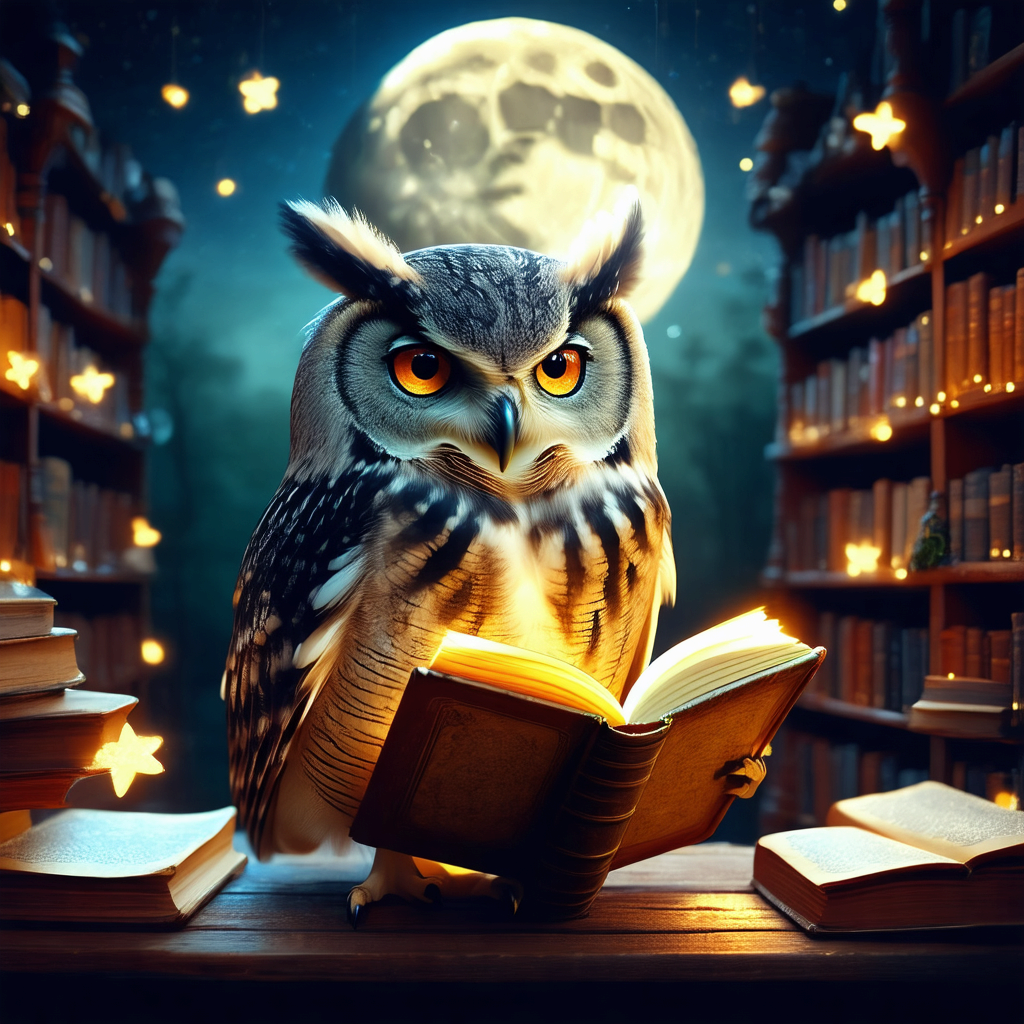}
        {SD 3.5-Large / GenAI-Bench-0623}
        {\rahfscores{0.77}{0.85}{0.78}{0.82}}
        {0.01}{0.08}
        {An owl with a tiny book in a moonlit library.}
        {The owl appears to read a glowing book amid floating lights, despite high RAHF scores.}%
    \hfill
    \baselinecase
        {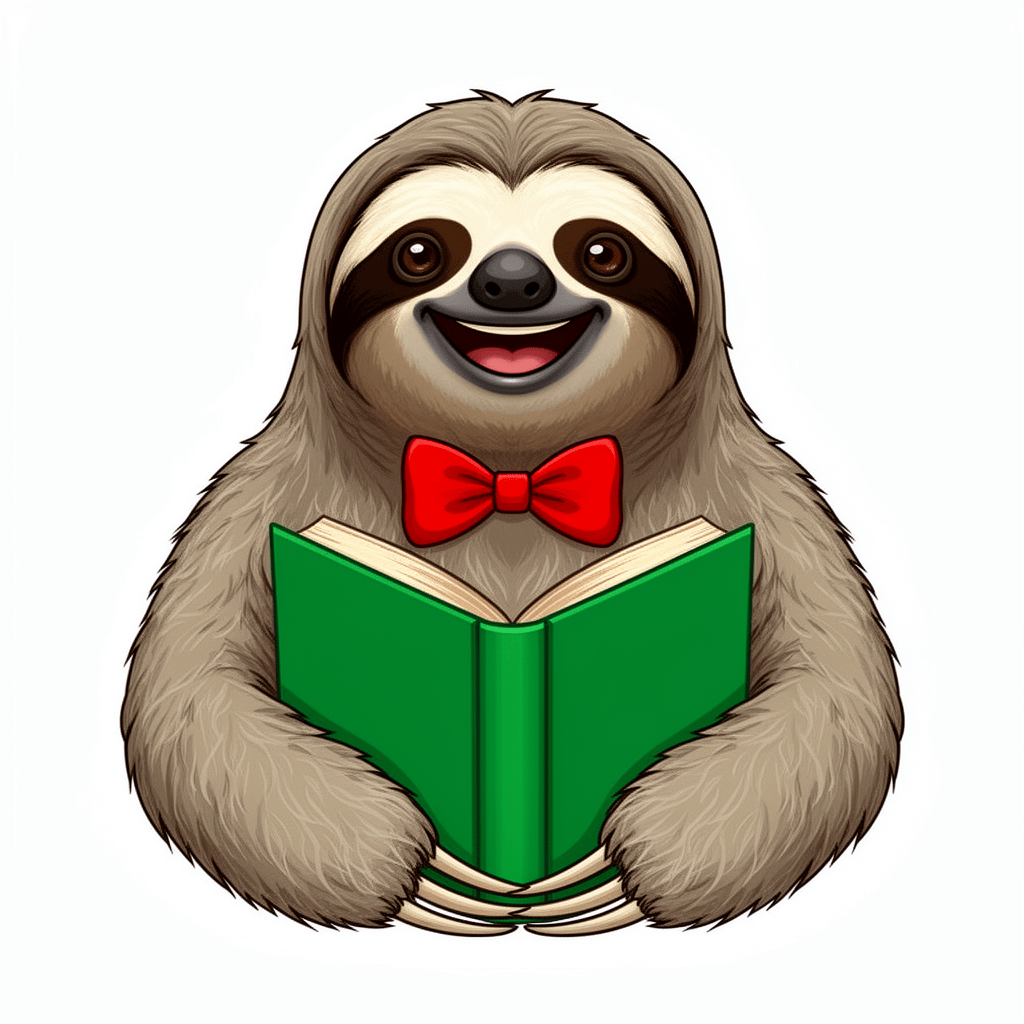}
        {FLUX.2-dev / GenAI-Bench-0545}
        {\rahfscores{0.71}{0.92}{0.69}{0.84}}
        {0.14}{0.08}
        {A smiling sloth wearing a bowtie and holding a book.}
        {The sloth has a human-like smile and holds a book; the stylized scene receives high RAHF scores.}
    \par\vspace{8pt}
    \casegroup{(b) Low RAHF but High Human World-Consistency Scores}{}%
    \baselinecase
        {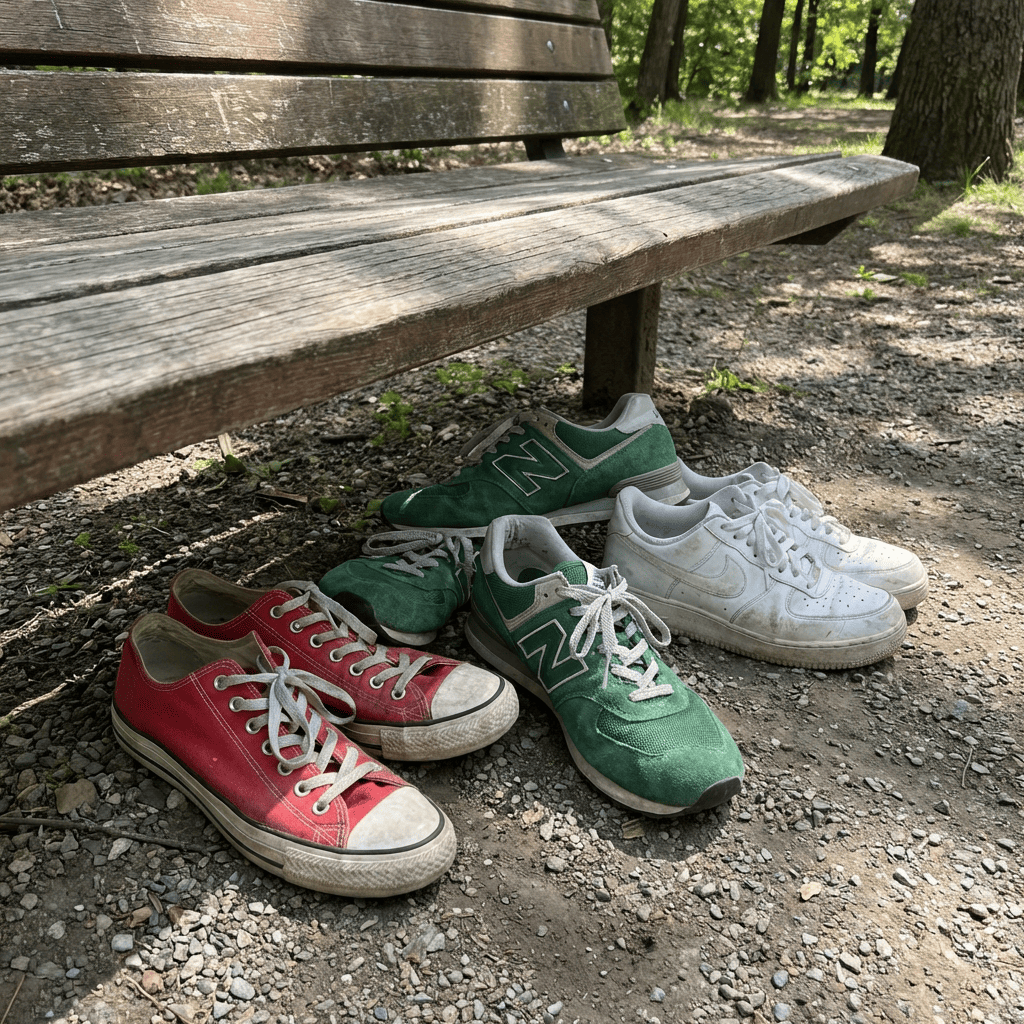}
        {Nano Banana Pro / GenAI-Bench-0906}
        {\rahfscores{0.38}{0.56}{0.34}{0.53}}
        {1.00}{1.00}
        {Under the bench, there are four pairs of sneakers: one pair is red, two pairs are green, and the last pair is white.}
        {The visible shoes and bench appear plausible; TerraVis and all human raters assign maximum scores.}%
    \hfill
    \baselinecase
        {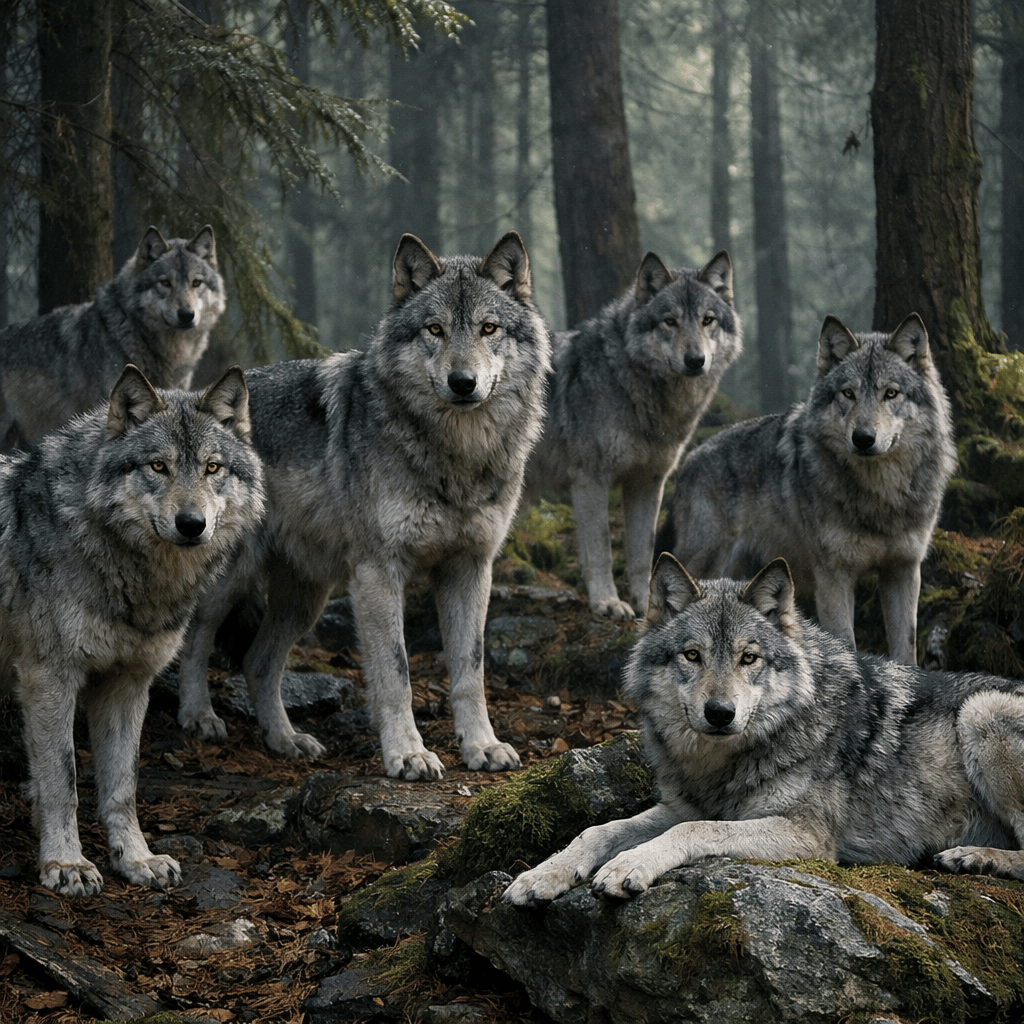}
        {GPT-Image-1.5 / GenAI-Bench-0874}
        {\rahfscores{0.41}{0.29}{0.60}{0.54}}
        {1.00}{1.00}
        {In the forest, there's a pack of wolves, all of them gray.}
        {The wolves appear plausible in their forest setting, despite a low RAHF plausibility score.}
    \par\vspace{7pt}\noindent%
    \baselinecase
        {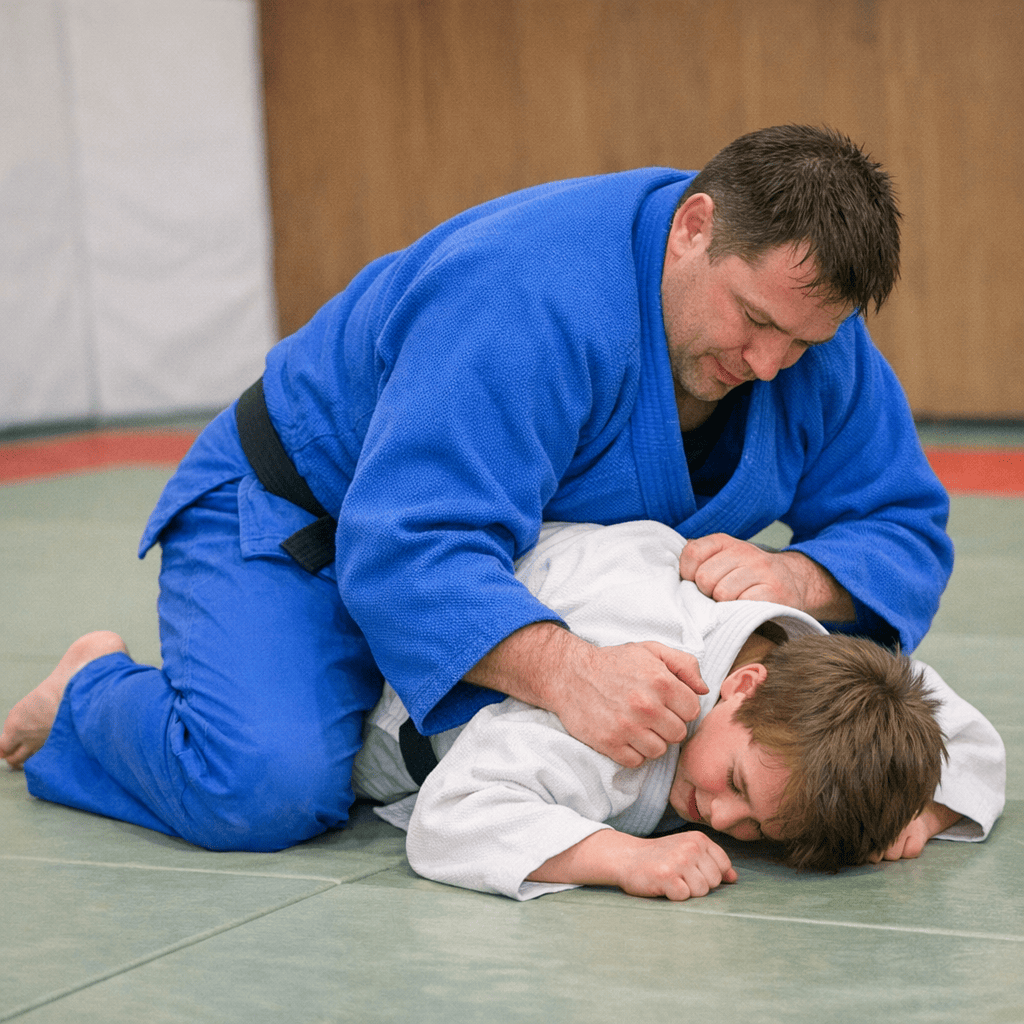}
        {GPT-Image-1.5 / GenAI-Bench-0196}
        {\rahfscores{0.41}{0.35}{0.56}{0.49}}
        {1.00}{1.00}
        {The larger person wears blue and the smaller person does not.}
        {The two people and their contact appear plausible, despite the low RAHF plausibility score.}%
    \hfill
    \baselinecase
        {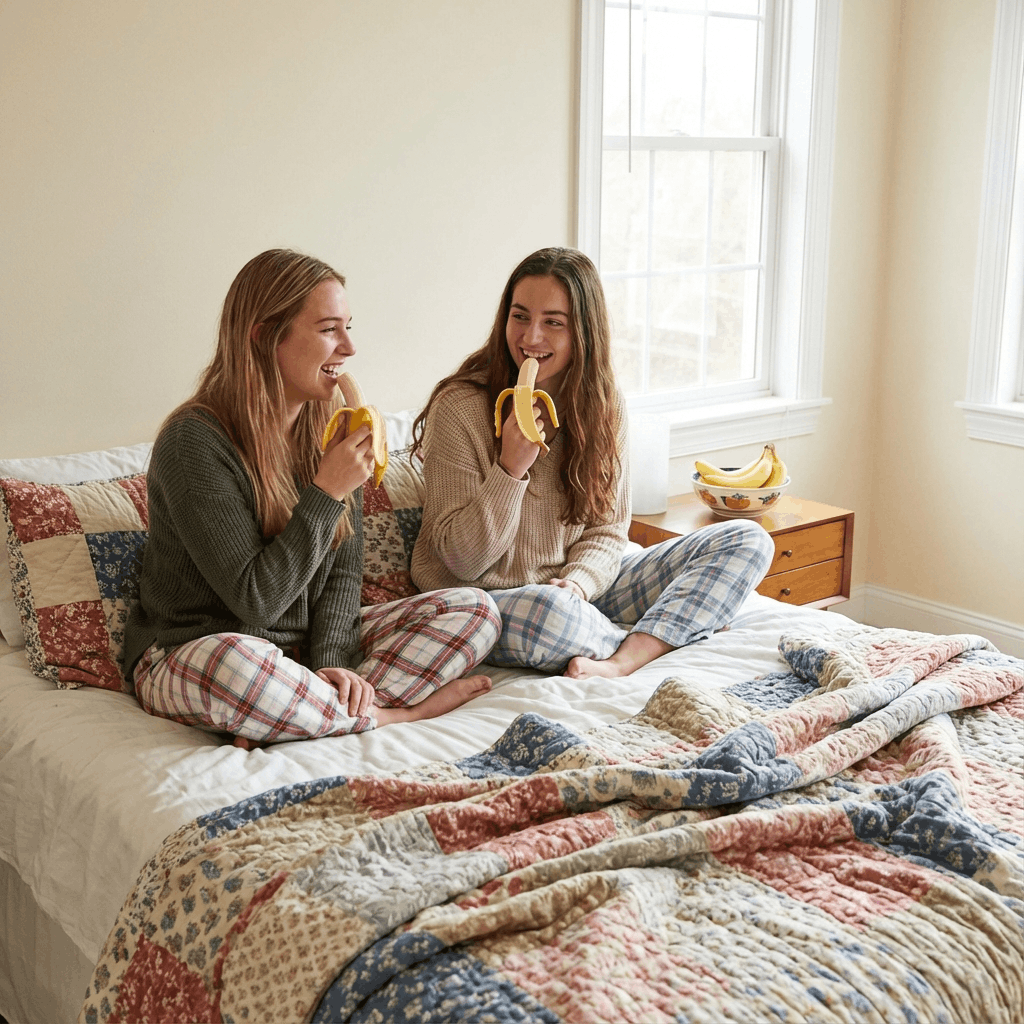}
        {Nano Banana Pro / COCO-T2I-106}
        {\rahfscores{0.42}{0.36}{0.65}{0.41}}
        {1.00}{1.00}
        {Two girls sitting on a bed eating bananas together.}
        {Both people and their interaction appear plausible, despite low RAHF plausibility and aesthetics scores.}
    \caption{
        \textbf{RAHF versus world consistency.}
        RAHF's overall score and its plausibility, alignment, and aesthetics scores are reported separately.
        (a) Images receive high RAHF scores despite low human world-consistency ratings.
        (b) Images receive low RAHF scores despite high human ratings.
        RAHF's plausibility head was trained to detect artifacts using RichHF-18K~\cite{rahf}, which limits its coverage to the types of artifacts represented in the training set and may not generalize well to images generated by newer text-to-image models. 
        For example, the first image in (b) suggests that RAHF incorrectly penalizes an artifact-free and world-consistent image generated by Nano Banana Pro.
    }
    \label{fig: apx_rahf_vs_wc}
\end{figure}

We provide more examples of four patterns of disagreement in world-consistency evaluation, as shown in \figref{fig: apx_disagreement}.
We observe that TerraVis can perform well in stylized images, such as the first example in Human Oversights, where human raters fail to find the failures that the mice stand calmly next to the cat instead of fleeing or showing fear despite being beside a predator, which is a behavior-event violation.
Also TerraVis can miss violations in stylized images, such as the first example in Missed Violations, where several lanterns appear to float below the branches without visible support, a support violation overlooked by TerraVis.
Furthermore, Figures~\ref{fig: apx_alignment_vs_wc}, \ref{fig: apx_hpsv3_vs_wc}, \ref{fig: apx_rahf_vs_wc} illustrate the differences between existing evaluation metrics, including text-image alignment, human preference, and artifact detection (i.e., RAHF), and world consistency evaluation. 
We summarize three key findings.

\textbf{First, prompt faithfulness and world consistency capture distinct dimensions}. 
Prompt alignment measures how faithfully an image reflects the content specified in the prompt, even when that content itself is implausible in the real-world. 
In contrast, world consistency evaluates whether the generated image conforms to real-world constraints. 
Therefore, a high degree of prompt faithfulness does not necessarily imply high world consistency.

\textbf{Second, HPSv3 primarily captures aesthetic preference rather than world consistency}. 
Its preference-oriented evaluation can correlate with perceived visual quality, but does not explicitly assess whether the depicted content is consistent with real-world knowledge or physical plausibility.

\textbf{Third, RAHF is primarily designed to detect rendering artifacts, and its coverage of world consistency is therefore limited}. 
In particular, its plausibility head was trained to detect artifacts using RichHF-18K~\cite{rahf}, which constrains its coverage to the types of artifacts represented in the training data. 
Consequently, it may not fully capture novel forms of artifacts arising from newer text-to-image models, such as Nano Banana Pro.

\end{document}